\documentclass[10pt,twocolumn,letterpaper]{article}
\usepackage[pagenumbers]{cvpr}

\usepackage{cuted} 

\usepackage{currfile} 

\usepackage{caption} 

\usepackage[T1]{fontenc}
\usepackage{microtype}
\usepackage{xurl}
\usepackage{etoolbox}
\makeatletter
\newbox\redate@labels\newif\ifredate@inlabel\newif\ifredate@newlist
\pretocmd{\@outputpage}{\global\setbox\redate@labels\box\@labels
  \if@inlabel\global\redate@inlabeltrue\else\global\redate@inlabelfalse\fi\global\@inlabelfalse
  \if@newlist\global\redate@newlisttrue\else\global\redate@newlistfalse\fi\global\@newlistfalse}{}{\errmessage{\string\@outputpage\space not patched}}
\apptocmd{\@outputpage}{\global\setbox\@labels\box\redate@labels
  \ifredate@inlabel\global\@inlabeltrue\fi\ifredate@newlist\global\@newlisttrue\fi}{}{\errmessage{\string\@outputpage\space not patched}}
\makeatother
\usepackage{newunicodechar}
\newunicodechar{−}{\ensuremath{-}}
\newunicodechar{✓}{\ding{51}}
\newunicodechar{✗}{\ding{55}}
\newunicodechar{↑}{\ensuremath{\uparrow}}
\newunicodechar{↓}{\ensuremath{\downarrow}}
\usepackage{pifont,colortbl,tabularx}
\definecolor{oursrow}{HTML}{F4E6E6} 
\usepackage{algorithm,listings}
\usepackage[varqu,scaled=0.95]{zi4}
\definecolor{codecomment}{HTML}{626262}
\usepackage{amsthm}
\theoremstyle{plain}
\newtheorem{theorem}{Theorem}
\newtheorem{proposition}{Proposition}
\newtheorem{corollary}{Corollary}
\newtheorem{lemma}{Lemma}
\newtheorem*{restatedplain}{\restatedname}
\theoremstyle{definition}
\newtheorem{assumption}{Assumption}
\newtheorem*{restateddefinition}{\restatedname}
\newcommand{\restatedname}{}
\AtEndPreamble{\crefname{assumption}{Assumption}{Assumptions}}
\usepackage{dblfloatfix}
\makeatletter
\usepackage[skins]{tcolorbox}
\usepackage{environ}
\pdfmapfile{+manrope.map}
\DeclareFontFamily{T1}{manrope}{}
\DeclareFontShape{T1}{manrope}{b}{n}{<-> s * [0.88] manrope-bold}{}
\DeclareFontShape{T1}{manrope}{eb}{n}{<-> s * [0.88] manrope-extrabold}{}
\definecolor{brand}{HTML}{003366}
\definecolor{cardback}{HTML}{E0E7ED} 
\definecolor{cardink}{HTML}{1C2B33}
\definecolor{ours}{HTML}{94070A}
\DeclareRobustCommand{\ours}{\textcolor{ours}{\textbf{ReDate}}}
\AtBeginDocument{\pdfstringdefDisableCommands{\def\ours{ReDate}}}
\def\card@abstract{}
\RenewEnviron{abstract}{\global\let\card@abstract\BODY}
\def\@maketitle{%
  \begin{tcolorbox}[enhanced,frame hidden,colback=cardback,arc=10pt,boxsep=0pt,left=0.55cm,right=0.55cm,top=0.4cm,bottom=0.4cm,
      width=\textwidth,grow to left by=0.55cm,grow to right by=0.55cm]
    \color{cardink}\centering
    {\usefont{T1}{manrope}{eb}{n}\fontsize{16pt}{19pt}\selectfont \textcolor{brand}{Old Is Not False}: Valid-Time Memory for Interactive Video World Models\par}
    \vskip 0.12in
    {\fontsize{10.8pt}{13pt}\selectfont
    \begin{tabular}[t]{c}
{\usefont{T1}{manrope}{b}{n}Ziming Wang\textsuperscript{1}}, {\usefont{T1}{manrope}{b}{n}Chuyue Gong\textsuperscript{1}}, {\usefont{T1}{manrope}{b}{n}Hongliang Cao\textsuperscript{2}},\\
{\usefont{T1}{manrope}{b}{n}Jingmin Zhu\textsuperscript{3}}, {\usefont{T1}{manrope}{b}{n}Chutian Wang\textsuperscript{4}}, {\usefont{T1}{manrope}{b}{n}Yuheng He\textsuperscript{4}}, {\usefont{T1}{manrope}{b}{n}Xiuyu Li\textsuperscript{5}}\\
\\
\fontsize{10pt}{12pt}\selectfont \textsuperscript{1}The Hong Kong University of Science and Technology, \textsuperscript{2}Beihang University,\\
\fontsize{10pt}{12pt}\selectfont \textsuperscript{3}Monash University, \textsuperscript{4}Sun Yat-sen University, \textsuperscript{5}Renmin University of China
    \end{tabular}\par}
    \vskip 0.16in
    \begin{minipage}{\dimexpr\linewidth-0.5cm\relax}
      \fontsize{11pt}{13.6pt}\selectfont\setlength{\parindent}{0pt}
      \card@abstract
    \end{minipage}
    \vskip 0.10in
    \begin{minipage}{\dimexpr\linewidth-0.5cm\relax}\centering
      \includegraphics[height=30pt]{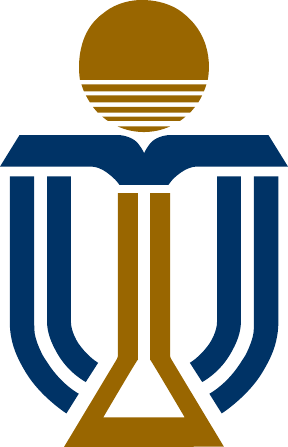}\hspace{18pt}\includegraphics[height=30pt]{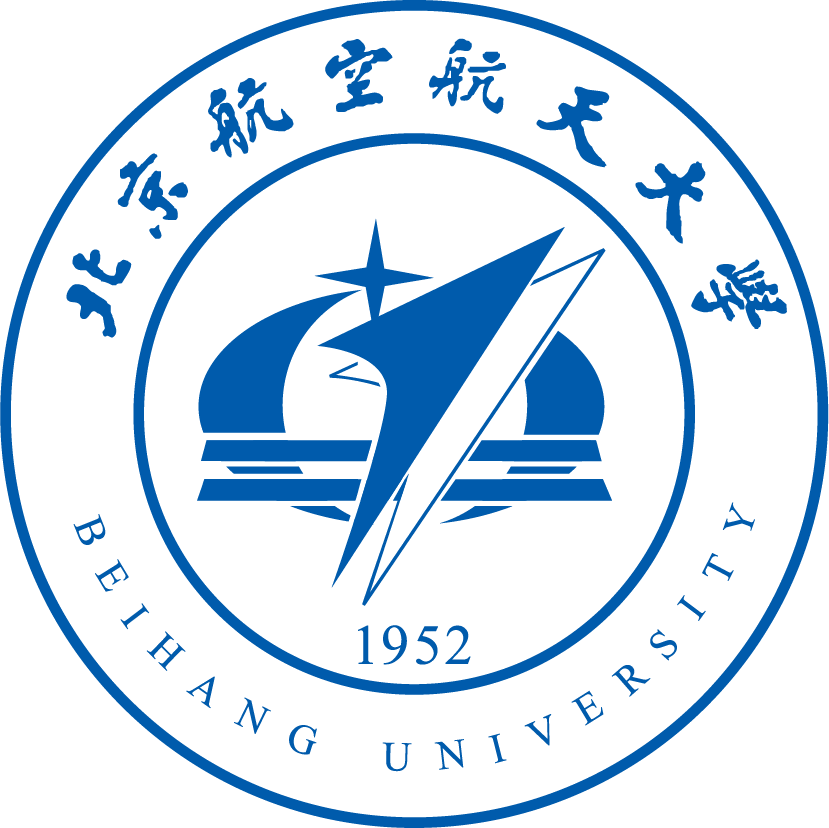}\hspace{18pt}\includegraphics[height=30pt]{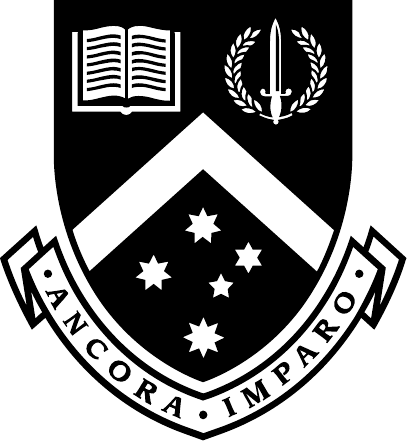}\hspace{18pt}\includegraphics[height=30pt]{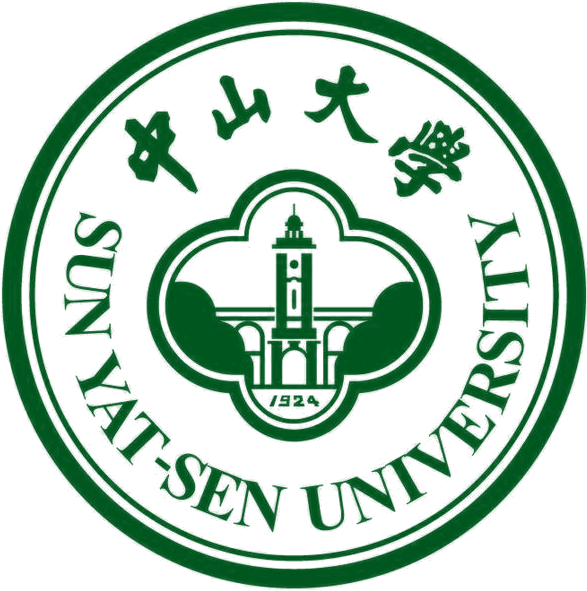}\hspace{18pt}\includegraphics[height=30pt]{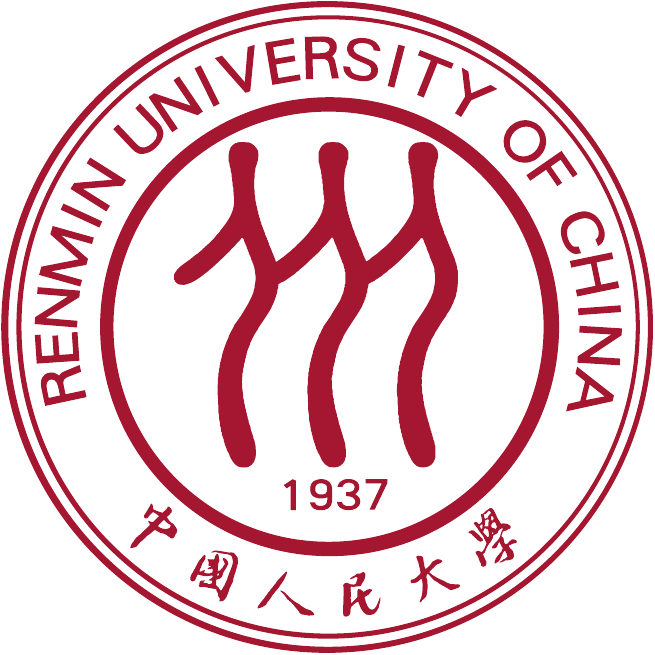}
    \end{minipage}
    \vskip 0.02in
  \end{tcolorbox}}
\makeatother
\definecolor{cvprblue}{rgb}{0.21,0.49,0.74}
\usepackage[pagebackref,breaklinks,colorlinks,allcolors=cvprblue]{hyperref}
\def\confName{CVPR}
\def\confYear{2027}
\title{Old Is Not False: Valid-Time Memory for Interactive Video World Models}
\author{}
\hypersetup{pdftitle={Old Is Not False: Valid-Time Memory for Interactive Video World Models},pdfauthor={Ziming Wang, Chuyue Gong, Hongliang Cao, Jingmin Zhu, Chutian Wang, Yuheng He, Xiuyu Li}}
\begin{document}
\begin{abstract}

Interactive video world models draw each frame from a memory bank of earlier frames stamped with their capture times. With frames, weights and noise fixed, reversing the timestamps of two views flips the state that generators of two families draw: they take the newest timestamp as the truth. That reading is right as long as states never return. After an agent undoes an action, the newest view of a region is outdated and the only view of the present is older; memories for changing worlds delete that older view, and memories that keep it hand the generator the past. Temporal databases separate when a record is stored from when its content holds, its valid time. \ours{} keeps every frame and dates each retrieved token by its valid time, read from the agent's action log. The frozen generator then reads a reverted region as one that has just reached a new state, so \ours{} needs no training. We prove that trusting the newest timestamp is right exactly when every current view is dated after every outdated one, that without a view taken after the return only the action log tells a reverted region from an up-to-date one, and that a generator able to draw from its own prediction gains more from re-dating than from deletion when the moved tokens keep little weight. After an undo, \ours{} raises the share of FAR's steps showing the current state from 0.36 to 0.62 and HY-WorldPlay's from 0.003 to 0.19. We also propose a calibrated state metric to test any memory after undos.

\end{abstract}

\maketitle

\begin{figure*}[t]
\centering
\includegraphics[width=\linewidth,height=0.82\textheight,keepaspectratio]{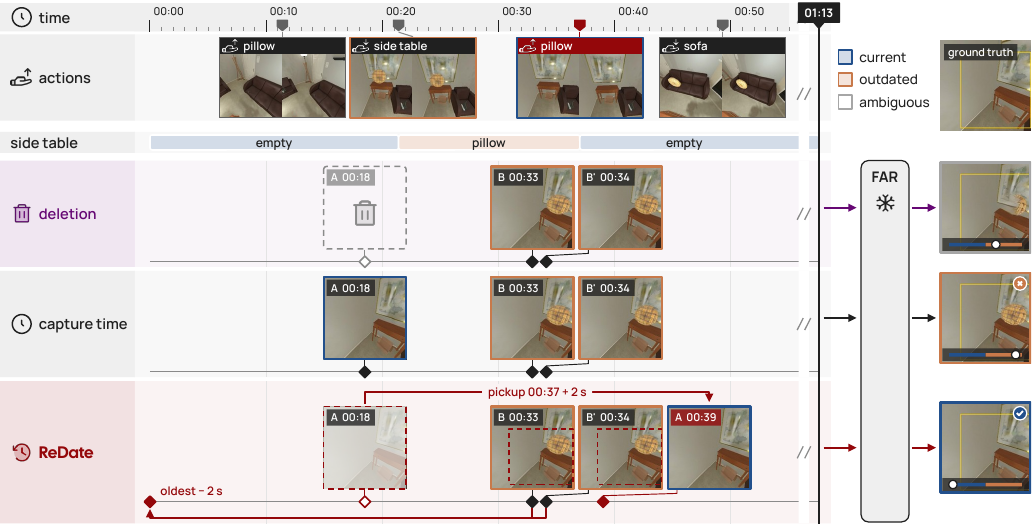}
\caption{One undo on an editor's timeline, read by FAR's frozen generator~\citep{kim2026learning}. Deletion leaves a gap where frame A was, capture time keeps A before B and B\('\), and \ours{} drags A to 2~s after the restoring pickup and moves the pillow's tokens before the oldest frame; the outputs show a ghost, the pillow and the empty table.\label{fig:teaser}}
\end{figure*}

\section{Introduction}\label{sec:introduction}

An \emph{interactive video world model} takes an agent's action and predicts the next frame of the scene~\citep{bruce2024genie,parkerholder2025genie,nvidia2025cosmos,chen2024diffusion}. Researchers use these models to train decision-making agents in imagined rollouts~\citep{hafner2025training}, evaluate robot and game policies in closed loop~\citep{team2025evaluating,quevedo2026worldgym,zhang2026world}, render playable worlds in real time~\citep{alonso2024diffusion,valevski2025diffusion,sun2025worldplay,zhang2026matrixgame4}, and plan navigation or manipulation trajectories~\citep{bar2025navigation,guo2026ctrl}. A video transformer attends to a short window of frames, so a rollout that returns to a place minutes later needs an external \emph{memory bank}~\citep{xiao2025worldmem,yu2025context,li2025vmem,kim2026learning,chen2026past}. As the agent moves, each frame enters this bank with its camera pose and a \emph{timestamp} recording its \emph{capture time}. At each step, retrieval selects \emph{context frames} that overlap with the \emph{target pose} the camera now visits~\citep{ren2025gen3c,wu2025video,ma2026closing}, or compresses past views into spatial tokens and scene states~\citep{huang2025memory,zhang2025frame}. In a static room, any frame from the target pose shows what the camera should see~\citep{duan2025worldscore,lian2025loopnav,ye2026mind}.

Rooms change when an agent acts in them, and everyday actions often undo earlier ones~\citep{weihs2021visual,mahajan2025do}. In unscripted EPIC-KITCHENS-100 videos~\citep{damen2022rescaling}, over half of opened cupboards, drawers and fridges are closed again, over half of switch-ons are followed by a switch-off, and most undos follow within half a minute. In the AI2-THOR~\citep{kolve2017ai2} room of \cref{fig:teaser}, the agent stores frame A of an empty side table from the doorway, puts a pillow on the table and stores frames B and B\('\) from the doorway, then picks the pillow up from across the room and returns. Frame A now shows the current state, the empty table, and the newer frames B and B\('\) show an outdated one. In FAR's benchmark~\citep{kim2026learning}, a third of the revisits that find a region in an earlier state follow such an \emph{undo}, which the revisited viewpoint never saw.

When a memory hands the frozen generator an older and a newer view of one region, which view does it draw, and what decides it? Holding context frames, weights and noise fixed, we alter only the timestamps of generators of two families (\cref{sec:finding}): FAR~\citep{kim2026learning} and the WorldMem baseline~\citep{xiao2025worldmem}, which add timestamp embeddings to frame tokens, and HY-WorldPlay~\citep{sun2025worldplay,wu2025hunyuanvideo}, which orders history through temporal rotary positions. Reversing the timestamps or swapping the rotary positions of two views flips the rendered region from the newer view's state to the older one's. A frozen generator reads timestamps as a soft \emph{latest-timestamp rule}, the last-writer-wins rule of replicated databases~\citep{johnson1975maintenance,shapiro2011conflict}, weighting views by how recently their timestamps claim they were confirmed~\citep{cao2026recency}.

Why do current memories show the past after an undo? Benchmarks test whether a world model keeps the states that actions create~\citep{chen2026memobench,lu2026current,zhang2026mbench,zhou2026worldmembench}, and memories for changing worlds treat a view that a newer one contradicts as false for good. \emph{Deletion-based memory} invalidates, masks or decays earlier views at a change~\citep{yang2026permavid,gu2026worldcraft,wu2026echo,zheng2026mem,wang2026latent} or replaces them with a symbolic or depth state~\citep{anon2026memworld,anon2026statecommit,venturoli2026statecraft}, erasing frame A and leaving a blurry ghost. \emph{State filtering} withholds frames that disagree with a state table~\citep{yang2026oneira,wang2026any}, dropping whole context frames whenever one region in them is outdated. Memories that keep every frame under capture time~\citep{xiao2025worldmem,kim2026learning} keep frame A older than B, so the generator draws the pillow. Long-rollout \emph{re-timing} rules shift or compress positions by frame age or attention role~\citep{sun2025worldplay,qian2026matrix,xu2026ucm,yi2025deep,kim2026memrope,yang2025longlive}, moving current and outdated views alike, so they cannot tell a reverted table from an up-to-date one.

Which timestamps make the generator draw the present, and is moving outdated content in time better than deleting it? Temporal databases separate the \emph{transaction time} when a record enters the store from the \emph{valid time} when its content holds~\citep{snodgrass1985taxonomy}, as do belief revision~\citep{alchourron1985logic,katsuno1991difference}, dynamic scene mapping~\citep{schmid2024khronos,saavedraruiz2025perpetua}, and temporal agent memory~\citep{rasmussen2025zep,wang2026toki}, each storing the validity intervals that its writers supply. Capture time is transaction time; frame A's valid time ended when the pillow arrived and began anew when it left, so old is not false. \ours{} derives valid time from the agent's \emph{action log}, keeps every frame and re-dates the retrieved tokens (\cref{sec:memory}). When an action restores a region to the state of an earlier frame, \ours{} dates that frame just after the restoring action and recalls it into the context; when a region has changed since capture, it moves only the frame's \emph{outdated tokens} before the oldest retrieved frame, which keeps the frame whole where deletion leaves a hole. Under these timestamps a reverted region reads as one that has just reached a new state, where the latest-timestamp reading is right, so \ours{} changes only what the frozen generator is shown. It raises FAR's accuracy on reverted steps from 0.36 to 0.62, ahead of rules adapted from ten published memories and of deleting the same tokens, and carries to HY-WorldPlay.

Each part of this study serves other memories: the \emph{timestamp probe} measures how firmly any generator follows its timestamps; the theory, with its deterministic results checked in Lean 4, states when the newest timestamp is right and when moving outdated content beats deleting it; \ours{} applies to any memory that dates its frames, and its log and context also raise state filtering and attention sinks; and the reverted steps of FAR's benchmark, a calibrated state metric and undo statistics of EPIC-KITCHENS-100 test any memory after undos.

\section{Related work}\label{sec:background}

\paragraph{Memory for video world models.}

Memories that retrieve, compress or render past views~\citep{li2025vmem,yu2026worldcrafter,wu2026world} are built for static scenes, where a stored view stays true; \ours{} keeps their retrieval and adds the time over which a view holds, for scenes that the agent changes and changes back.

\paragraph{Memory in changing worlds.}

World models simulate change by the agent, by other actors and out of view~\citep{chen2026out,duan2026liveworld,wang2026worlddirector}. Their memories invalidate, filter or replace a contradicted view~\citep{zheng2026mem,wang2026any,zhang2026matrixgame4}; \ours{} keeps it and re-dates its tokens when the region returns.

\paragraph{Time in the context of a generator.}

Long-rollout rules shift context times, move early frames next to the present as attention sinks, or set positions by role, relevance or novelty~\citep{liu2025rolling,xu2026recap}; \ours{} moves a token only when the agent's actions confirm or contradict its content.

\begin{figure*}[t]
\centering
\includegraphics[width=\linewidth,height=0.82\textheight,keepaspectratio]{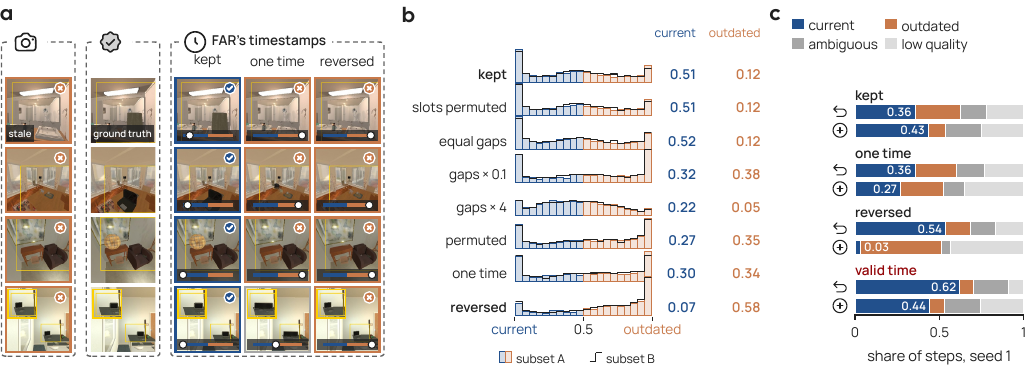}
\caption{FAR with fixed frames, weights and noise: (a) outputs under kept, one-time and reversed timestamps, (b) current and outdated shares under eight timestamp conditions on the 1,478 steps of subset A, and (c) outcomes on reverted and new-state steps at seed 1.\label{fig:timestamps}}
\end{figure*}

\section{Time decides what the generator draws}\label{sec:finding}

We isolate how FAR reads time by holding its 12 context frames, weights and noise fixed and rewriting only the timestamps. At each step a \emph{ground-truth frame} at the target pose shows a region's present state and a \emph{stale frame} stored there shows another. A step is \emph{new-state} when the region holds a state it never had, \emph{repeat-state} when it holds an earlier state, and a repeat-state step is \emph{reverted} when the memory bank holds no frame from that pose captured since the return. Each output shows the \emph{current} state, the \emph{outdated} state or neither, by whether its changed region lies close to the ground-truth or the stale frame, and is \emph{low quality} below a calibrated PSNR floor (\cref{sec:setup}).

At seed 0 on subset A, the 1,478 evaluated steps of the first 102 test clips, on which every rule was chosen, FAR draws the current state on 0.51 and the outdated state on 0.12 under kept timestamps, and 0.07 and 0.58 when reversed (\cref{fig:timestamps}). One shared time gives 0.30 and 0.34, and equal gaps in the true order give 0.52, so order decides which state FAR copies. Tenfold gap compression raises the outdated share to 0.38; fourfold stretching lowers it to 0.05 and sends 0.62 of the outputs below the quality floor, outside the span of timestamps FAR was trained on. On the 352 reverted steps of seed 1, reversal raises the current share from 0.36 to 0.54 and drops it on new-state steps from 0.43 to 0.03 (\cref{fig:timestamps}c): capture order fits new states and fails undos, and no order of capture times fits both.

The order shifts the shares without making either state certain, and generators trained with geometric retrieval shift them less, so a generator reads time as a \emph{soft} last-writer-wins rule. We measure how firmly a generator follows the order by its \emph{timestamp sensitivity} \(s=\tfrac12\big[\log(p_C/p_O)_{\mathrm{kept}}-\log(p_C/p_O)_{\mathrm{reversed}}\big]\), with \(p_C\) and \(p_O\) the current and outdated shares: 1.81 for FAR, 1.21 for the FAR generator under geometric retrieval and 0.63 for the WorldMem baseline~\citep{xiao2025worldmem}, the last two with the frames of the target pose added.

On the 218 reverted steps of seed 1 with an earlier current and a later outdated view at the target pose, we feed HY-WorldPlay~\citep{sun2025worldplay,wu2025hunyuanvideo} both views and exchange their rotary positions. At the fixed threshold 0.2, placing the outdated view later gives 71 outdated and 1 current render; exchanging them gives 70 current and 5 outdated, and one shared position gives either state on only 16 steps, since without an order the two views leave open which state holds (\cref{lem:timestamp} in \cref{sec:proofs}). On 31 undo events in real homes~\citep{sigurdsson2016hollywood}, the same exchange moves the renders toward the frame after the closing.

\section{Valid-time memory}\label{sec:memory}

\subsection{Task and setting}

In a scene of regions, with time running over the natural numbers, region \(x\) has \emph{state} \(w(x,t)\) at time \(t\), such as open or closed or the items on a surface. An \emph{observation} \((x,t,v)\) records \(v=w(x,t)\), a frame holds observations from one time, and the memory reads the action log \(L\) of the agent's actions with their regions and times.

\begin{assumption}[Only the agent changes the scene]\label{asm:agent} A region's state changes only when the agent acts on it.\end{assumption}

\begin{assumption}[The action log reveals reversions]\label{asm:log} A function \(\mathrm{same}_L(x,t,q)\) of \(L\) alone equals \(1\) exactly when \(w(x,t)=w(x,q)\), for all \(x\), \(t\le q\), and worlds consistent with \(L\) and \cref{asm:agent}.\end{assumption}

\Cref{asm:log} holds whenever the actions on a region act freely on its states as a group, since the region has returned to the state of time \(t\) exactly when the product of the actions on it since \(t\) is the identity (\cref{lem:log}). Opening and closing a door generate the group of order two, and putting an item on a receptacle or taking it away toggles that item's membership, so the action log of AI2-THOR meets \cref{asm:log} for doors and receptacles, and the errors of our state tracker come from locating regions in frames. The spot where an item stood falls outside this form, since the simulator returns items to new spots (\cref{sec:tracker}). A memory keeps observations and gives each a timestamp \(\tau\), a function of the observation, \(L\), and the query time \(q\). At \(q\), a \emph{selection rule} models the generator: it receives the kept observations of \(x\) with their timestamps in no order and selects one, \emph{correctly} when \(v=w(x,q)\). The \emph{latest-timestamp rule}, last-writer-wins, selects one with the latest \(\tau\). An observation is \emph{current} when \(v=w(x,q)\) and \emph{outdated} otherwise. Region \(x\) is \emph{up to date} when its latest observation is current, \emph{reverted} when it has a current observation and an outdated latest one, and \emph{unobserved} when it has no current observation, where no memory is correct.

\subsection{What to keep}

A \emph{deletion-based} memory~\citep{yang2026permavid,gu2026worldcraft} removes an observation of \(x\) once a later observation of \(x\) shows another state.

\begin{theorem}[Keep]\label{thm:keep} On a reverted region a deletion-based memory holds no current observation, so every selection rule is wrong; keeping every observation holds a current observation of every reverted region.\end{theorem}

Deleting the doorway view of the empty table removes the only picture of the restored state, so \ours{} keeps every frame. Without timestamps no rule can tell which of two states came last, and capture timestamps make the latest-timestamp rule correct on up-to-date regions (\cref{lem:timestamp}).

\subsection{Which time}\label{sec:validtime}

Let \(e_x\) be the time of the last action at or before \(q\) changing \(x\). \ours{} \emph{re-dates} \((x,t,v)\) with \(t<e_x\) and \(\mathrm{same}_L(x,t,q)=1\) to \(e_x\), and leaves other observations at \(t\); in \cref{fig:teaser} this moves frame A to the pickup of the pillow.

\begin{theorem}[Valid time]\label{thm:valid} (i) Under \ours{}'s timestamps the latest-timestamp rule is correct on every up-to-date and reverted region. (ii) For any two states, two worlds share one history with \(x\) up to date in one and reverted in the other; a memory ignoring \(L\) is correct in at most one under a deterministic rule, and with probabilities summing to at most \(1\) under a randomized rule. (iii) Timestamps make the latest-timestamp rule correct on every up-to-date and reverted region of every history exactly when they place every current observation of \(x\) after every outdated one. (iv) Among assignments keeping outdated observations at capture time and never moving observations earlier, every one satisfying (iii) places each current observation with \(t<e_x\) at \(e_x\) or later, with equality under \ours{}.\end{theorem}

By (ii), repair without a post-return view needs \(L\); by (iii) and (iv), valid time is the earliest timestamp ordering current content after outdated content. Strictly increasing maps of capture times keep capture order's selections, and log-free re-timings fail on one of the two histories (\cref{lem:retime}).

\begin{corollary}[No return in the input]\label{cor:noreturn} Under the timestamps of \ours{}, the entries of every up-to-date or reverted region are, under capture timestamps, the entries of a history of a world consistent with \cref{asm:agent} in which the region is up to date and never held its current state before \(e_x\).\end{corollary}

Under these timestamps a reverted region reads as one that has just reached a new state, the case in which the latest-timestamp reading is right, so a generator that learned this reading needs no training to use them.

A generator also draws from its own prediction, which lacks states held only in memory. The \emph{soft rule}, the conditional logit of discrete choice~\citep{mcfadden1974conditional} with the generator's own prediction as its outside option, selects a kept observation \(o\) of \(x\) with probability \[P(o)=\frac{e^{\beta\tau(o)}h(o)}{\sum_{o'\text{ kept of }x}e^{\beta\tau(o')}h(o')+h_0e^{\beta\tau_q}},\] and draws its own prediction with the remaining probability, for sharpness \(\beta>0\), pose-content weight \(h(o)\in[h_{\min},h_{\max}]\) with \(h_{\min}>0\), prediction weight \(h_0>0\), and generated-frame time \(\tau_q\), converging to the latest-timestamp rule on memory draws as \(\beta\to\infty\). Under a constant hazard \(\beta\) of change and no return, a view captured at \(\tau\) survives to \(q\) unchanged with probability \(e^{-\beta(q-\tau)}\), which is proportional to the soft rule's weight \(e^{\beta\tau}\) with sharpness equal to the hazard. \Cref{sec:finding} shows generators that learned this reading; an undo breaks the premise because a restoring action confirms older content again at \(e_x\), and valid time restarts the clock there.

\begin{proposition}[Soft rule]\label{prop:soft} Under the soft rule, (i) the probability of a correct selection rises with every current timestamp and falls with every outdated one, (ii) if every current timestamp exceeds every outdated one by \(\Delta\), an outdated observation is selected with probability at most \(\frac{n_Oh_{\max}}{n_Ch_{\min}}e^{-\beta\Delta}\) for \(n_C\) current and \(n_O\) outdated observations, vanishing as \(\beta\to\infty\), and (iii) on a reverted region, deleting outdated observations under capture time is correct with probability \(A/(A+h_0e^{\beta\tau_q})\), which \ours{} exceeds whenever its outdated weight is below \(h_0e^{\beta\tau_q}(A'/A-1)\), for current weights \(A\) under capture time and \(A'>A\) under valid time.\end{proposition}

Deletion removes the weight of outdated content. Re-dating removes it as well, raises the weight of current content against the generator's own prediction, and keeps the outdated tokens in the context, a gain in image quality that the selection model leaves out and \cref{sec:ablation} measures.

\subsection{Frames and views}

A frame is \emph{consistent} when no region of the target view is outdated in it~\citep{yang2026oneira}. \emph{Filtering} keeps only consistent frames, losing current observations of other regions in withheld frames, and \ours{} keeps inconsistent frames at capture time.

\begin{theorem}[Re-date, keep the frame]\label{thm:frame} For region \(z\) of the target view with a current observation, (i) with one timestamp per frame, filtering is correct on \(z\) exactly when a current observation of \(z\) lies in a consistent frame, and \ours{} exactly when this holds or \(z\) is up to date; (ii) if two frames cross, one showing \(y\) current and \(z\) outdated and the other \(y\) outdated and \(z\) current, with no other frame showing \(y\) or \(z\), every per-frame assignment is wrong on \(y\) or \(z\) and filtering is wrong on both; (iii) with a timestamp per token, \ours{} is correct on every up-to-date and reverted region of the target view.\end{theorem}

When no frame from the target pose shows the present, the view decides. Let \(N(g)\) be the poses near target pose \(g\).

\begin{assumption}[The generator copies from the target view]\label{asm:view} For target pose \(g\), the selection rule selects among observations with poses in \(N(g)\).\end{assumption}

\begin{theorem}[View]\label{thm:view} Under \cref{asm:view}, (i) if all current observations of \(x\) lie outside \(N(g)\), every memory of captured observations is wrong under any timestamps; (ii) \ours{} is correct on every up-to-date and reverted region with a current observation in \(N(g)\); (iii) if a memory adds a current observation in \(N(g)\) to every history, the latest-timestamp rule is correct on every such region of every history exactly when current observations in \(N(g)\) follow outdated ones in \(N(g)\), which \(e_x\) satisfies under \ours{} and any timestamp before an outdated observation in \(N(g)\) violates; (iv) \ours{} is correct wherever any memory of captured observations is correct under any timestamps and selection rule.\end{theorem}

Wherever \ours{} is wrong in this model every memory of captured frames is wrong, and the remaining errors need a view of the current state from \(g\), which no timestamp supplies. \Cref{tab:theory} collects the results, \cref{sec:proofs} proves them, and Lean 4 checks the deterministic ones.

\begin{table}[tb]
\centering
\caption{What the latest-timestamp rule shows on one region. ✓ current, ✗ outdated, ½ current on at most one of two histories with the same frames or crossing regions, and (✓) current when a consistent frame holds a current observation.\label{tab:theory}}
\footnotesize\setlength{\tabcolsep}{5pt}
\begin{tabular}{lcccc}
\toprule
Memory & \begin{tabular}[b]{@{}c@{}}Up to\\date\end{tabular} & Reverted & Crossing & \begin{tabular}[b]{@{}c@{}}Other\\poses\end{tabular} \\
\midrule
Deletion & ✓ & ✗ & ½ & ✗ \\
Shared time or none & ½ & ½ & ½ & ✗ \\
Capture order & ✓ & ✗ & ½ & ✗ \\
Sink at the present & ½ & ½ & ½ & ✗ \\
State filtering & (✓) & (✓) & ✗ & ✗ \\
\ours{} per frame & ✓ & (✓) & ½ & ✗ \\
\rowcolor{oursrow}\textbf{\ours{} (ours)}, per token & ✓ & ✓ & ✓ & ✗ \\
+ frame at target pose & ✓ & ✓ & ✓ & ✓ \\
\bottomrule
\end{tabular}
\end{table}

\subsection{\ours{} in a retrieval memory}\label{sec:redate}

\begin{figure*}[t]
\centering
\includegraphics[width=\linewidth,height=0.82\textheight,keepaspectratio]{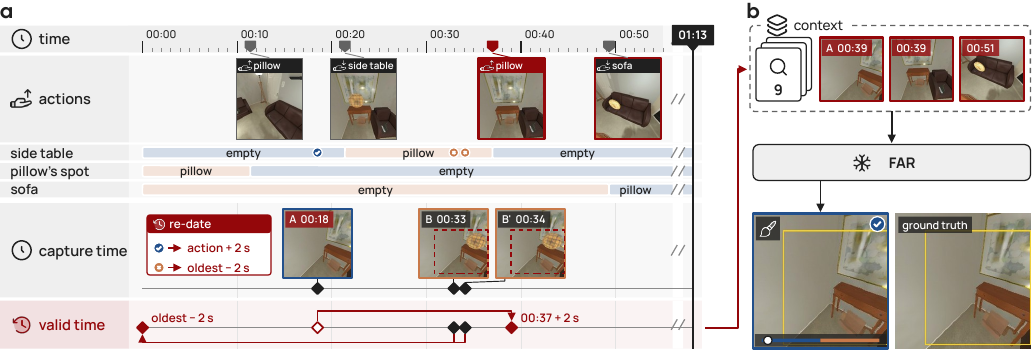}
\caption{(a) The state tracker reads the action log and re-dates the frames of the target pose: a frame whose region is as at capture (tick) moves to 2~s after the restoring action, and the boxed tokens of a changed region (cross) move 2~s before the oldest retrieved frame. (b) The re-dated frame and the views after the last two actions join FAR's context.\label{fig:method}}
\end{figure*}

From \(L\), a \emph{state tracker} follows each \emph{manipulated region}, a door, a receptacle or the spot where an item first stood (\cref{fig:method,sec:implementation}). Its tokens lie in a disc of radius 0.8~m centered 1~m in front of the agent at its actions.

One principle governs the timestamps: every retrieved token is dated by the last time its content was observed or re-established, at its capture or at the action that restored it, and a token whose content the log contradicts is dated before every retrieved frame. Because FAR adds one timestamp embedding to every token of a frame and a frame shows outdated regions beside current ones, \ours{} gives each token \(k\) of a retrieved frame \(o\) with capture time \(\tau(o)\) its own timestamp, as \cref{thm:frame}(iii) requires. A manipulated region of the target view acted on since \(\tau(o)\) is \emph{restored} when it holds the state in \(o\), and \emph{outdated} otherwise. Token \(k\) receives \[\tau'(k)=\begin{cases}\tau_{\min}-2\,\mathrm{s}&k\text{ outdated,}\\a(o)+2\,\mathrm{s}&k\text{ not outdated, }o\text{ restored,}\\\tau(o)&\text{otherwise,}\end{cases}\] with \(o\) restored when it shows a restored region, \(a(o)\) the last action that restored one, and \(\tau_{\min}\) the oldest timestamp of the context. The restored date meets \cref{thm:valid}(iii), and the outdated date widens the gap \(\Delta\) of \cref{prop:soft}(ii) and stays just outside the span the retrieved frames cover.

FAR's retriever ranks by capture time, so \ours{} recalls the two re-dated frames nearest \(g\) and the frames 2~s after the last two actions (\emph{post-action retrieval}) into its 12 context frames, replacing FAR's lowest-ranked ones. \ours{} changes no weight, no sampler setting and no training; it costs one table update per action and one embedding difference per token.

\section{Experiments}\label{sec:experiments}

\subsection{Setup}\label{sec:setup}

Every memory runs without training in FAR's frozen diffusion transformer~\citep{kim2026learning,peebles2023scalable,bar2025navigation}. Its benchmark holds 1,309 test clips of AI2-THOR rooms~\citep{kolve2017ai2}, tours in which an agent picks up, puts down and opens objects and later returns to them, with action logs and simulator change masks. FAR retrieves 12 frames from a bank of 1,000 and samples with 20 DDIM steps~\citep{song2021denoising}, sharing noise across memories within a seed. At seed 1 the 249 clips with a repeat-state step hold 352 reverted, 1,050 repeat-state and 2,968 new-state steps; 52 of these clips lie in the first 200, on which every rule was chosen or first run, and the other 197 are \emph{held-out clips}.

The benchmark's LPIPS test calls a changed-region crop correct when it lies nearer the ground-truth frame than the stale frame, accepting 0.31 of frames from a memory that never saw the current state, so we classify each frame by its state~\citep{tiwari2026video}. With latent errors \(\delta_t\) and \(\delta_s\) to the ground-truth and stale frames and error \(d\) between them, a frame is current when \(\delta_t<\delta_s\) and \(\delta_t\le\kappa d\), outdated when \(\delta_s<\delta_t\) and \(\delta_s\le\kappa d\), and \emph{ambiguous} otherwise. We calibrate \(\kappa\) on degraded simulator frames of the current state; below 30.4~dB, where calibration no longer separates the states, a frame is low quality (\cref{sec:evaluation}). \emph{Accuracy} is the share of steps that show the current state; it accepts 0.95 of degraded frames of the current state and 0.01 of the frames of a memory that never saw it. Because it calls 0.43 of even pixel mixtures of the two frames current, we repeat every comparison at the \emph{fixed threshold} \(\kappa=0.2\) with no floor, which calls 0.19 of them current. We compare FAR with 13 rules adapted from 10 published memories (\cref{tab:main}); each rule decides what to delete, filter or re-time inside FAR's memory bank and frozen generator, isolating the rule from its source system's training. State filtering reads \ours{}'s state tracker~\citep{yang2026oneira}, and four rules (†) read the simulator.

\begin{table*}[t]
\centering
\caption{Memory rules inside FAR at seed 1: published memories; \ours{}'s parts, published rules given \ours{}'s log or context, and \ours{}; and rules that read the simulator (†). Acc. is accuracy, At 0.2 the fixed-threshold current share, Seen and Unseen split steps by whether the target pose stored the current state, Outd. and Low q. are outdated and low-quality shares, and Repeat and New give accuracy on all steps of each kind. ‡ reads the action log; bold and underline mark the best and second best without †.\label{tab:main}}
\footnotesize\setlength{\tabcolsep}{4pt}
\begin{tabular}{lrrrrrrrrrr}
\toprule
 & \multicolumn{8}{c}{Reverted steps} & \multicolumn{2}{c}{All steps} \\
\cmidrule(lr){2-9}\cmidrule(lr){10-11}
Memory rule & Acc.↑ & At 0.2↑ & Seen↑ & Unseen↑ & Outd.↓ & Low q.↓ & LPIPS acc.↑ & PSNR↑ & Repeat↑ & New↑ \\
\midrule
FAR~\citep{kim2026learning} & 0.36 & 0.32 & 0.54 & 0.05 & 0.27 & 0.22 & 0.62 & 36.5 & 0.58 & 0.43 \\
View-overlap retrieval~\citep{yu2025context} & 0.08 & 0.08 & 0.12 & 0.01 & 0.03 & 0.88 & 0.14 & 22.5 & 0.09 & 0.01 \\
Geometric retrieval~\citep{xiao2025worldmem} & 0.03 & 0.01 & 0.03 & 0.02 & 0.09 & 0.82 & 0.20 & 23.3 & 0.26 & 0.23 \\
Decay by age~\citep{zheng2026mem} & 0.40 & 0.37 & 0.61 & 0.06 & 0.24 & 0.19 & 0.64 & 37.5 & 0.60 & \underline{0.44} \\
Re-timing in the true order~\citep{sun2025worldplay} & 0.26 & 0.23 & 0.39 & 0.06 & 0.36 & 0.24 & 0.53 & 35.7 & 0.53 & \underline{0.44} \\
Sink at the present~\citep{yi2025deep} & 0.52 & 0.49 & 0.80 & 0.05 & 0.19 & 0.15 & 0.73 & 39.1 & 0.68 & 0.21 \\
Frame deletion by pose~\citep{wu2026echo} & 0.11 & 0.03 & 0.15 & 0.05 & 0.26 & 0.19 & 0.64 & 34.2 & 0.51 & \underline{0.44} \\
Token deletion by later frame~\citep{wu2026echo} & 0.14 & 0.05 & 0.20 & 0.05 & 0.28 & 0.21 & 0.61 & 34.1 & 0.50 & \underline{0.44} \\
Deletion of all earlier frames‡~\citep{gu2026worldcraft} & 0.02 & 0.01 & 0.03 & 0.02 & \textbf{0.00} & 0.95 & 0.22 & 21.1 & 0.20 & 0.25 \\
State filtering‡~\citep{yang2026oneira} & 0.50 & 0.45 & 0.76 & 0.07 & \underline{0.01} & 0.39 & 0.71 & 33.7 & 0.69 & 0.42 \\
\midrule
FAR + post-action retrieval‡ & 0.41 & 0.38 & 0.63 & 0.05 & 0.24 & 0.18 & 0.66 & 37.5 & 0.66 & 0.43 \\
FAR + valid time‡ & 0.55 & 0.51 & 0.84 & 0.08 & 0.19 & \underline{0.11} & 0.76 & 39.2 & 0.74 & \textbf{0.46} \\
Token deletion, \ours{}'s log, capture time‡ & 0.52 & 0.48 & 0.81 & 0.05 & \underline{0.01} & 0.39 & 0.70 & 34.0 & 0.73 & 0.40 \\
Token deletion, \ours{}'s log, valid time‡ & 0.56 & 0.53 & 0.87 & 0.06 & \underline{0.01} & 0.36 & 0.70 & 35.0 & 0.72 & 0.41 \\
State filtering, \ours{}'s context‡~\citep{yang2026oneira} & 0.53 & 0.49 & 0.80 & 0.08 & \underline{0.01} & 0.38 & 0.73 & 34.0 & 0.74 & 0.42 \\
Sink at the present, \ours{}'s context‡~\citep{yi2025deep} & 0.57 & 0.52 & 0.87 & 0.08 & 0.14 & 0.12 & 0.79 & \textbf{39.8} & 0.74 & 0.21 \\
\rowcolor{oursrow}\textbf{\ours{} (ours)}‡ & \textbf{0.62} & \textbf{0.55} & \textbf{0.94} & \textbf{0.10} & 0.09 & \textbf{0.09} & \textbf{0.85} & \underline{39.7} & \textbf{0.81} & \underline{0.44} \\
\ours{}, log without object names‡ & \underline{0.61} & \textbf{0.55} & \underline{0.92} & \textbf{0.10} & 0.09 & \underline{0.11} & \underline{0.84} & 39.5 & \underline{0.80} & \underline{0.44} \\
\midrule
Frame deletion at each change†~\citep{yang2026permavid} & 0.05 & 0.02 & 0.06 & 0.05 & 0.01 & 0.82 & 0.41 & 26.9 & 0.41 & 0.42 \\
Token deletion at each change†~\citep{yang2026permavid} & 0.10 & 0.01 & 0.11 & 0.07 & 0.01 & 0.45 & 0.56 & 30.9 & 0.46 & 0.42 \\
Oracle token deletion†~\citep{yang2026permavid} & 0.43 & 0.38 & 0.64 & 0.08 & 0.10 & 0.24 & 0.67 & 36.7 & 0.61 & 0.43 \\
Changing content never stored†~\citep{wang2026latent} & 0.09 & 0.01 & 0.11 & 0.05 & 0.01 & 0.47 & 0.51 & 30.8 & 0.16 & 0.10 \\
\bottomrule
\end{tabular}
\end{table*}

\subsection{Reverted steps}

\begin{figure*}[t]
\centering
\includegraphics[width=\linewidth,height=0.82\textheight,keepaspectratio]{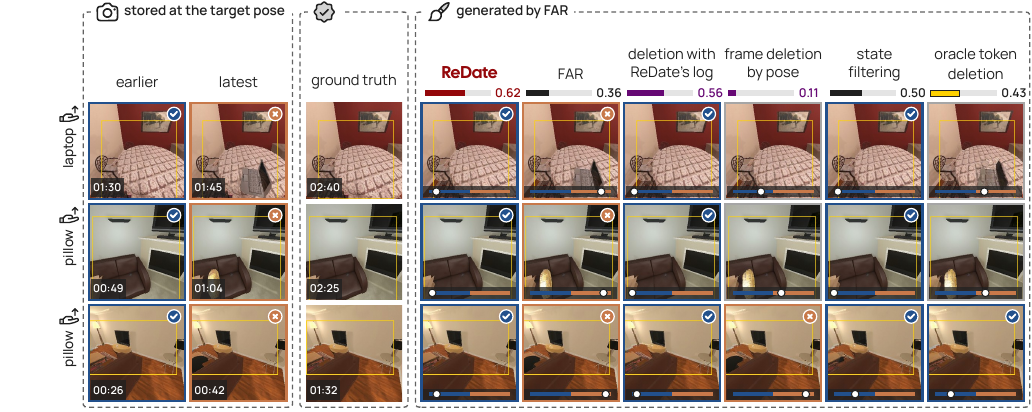}
\caption{Reverted steps at seed 1, with accuracy on all 352 steps (yellow reads the simulator). FAR draws the outdated state, frame deletion by pose leaves a ghost, and \ours{} draws the current state.\label{fig:examples}}
\end{figure*}

\begin{table*}[t]
\centering
\caption{Re-dating against token deletion with the same log, frames and tokens on the 352 reverted steps of seed 1; columns and marks as in \cref{tab:main}.\label{tab:samelog}}
\footnotesize\setlength{\tabcolsep}{5pt}
\begin{tabular}{llrrrrrrr}
\toprule
Recalled frames & Outdated tokens & Acc.↑ & \begin{tabular}[b]{@{}r@{}}At\\0.2↑\end{tabular} & \begin{tabular}[b]{@{}r@{}}LPIPS\\acc.↑\end{tabular} & PSNR↑ & Outd.↓ & \begin{tabular}[b]{@{}r@{}}Low\\q.↓\end{tabular} & New↑ \\
\midrule
capture time & kept & 0.39 & 0.37 & 0.65 & 37.3 & 0.25 & 0.19 & 0.44 \\
capture time & hidden & 0.52 & 0.48 & 0.70 & 34.0 & \textbf{0.01} & 0.39 & 0.40 \\
capture time & moved before the oldest frame & \underline{0.58} & 0.51 & \underline{0.82} & 39.4 & 0.10 & 0.12 & \textbf{0.45} \\
valid time & kept & \underline{0.58} & 0.53 & 0.80 & \underline{39.6} & 0.15 & \underline{0.10} & \textbf{0.45} \\
valid time & hidden & 0.56 & 0.53 & 0.70 & 35.0 & \textbf{0.01} & 0.36 & 0.41 \\
\rowcolor{oursrow}\textbf{valid time, \ours{} (ours)} & moved before the oldest frame & \textbf{0.62} & \textbf{0.55} & \textbf{0.85} & \textbf{39.7} & 0.09 & \textbf{0.09} & 0.44 \\
valid time, control & moved, and as many tokens hidden half the grid away & \underline{0.58} & \textbf{0.55} & 0.74 & 34.9 & \textbf{0.01} & 0.36 & 0.40 \\
\bottomrule
\end{tabular}
\end{table*}

On the 352 reverted steps \ours{} raises accuracy from FAR's 0.36 to 0.62 and cuts the outdated share from 0.27 to 0.09, higher on 54 of 141 clips and lower on 2 (\cref{fig:examples}), and the order holds in LPIPS features and in pixels.

Each baseline family behaves as \cref{tab:theory} predicts. Deletion rules, built for changes that last, remove every frame of the current state at the target pose and reach 0.02 to 0.14 (\cref{thm:keep}). Re-timing in the true order keeps capture order and drops to 0.26, since the soft rule also reads the gaps it alters. Deep Forcing's sink, which moves the two oldest retrieved frames next to the present whatever they show, reaches 0.52, or 0.57 with \ours{}'s context, and drops new-state steps from 0.43 to 0.21; one of the two histories of \cref{thm:valid}(ii) loses what the other gains. Oracle token deletion reaches 0.43 with 0.24 at low quality, as its current tokens keep their capture timestamps and the generator draws more of the region from its own prediction (\cref{prop:soft}(iii)). On the 219 \emph{seen} steps, whose target pose stored the current state, \ours{} reaches 0.94 against 0.54 for FAR and 0.76 for state filtering, which withholds whole frames and leaves 0.39 of its renders at low quality (\cref{thm:frame}). On seen steps FAR draws new states on 0.88 and returns on only 0.54, and \ours{} draws returns on 0.94 and new states on 0.90: a re-dated return is drawn as well as a new state (\cref{cor:noreturn}). On the 133 \emph{unseen} steps, whose target pose did not store the current state, all memories stay at or below 0.10, near the zero of \cref{thm:view}(i).

\subsection{Ablation}\label{sec:ablation}

Adding \ours{}'s parts to FAR one at a time, valid time on FAR's own retrieval raises accuracy on reverted steps from 0.36 to 0.55, the two recalled frames raise it to 0.57, post-action retrieval to 0.58, which we call \emph{\ours{} per frame}, and a timestamp per token to 0.62. The timestamp per token nearly halves the outdated share, from 0.15 to 0.09 at seed 1 and from 0.13 to 0.06 at seed 0, since it also moves the outdated tokens of frames that keep their capture time. The recalled frames act through their timestamps: under capture time or one shared time they give 0.39 and 0.21, and the same new timestamps on as many random frames give 0.30.

\Cref{tab:samelog} crosses capture or valid time on recalled frames with keeping, hiding, or moving outdated tokens 2~s before the oldest frame; valid time with the tokens kept, re-dated per token, matches \ours{} per frame at 0.58. Against moving them, hiding the tokens leaves a hole that costs over 4~dB of PSNR and sends over a third of the renders to low quality; moving them into the past keeps the image whole and gives the highest accuracy and the fewest low-quality renders, with the same order at seed 0. A hole clears outdated renders by degrading them: a control that keeps \ours{}'s moved tokens and hides as many tokens half the grid away agrees with token deletion on 324 of 352 steps and sends 24 of \ours{}'s 30 outdated renders below the quality floor. Valid time raises accuracy under every treatment (\cref{prop:soft}(iii)), and varying constants moves it by at most 0.01.

\subsection{Held-out clips, the benchmark's evaluator and a second generator}\label{sec:robustness}

Across seeds 0, 1 and 2 (366, 352 and 50 reverted steps), \ours{} reaches 0.64, 0.62 and 0.48 against FAR's 0.38, 0.36 and 0.32, higher at seed 1 in 15 of the 20 rooms and lower in 1, and 0.58 against 0.34 on the 590 reverted steps of the 197 held-out clips across seeds 0 and 1. On the complete index at seed 0, it raises the 1,050 repeat-state steps from 0.54 to 0.80 and all 18,006 steps from 0.49 to 0.52. Under FAR's own evaluator, whose reruns of FAR and the WorldMem baseline match their reported values within 0.003, \ours{} raises accuracy from 0.802 to 0.824 on surfaces and from 0.760 to 0.817 in containers, ahead of post-action retrieval alone.

In HY-WorldPlay, the views of the target pose in \ours{}'s context take the rotary positions of their dates, and no ground-truth frame or change mask enters the run. At the fixed threshold, \ours{} raises the current share from 0.003 under capture time to 0.188 at seed 1, 0.208 at seed 0 and 0.181 on held-out clips, cutting the outdated share from 0.281 to 0.091. Given \ours{}'s log, token deletion and state filtering reach 0.233 and 0.173 at 25.7~dB and 24.0~dB of PSNR, against \ours{}'s 0.188 at 28.0~dB: the log carries the gain, and re-dating keeps the image. Under geometric retrieval with target-pose and post-action frames, \ours{} raises the FAR generator from 0.38 to 0.60 and the WorldMem baseline from 0.40 to 0.43, so each generator gains in the order of its timestamp sensitivity. With every generated frame written back into memory, \ours{} gives 0.61 and FAR 0.35. A rollout step takes 0.47~s with \ours{} and 0.44~s with FAR at the same 6.1~GiB peak memory.

\subsection{Views and the action log}\label{sec:analysis}

\ours{}'s remaining errors lie where the target pose never stored the present: 120 of its 134 misses on reverted steps fall in the 133 unseen steps, and the 2,423 new-state steps whose split is recorded divide the same way, 0.90 on the 1,199 whose target pose stored the new state and 0.06 on the 1,224 seen only from other poses. A simulator frame of the present stored at the target pose under its valid time raises \ours{} per frame on unseen reverted steps from 0.07 to 0.76 (0.84 on such new-state steps), and timestamped before the outdated frames it gives 0.53 and 0.48, as \cref{thm:view}(iii) states. The simulator's changed tokens alone, written into the newest target-pose frame, give 0.73, so the remaining errors lie in carrying the present to the target pose, where a relay through FAR's rollout and VGGT depth reprojection~\citep{wang2025vggt} reach 0.13 and 0.10.

At the target pose, 0.88 of frames marked outdated disagree with the ground-truth frame, and the state tracker re-dates target-pose frames with precision 0.62 and recall 0.56 (\cref{sec:tracker}). A log with only the verb and place of each action lowers precision and recall to 0.54 and 0.47 and leaves \ours{} at 0.61; dropping or misattributing up to half of the actions leaves \ours{} per frame between 0.44 and 0.50, above FAR's 0.36. Valid time also tells a memory what to keep: on the 222 reverted steps whose context holds a re-dated view, a memory of the last 60~s of capture keeps that view on 0.28 of the steps, and a memory of the last 60~s of valid time on 0.82.

\section{Conclusion}

Given an older and a newer view of a region, a frozen video world model draws the one with the newer timestamp, whether time enters through embeddings or rotary positions. After an undo, deletion and state filtering drop the view of the present, and capture time and re-timing without the log keep it older than the outdated view. Dating retrieved tokens by their valid time lets the frozen generator draw the present without training, and moving outdated tokens keeps frames whole where deletion leaves a hole. The remaining errors lie where only other poses saw the present; rendering a valid-time view of it at the target pose and logging what other actors change are the next problems.

\label{cvpr:body-end}

\section*{LLM Usage Statement}
\addcontentsline{toc}{section}{LLM Usage Statement}

We used Claude through Claude Code (Anthropic), Codex (OpenAI) and Gemini (Google) mainly to assist our research and its execution: programming and running the experiments, searching for related work and checking each reference against its source, and editing the language and grammar of our text. The authors conceived the theory and the method, divided the work and wrote the draft of the paper, and they checked every statement, number, figure and citation.
{\small
\bibliographystyle{ieeenat_fullname}
\bibliography{arxiv-references}
}
\renewcommand{\topfraction}{0.9}\renewcommand{\dbltopfraction}{0.9}\renewcommand{\bottomfraction}{0.8}
\renewcommand{\textfraction}{0.07}\renewcommand{\floatpagefraction}{0.9}\renewcommand{\dblfloatpagefraction}{0.9}
\setcounter{topnumber}{4}\setcounter{dbltopnumber}{4}\setcounter{bottomnumber}{2}\setcounter{totalnumber}{6}
\makeatletter
\setlength{\@fptop}{0pt}\setlength{\@fpbot}{0pt plus 1fil}\setlength{\@dblfptop}{0pt}\setlength{\@dblfpbot}{0pt plus 1fil}
\setlength{\@fpsep}{\floatsep}\setlength{\@dblfpsep}{\dblfloatsep}
\makeatother
\appendix
\crefalias{section}{appendix}\crefalias{subsection}{appendix}
\section{Proofs}\label{sec:proofs}
\suppressfloats[t]

\begin{figure*}[tp]
\centering
\includegraphics[width=\linewidth,height=0.82\textheight,keepaspectratio]{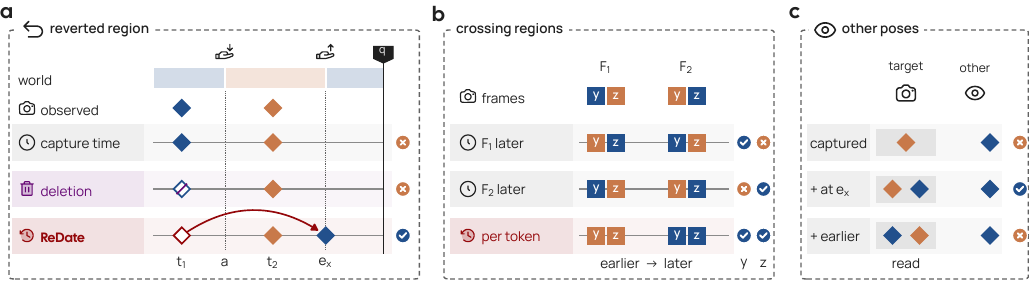}
\caption{The three cases in the proofs; each badge gives the state of the entry with the latest timestamp. (a) On a reverted region capture time selects the outdated observation, deletion removes the current one, and \ours{} moves the current one to \(e_x\). (b) On crossing regions one timestamp per frame is wrong on \(y\) or \(z\), and a timestamp per token is right on both. (c) When only another pose saw the current state, an observation added at the target pose is selected at \(e_x\) and missed before the outdated one.\label{fig:theory}}
\end{figure*}

This appendix states the model of \cref{sec:memory} in full and proves its results in order, with \cref{lem:log}, on which the state tracker rests, and \cref{lem:timestamp,lem:retime}, which the main paper uses in passing. \Cref{fig:theory} illustrates the three cases behind the proofs, \cref{tab:s-notation} collects the notation, and \cref{tab:s-results} lists each result with its assumptions, the main-paper measurement that tests it, and its verification in Lean 4 in \texttt{ReDate.lean} of the supplementary code, which checks the deterministic results.

\begin{table}[tbp]
\centering
\caption{Notation of the proofs.\label{tab:s-notation}}
\footnotesize\setlength{\tabcolsep}{6pt}
\begin{tabular}{ll}
\toprule
Symbol & Meaning \\
\midrule
\(X\), \(V\) & regions and their states \\
\(w(x,t)\) & the state of region \(x\) at time \(t\) \\
\(L\) & the action log \\
\((x,t,v)\) & an observation of \(x\) at \(t\), \(v=w(x,t)\) \\
\(H\), \(q\) & a history and the query time \\
\(p=w(x,q)\) & the current state of \(x\) \\
\(\mathrm{same}_L(x,t,q)\) & one exactly when \(w(x,t)=w(x,q)\) \\
\(e_x\) & the last action at or before \(q\) changing \(x\) \\
\(d_{L,q}\) & a timestamp assignment \\
\(F\), \(t_F\), \(T\) & a frame, its capture time, the target view \\
\(g\), \(N(g)\) & the target pose and the poses near it \\
\(\tau\), \(\beta\), \(h\), \(h_0\) & a timestamp; the soft rule's sharpness and weights \\
\(\ell\) & the log-odds, current against outdated \\
\bottomrule
\end{tabular}
\end{table}

\begin{table*}[tp]
\centering
\caption{The proved results, their assumptions, the main-paper measurements that test them, and the scope of the Lean 4 verification.\label{tab:s-results}}
\small\setlength{\tabcolsep}{4pt}
\begin{tabularx}{\linewidth}{>{\raggedright\arraybackslash\hsize=0.697\hsize}Xl>{\raggedright\arraybackslash\hsize=1.714\hsize}X>{\raggedright\arraybackslash\hsize=0.589\hsize}X}
\toprule
Result & Assumes & Tested by & Lean \\
\midrule
\Cref{lem:log}, group actions reveal reversions & \Cref{asm:agent} & the state tracker on receptacles and doors (\cref{sec:tracker}) &  \\
\Cref{thm:keep}, keep & none & deletion-based memories at 0.02 to 0.14 on reverted steps & ✓ \\
\Cref{lem:timestamp}, timestamp & \Cref{asm:agent} & one time for every frame, current state from 0.51 to 0.30; HY-WorldPlay with both views at one position, either state on 16 of 218 steps & ✓ \\
\Cref{thm:valid}, valid time & \Cref{asm:agent,asm:log} & \ours{} at 0.94 where the target pose stored the current state; reversed capture timestamps trade new-state steps for reverted ones; HY-WorldPlay from 1 to 70 current renders as two positions exchange & ✓ \\
\Cref{cor:noreturn}, no return in the input & \Cref{asm:agent,asm:log} & \ours{} at 0.94 on seen reverted steps against FAR's 0.88 on seen new-state steps & ✓ \\
\Cref{lem:retime}, re-timing & \Cref{asm:agent} & re-timing in the true order at 0.26; the sink at 0.52 with new-state steps at 0.21 & first part \\
\Cref{prop:soft}, soft rule & \Cref{asm:agent,asm:log} & oracle token deletion with 0.24 low quality; restored tokens at capture raise it from 0.09 to 0.12; deletion with \ours{}'s log at 0.52 under capture time and 0.56 under valid time, and 0.56 and 0.59 at seed 0 & three inequalities, integer weights \\
\Cref{cor:sensitivity}, sensitivity & none & \ours{}'s gain in log-odds in the order of the timestamp sensitivity, at 0.85 to 1.13 times it, for FAR, the FAR generator under geometric retrieval and the WorldMem baseline (\cref{sec:robustness}) &  \\
\Cref{thm:frame}, frames & \Cref{asm:agent,asm:log} & per-token timestamps at 0.62 against 0.58 per frame; filtering with 0.39 low quality & (ii), (iii) \\
\Cref{thm:view}, view & \Cref{asm:agent,asm:log,asm:view} & every memory at or below 0.10 where only other poses saw the current state; a stored ground-truth frame at 0.76 under valid time and 0.53 before & deterministic rules \\
\Cref{cor:table}, \cref{tab:theory} & \Cref{asm:agent,asm:log,asm:view} & the rows of \cref{tab:main} &  \\
\bottomrule
\end{tabularx}
\end{table*}

\subsection{The model}

Let \(X\) be a set of regions, let time run over the natural numbers, and let \(V\) be a set of states. A \emph{world} is a function \(w\) from regions and times to states. The agent acts at finitely many times, each action touches one region, and the \emph{action log} \(L\) lists the actions with their regions and times.

\renewcommand{\restatedname}{Assumption~\ref*{asm:agent}}\begin{restateddefinition}[Only the agent changes the scene] For every region \(x\) and time \(t\ge1\), if no action touches \(x\) at time \(t\) then \(w(x,t)=w(x,t-1)\).\end{restateddefinition}

A world is \emph{admissible} for a log \(L\) when it satisfies \cref{asm:agent} and each logged action applies the change that the log names.

\renewcommand{\restatedname}{Assumption~\ref*{asm:log}}\begin{restateddefinition}[The action log reveals reversions] There is a function \(\mathrm{same}_L(x,t,q)\) of the log alone that equals one exactly when \(w(x,t)=w(x,q)\), for every admissible world \(w\) of \(L\), every region \(x\) and all times \(t\le q\).\end{restateddefinition}

\Cref{lem:log} gives a condition on the actions under which a state tracker computes \(\mathrm{same}_L\).

An \emph{observation} is a triple \((x,t,w(x,t))\), and a \emph{history} \(H\) at a query time \(q\) is a finite set of observations with \(t\le q\). A \emph{memory} maps the history, the action log and the query time to a finite set of \emph{entries}, each an observation of the history with a timestamp. A \emph{timestamp assignment} is a function \(d_{L,q}(x,t,v)\) of an observation, the log and the query time. A \emph{selection rule} receives the entries about a region \(x\) as an unordered set, reads of each entry its state, its timestamp and its pose, and maps them to one of them, possibly at random, so that the capture time of an entry reaches the rule only through its timestamp; it \emph{is correct} when the selected entry has the state \(w(x,q)\). A selection rule with no entry to select is wrong. The \emph{latest-timestamp rule} selects an entry about \(x\) with the latest timestamp, and it is correct when every entry with the latest timestamp has the state \(w(x,q)\).

Fix a region \(x\) and write \(p=w(x,q)\) for its current state. The observations of \(x\) in \(H\) with state \(p\) are its \emph{current observations} and the others its \emph{outdated observations}. The region is \emph{up to date} when its latest observation is current, \emph{reverted} when it has a current observation and its latest observation is outdated, and \emph{unobserved} when it has no current observation. Every region with an observation is of one of these three kinds. Memory entries are observations of the history, so on an unobserved region no entry has state \(p\) and every selection rule of every memory is wrong.

Four memories appear in the theorems (\cref{fig:theory}a). The \emph{deletion-based} memory keeps an observation \((x,t,v)\) exactly when no observation \((x,t',v')\) of the history has \(t'>t\) and \(v'\ne v\), under any timestamps. The memory with \emph{capture timestamps} keeps every observation under its capture timestamp \(t\). \emph{\ours{}} keeps every observation. Let \(e_x\) be the time of the last action at or before \(q\) that changed the state of \(x\), with \(e_x=-\infty\) when no action did, so that \(x\) has held the state \(p\) at every time from \(e_x\) to \(q\). \ours{} writes on an observation \((x,t,v)\) the timestamp \(e_x\) when \(t<e_x\) and \(\mathrm{same}_L(x,t,q)=1\), and the timestamp \(t\) otherwise.

A \emph{frame} is a set of observations of distinct regions with one time of capture \(t_F\), and a memory of frames hands the selection rule whole frames under one timestamp each, so that the entries about \(x\) are the observations of \(x\) in the frames handed over. A query has a \emph{target view} \(T\subseteq X\), and a frame \(F\) is \emph{consistent} when \(w(s,t_F)=w(s,q)\) for every \(s\in F\cap T\). The \emph{filtering} memory hands the selection rule the consistent frames under their capture timestamps, deciding consistency by \(\mathrm{same}_L\). On frames, \ours{} timestamps \(F\) at \(\max\{e_s: s\in F\cap T,\ t_F<e_s\}\) when this set is nonempty and \(\mathrm{same}_L(s,t_F,q)=1\) for each of its regions, and at \(t_F\) otherwise. With a \emph{timestamp per token}, each observation in a frame carries its own timestamp (\cref{fig:theory}b). \ours{} writes on it the timestamp of the region rule above, and on an outdated observation \((x,t,v)\) with \(t<e_x\) a timestamp before every capture time of the history. This keeps every current entry of a region after every outdated one, so the latest-timestamp rule selects as under the region rule, and it widens the gap of \cref{prop:soft}. Deleting a whole stored frame removes all of its observations and deleting tokens removes some of them, so \cref{thm:keep} covers both, and state filtering in \cref{sec:experiments} is the filtering memory.

\Cref{sec:redate} implements these timestamps with two changes that keep every current entry of a region after every outdated one (\cref{alg:redate}). A frame with a restored region takes the time 2~s after its latest restoring action on all of its tokens outside outdated regions. A frame logged with an action shows the state before it, and every other region of the target view that the frame shows has stayed unchanged since its capture, so its observation is current. Outdated tokens take a time before every retrieved frame.

Let every observation carry the camera pose \(c\) of its frame, and let a query have a target pose \(g\) with a set \(N(g)\) of poses near it. A region with a current observation is \emph{seen from the target pose} when some current observation has its pose in \(N(g)\), and \emph{seen only from other poses} otherwise. A \emph{transported observation} of \(x\) is an entry with state \(w(x,q)\) and a pose in \(N(g)\) that a memory adds without a capture, as a render of the region from the target pose (\cref{fig:theory}c).

\renewcommand{\restatedname}{Assumption~\ref*{asm:view}}\begin{restateddefinition}[The generator copies from the target view] The selection rule of a query with target pose \(g\) selects among the entries whose pose lies in \(N(g)\).\end{restateddefinition}

\subsection{Proof of Lemma 1}

\begin{lemma}[Group actions reveal reversions]\label{lem:log} Let the actions on a region \(x\) act on its states through a group \(G_x\) that acts freely on \(V\): each logged action on \(x\) names an element \(g\in G_x\), and an admissible world applies it as \(w(x,s)=g\cdot w(x,s-1)\) at the time \(s\) of the action. Then \(w(x,q)=w(x,t)\) exactly when the product of the elements named by the actions on \(x\) in \((t,q]\) is the identity, so \cref{asm:log} holds for \(x\).\end{lemma}

\begin{proof} Let \(g_1,\dots,g_k\) be the elements named by the actions on \(x\) in \((t,q]\), in the order of their times, and let \(h=g_k\cdots g_1\), a function of the log alone. By \cref{asm:agent} the state of \(x\) is constant between actions, so \(w(x,q)=h\cdot w(x,t)\). The action is free, so \(h\cdot v=v\) holds exactly when \(h\) is the identity, and \(\mathrm{same}_L(x,t,q)\), which equals one exactly when \(h\) is the identity, has the property that \cref{asm:log} requires.\end{proof}

A door is acted on by the group of order two, whose generator exchanges open and closed, and a receptacle by the group of subsets of items under symmetric difference, since putting an item on it and taking that item away each toggle the item's membership; both actions are free.

\subsection{Proof of Theorem 1}

\renewcommand{\restatedname}{Theorem~\ref*{thm:keep}}\begin{restatedplain}[Keep] On a reverted region a deletion-based memory holds no current observation, so every selection rule is wrong on it. A memory that keeps every observation holds a current observation of every reverted region.\end{restatedplain}

\begin{proof} Let \((x,t^\star,v^\star)\) be the latest observation of \(x\). The region is reverted, so \(v^\star\ne p\). Every current observation \((x,t,p)\) has \(t<t^\star\), because the latest observation is outdated, and \(p\ne v^\star\), so the deletion-based memory removes it. The entries of the deletion-based memory about \(x\) are therefore outdated observations, and a selection rule that selects among them is wrong. A reverted region has a current observation by definition, and a memory that keeps every observation keeps that one.\end{proof}

\subsection{Proof of Lemma 2}

\begin{lemma}[Timestamps]\label{lem:timestamp} Let a memory keep every observation and hand the selection rule no timestamps. For two states \(u\ne v\), and for up-to-date and for reverted regions alike, there are two histories that hand the rule the same two states and have different current states. Every deterministic selection rule is wrong on one of them, and a randomized rule is correct on them with probabilities that sum to at most one. Under capture timestamps the latest-timestamp rule is correct on every up-to-date region.\end{lemma}

\begin{proof} Let \(t_1<t_2\le q\) be two times. For up-to-date regions take \(H=\{(x,t_1,u),(x,t_2,v)\}\), which arises in an admissible world of a log with one action on \(x\) in \((t_1,t_2]\) and has current state \(v\), and \(H'=\{(x,t_1,v),(x,t_2,u)\}\), which arises in the same way and has current state \(u\). For reverted regions add a second action on \(x\) after \(t_2\) that restores the earlier state, so that \(H\) has current state \(u\) and \(H'\) has current state \(v\). In each pair both histories hand the selection rule the same two states, so it selects the observation with state \(v\) with one probability \(a\) in both and the observation with state \(u\) with one probability \(b\) in both, and \(a+b\le1\). The rule is correct with probability \(a\) on the history whose current state is \(v\) and with probability \(b\) on the other, so the two probabilities sum to at most one, and a deterministic rule has \(\{a,b\}=\{0,1\}\) and is wrong on one of the two. Under capture timestamps, on an up-to-date region the observation with the latest timestamp is the latest observation, which is current by definition.\end{proof}

\subsection{Proof of Theorem 2 and Corollary 1}

\renewcommand{\restatedname}{Theorem~\ref*{thm:valid}}\begin{restatedplain}[Valid time] (i) Under \ours{}'s timestamps the latest-timestamp rule is correct on every up-to-date and every reverted region. (ii) For two states \(u\ne v\) there are two worlds that produce the same history, one in which the region is up to date and one in which it has reverted, so a memory that does not read the action log is correct with any deterministic selection rule in at most one of them, and with a randomized rule in the two with probabilities that sum to at most one. (iii) A timestamp assignment makes the latest-timestamp rule correct on every up-to-date and every reverted region of every history exactly when it timestamps every current observation of a region after every outdated observation of it. (iv) Let a timestamp assignment keep every outdated observation at its capture time and never timestamp an observation before its capture. If it makes the latest-timestamp rule correct on every up-to-date and every reverted region of every history, it timestamps every current observation captured before \(e_x\) at \(e_x\) or later, and \ours{}'s timestamps meet this bound with equality.\end{restatedplain}

\begin{proof} Part (i). Write \(e=e_x\) and consider the three groups of observations of \(x\). An observation with \(t\ge e\) was taken after the state of \(x\) last changed, so its state is \(p\), and its timestamp is \(t\ge e\). An observation with \(t<e\) and state \(p\) has \(\mathrm{same}_L(x,t,q)=1\) by \cref{asm:log}, and its timestamp is \(e\). An observation with \(t<e\) and another state has \(\mathrm{same}_L(x,t,q)=0\), and its timestamp is \(t<e\). Every current observation therefore has a timestamp of at least \(e\), and every outdated observation has a timestamp below \(e\). An up-to-date or reverted region has a current observation, so the latest timestamp is at least \(e\), every entry that carries it is a current observation, and the latest-timestamp rule is correct.

Part (ii). For \(t_1<t_2<q\) the history \(\{(x,t_1,u),(x,t_2,v)\}\) arises in an admissible world with one action on \(x\) in \((t_1,t_2]\), where \(x\) is up to date with current state \(v\) and \(\mathrm{same}_L(x,t_1,q)=0\), and in an admissible world with a second action in \((t_2,q]\), where \(x\) has reverted to \(u\) and \(\mathrm{same}_L(x,t_1,q)=1\). A memory that does not read the log sees the same history, times and query in both, so its selection rule selects the observation of state \(v\) with one probability \(a\) and the observation of state \(u\) with one probability \(b\) in both, with \(a+b\le1\). It is correct with probability \(a\) in the first world and \(b\) in the second, and a deterministic rule is correct in at most one of them.

Part (iii). When every current observation is timestamped after every outdated observation of its region, every entry with the latest timestamp is a current observation, and a region with a current observation makes the latest-timestamp rule correct. When an outdated observation \(b\) is timestamped at or after a current observation \(a\) of one region, the history \(\{a,b\}\) arises in an admissible world, a timestamp assignment of the observation gives \(a\) and \(b\) the same timestamps there, and \(b\) carries the latest timestamp, so the latest-timestamp rule is wrong. Part (i) shows that \ours{}'s timestamp assignment places every current observation after every outdated one.

Part (iv). Let \(o=(x,t,p)\) be a current observation with \(t<e_x\). The action at \(e_x\) changed the state of \(x\), so \(w(x,e_x-1)\ne p\), and \(b=(x,e_x-1,w(x,e_x-1))\) is an outdated observation of the same world. On the history \(\{o,b\}\) the region has reverted, the assignment keeps \(b\) at \(e_x-1\), and part (iii) requires \(d(o)>e_x-1\), so \(d(o)\ge e_x\). \ours{} gives \(o\) the timestamp \(e_x\) and leaves every observation captured since \(e_x\) and every outdated one at its capture time, so it meets the bound with equality and is correct by part (i).\end{proof}

\renewcommand{\restatedname}{Corollary~\ref*{cor:noreturn}}\begin{restatedplain}[No return in the input] Under the timestamps of \ours{}, the entries of every up-to-date or reverted region are, under capture timestamps, the entries of a history of a world consistent with \cref{asm:agent} in which the region is up to date and never held its current state before \(e_x\).\end{restatedplain}

\begin{proof} Fix an up-to-date or reverted region \(x\) with current state \(p\), and write \(e=e_x\). If no action changed \(x\), every observation of \(x\) shows \(p\), \ours{} keeps every capture time, and the world \(w\) itself has the property. Otherwise the action at \(e\) changed \(x\), so \(u=w(x,e-1)\) is a state other than \(p\). Define \(w'\) equal to \(w\) on every other region and, on \(x\), by \(w'(x,s)=p\) for \(s\ge e\), \(w'(x,s)=w(x,s)\) for \(s<e\) with \(w(x,s)\ne p\), and \(w'(x,s)=u\) for \(s<e\) with \(w(x,s)=p\). By the proof of \cref{thm:valid}(i), every current entry carries a timestamp \(d\) with \(e\le d\le q\) and shows \(p=w'(x,d)\), and every outdated entry keeps its capture time \(t<e\) and shows \(w(x,t)\ne p\), which is \(w'(x,t)\). Each entry, read with its timestamp as its capture time, is therefore an observation of \(w'\). The world \(w'\) satisfies \cref{asm:agent} for the log that lists an action on \(x\) at every time at which \(w'(x,\cdot)\) changes, beside the actions of \(L\) on other regions. It holds \(p\) from \(e\) to \(q\), so the region is up to date, and it never holds \(p\) before \(e\).\end{proof}

The generator is shown, on every region where a correct selection is possible, a history in which the current state is new, which the latest-timestamp rule reads correctly. The per-token rule of \cref{sec:redate} moves the outdated tokens of a frame to one time; it keeps every current entry after every outdated one, and the construction applies whenever a region's outdated entries at one timestamp share their state.

\subsection{Proof of Lemma 3}

\begin{lemma}[Re-timing that ignores content]\label{lem:retime} A timestamp assignment that moves the capture times by one strictly increasing map, as an even spacing in the true order or a common rescaling of the gaps, selects the observation that capture time selects, so it is correct on every up-to-date and wrong on every reverted region. An assignment that does not read the action log, as one time for every frame or a sink moved next to the present, is wrong on one of the two histories of \cref{thm:valid}(ii).\end{lemma}

\begin{proof} A strictly increasing map \(f\) keeps the order of any two capture times, \(f(t)\le f(t')\) exactly when \(t\le t'\), so the entries with the latest timestamp \(f(t)\) are those with the latest capture time, and the latest-timestamp rule selects what capture time selects. Capture time is correct on up-to-date regions by \cref{lem:timestamp} and wrong on reverted regions, whose latest observation is outdated. A memory whose timestamps do not read the log is a memory that does not read the log, and \cref{thm:valid}(ii) applies to it.\end{proof}

\subsection{Proofs of Proposition 1 and Corollary 2}

The \emph{soft rule} selects an entry \(o\) about \(x\) with probability \(e^{\beta\tau(o)}h(o)/Z\) and draws the region from the generator's own prediction with probability \(G/Z\). Here \(\tau(o)\) is the timestamp of \(o\), \(h(o)\in[h_{\min},h_{\max}]\) a positive weight independent of \(\tau\), \(G=h_0e^{\beta\tau_q}\) with a weight \(h_0\ge0\) and the timestamp \(\tau_q\) of the generated frame, and \(Z\) the sum of \(G\) and the numerators over the entries about \(x\). A draw from the prediction does not show a state that only the memory holds, so on a region the agent changed it is wrong. With \(h_0=0\) the rule draws only from the memory. The soft rule is the conditional logit choice of McFadden~\citep{mcfadden1974conditional}, with the prediction as its outside option. Under a constant hazard \(\beta\) of change and no return, \(e^{-\beta(q-\tau)}\) is the probability that a view captured at \(\tau\) remains current at \(q\), the weight of the soft rule; valid time restarts this clock at the action that confirms the content again.

\renewcommand{\restatedname}{Proposition~\ref*{prop:soft}}\begin{restatedplain}[Soft rule] Under the soft rule, (i) the probability of a correct selection rises with the timestamp of every current entry and falls with the timestamp of every outdated one, (ii) when every current timestamp exceeds every outdated one by \(\Delta\), an outdated entry is selected with probability at most \(\frac{n_O h_{\max}}{n_C h_{\min}}e^{-\beta\Delta}\), for \(n_C\) current and \(n_O\) outdated entries, which vanishes as \(\beta\to\infty\), and (iii) on a reverted region, deleting every outdated entry under capture timestamps is correct with probability \(A/(A+h_0e^{\beta\tau_q})\), which \ours{}'s timestamps exceed whenever their outdated weight is below \(h_0e^{\beta\tau_q}(A'/A-1)\), for current weights \(A\) under capture time and \(A'>A\) under valid time.\end{restatedplain}

\begin{proof} Let \(A\) be the sum of \(e^{\beta\tau(o)}h(o)\) over the current entries and \(B\) the sum over the outdated ones, so that a correct selection has probability \(A/(A+B+G)\). Part (i). For \(A\le A'\) and \(B'\le B\), \(AB'\le A'B\) and \(AG\le A'G\), so \(A(A'+B'+G)\le A'(A+B+G)\) and \(A/(A+B+G)\le A'/(A'+B'+G)\). Raising the timestamp of a current entry raises \(A\) and leaves \(B\) and \(G\), and lowering the timestamp of an outdated entry lowers \(B\) and leaves \(A\) and \(G\). Part (ii). Let \(\tau_C\) be the least current timestamp, so that every outdated timestamp is at most \(\tau_C-\Delta\). Then \(A\ge n_C h_{\min}e^{\beta\tau_C}\) and \(B\le n_O h_{\max}e^{\beta(\tau_C-\Delta)}\), and an outdated entry is selected with probability \(B/(A+B+G)\le B/A\le\frac{n_O h_{\max}}{n_C h_{\min}}e^{-\beta\Delta}\). For \(\Delta>0\) the bound vanishes as \(\beta\to\infty\), and so does the share \(B/(A+B)\le B/A\) of outdated entries among the draws from memory, which then follow the latest-timestamp rule. Part (iii). Deleting every outdated entry leaves the current entries under their capture timestamps, with weight \(A\), and no outdated entry, so a correct selection has probability \(A/(A+G)\). On a reverted region the latest observation is outdated and was captured before \(e_x\), so every current observation was captured before \(e_x\), and \ours{} moves each to \(e_x\), which raises every term of \(A\) for \(\beta>0\) and gives \(A'>A\). With outdated weight \(B'\) under \ours{}, \(A'/(A'+B'+G)>A/(A+G)\) exactly when \(A'(A+G)>A(A'+B'+G)\), that is when \((A'-A)G>AB'\), or \(B'<G(A'/A-1)\). With \(h_0=0\) deletion is correct with probability one and the bound is zero.\end{proof}

The log-odds of the current against the outdated state, the quantity of \cref{sec:finding,sec:robustness}, is linear in the timestamps under the soft rule, and the generator's prediction drops out of it.

\begin{corollary}[Sensitivity]\label{cor:sensitivity} Under the soft rule, let the current entries about a region share one timestamp \(\tau_C\) and the outdated entries one timestamp \(\tau_O\), with summed weights \(H_C\) and \(H_O\). (i) The log-odds of drawing the current state against drawing the outdated one is \(\ell=\beta(\tau_C-\tau_O)+\log(H_C/H_O)\), whatever the weight \(h_0\) of the prediction. (ii) Reversing the two timestamps changes \(\ell\) by \(2\beta|\tau_C-\tau_O|\), so half that change, the timestamp sensitivity, is \(\beta|\tau_C-\tau_O|\). (iii) For two generators of sharpness \(\beta_1\) and \(\beta_2\) that read the same entries, every change of the two timestamps changes their log-odds in the ratio \(\beta_1/\beta_2\), the ratio of their sensitivities.\end{corollary}

\begin{proof} With \(A=H_Ce^{\beta\tau_C}\), \(B=H_Oe^{\beta\tau_O}\) and \(G=h_0e^{\beta\tau_q}\), the soft rule draws the current state with probability \(A/(A+B+G)\) and the outdated one with probability \(B/(A+B+G)\), so their ratio is \(A/B\) and \(\ell=\log(A/B)=\beta(\tau_C-\tau_O)+\log(H_C/H_O)\), which omits \(G\). Part (ii) follows by exchanging \(\tau_C\) and \(\tau_O\). For part (iii), a change from \((\tau_C,\tau_O)\) to \((\tau'_C,\tau'_O)\) changes \(\ell\) by \(\beta[(\tau'_C-\tau'_O)-(\tau_C-\tau_O)]\), which is proportional to \(\beta\) with a factor that the entries and timestamps alone fix, and by part (ii) the sensitivities of the two generators stand in the ratio \(\beta_1/\beta_2\).\end{proof}

When entries of one state carry different timestamps, \(\ell\) is a difference of two log-sum-exp terms, and the proportion holds to the extent that each state's timestamps move together, as they do when \ours{} re-dates a frame's restored region and moves its outdated tokens to one timestamp.

\subsection{Proof of Theorem 3}

\renewcommand{\restatedname}{Theorem~\ref*{thm:frame}}\begin{restatedplain}[Re-date, keep the frame] Let \(z\) be a region of the target view with a current observation. (i) Let every frame carry one timestamp. The filtering memory is correct on \(z\) exactly when some current observation of \(z\) lies in a consistent frame. \ours{} with the latest-timestamp rule is correct on \(z\) exactly when \(z\) is up to date or some current observation of \(z\) lies in a consistent frame, so it is correct wherever filtering is, and also on every up-to-date region whose current observations all lie in inconsistent frames. (ii) Let two frames cross: one shows a region \(y\) in its current state and \(z\) in an outdated state, the other shows \(y\) outdated and \(z\) current, and no other frame shows \(y\) or \(z\). Every assignment of one timestamp per frame leaves the latest-timestamp rule wrong on \(y\) or on \(z\), and filtering is wrong on both. (iii) Let every token carry a timestamp of its own. \ours{}'s timestamps then make the latest-timestamp rule correct on every up-to-date and every reverted region of the target view.\end{restatedplain}

\begin{proof} Part (i), filtering. A frame that shows \(z\) in an outdated state is inconsistent, so every observation of \(z\) that the filtering memory hands over is current, and every selection rule is correct exactly when it receives one. The memory hands over none exactly when no current observation of \(z\) lies in a consistent frame.

Part (i), \ours{}. A frame that \ours{} timestamps after its capture has every region of \(F\cap T\) acted on since \(t_F\) at its state of \(t_F\), and every other region of \(F\cap T\) unchanged by \cref{asm:agent}, so it is consistent. If \(z\) is up to date, its latest observation is current, every frame timestamped after the capture of that observation is a re-dated frame, which is consistent and shows \(z\) at \(p\), and every other frame that shows \(z\) keeps a timestamp at most that of the latest observation, so the entries about \(z\) with the latest timestamp show \(p\). If \(z\) has reverted with latest observation at \(t^\star\), a current observation of \(z\) in a consistent frame \(F\) has \(t_F<t^\star<e_z\), so \(F\) is timestamped at least \(e_z>t^\star\), and every entry about \(z\) timestamped after \(t^\star\) lies in a consistent frame and shows \(p\). When no current observation lies in a consistent frame, every current observation of a reverted \(z\) keeps a timestamp below \(t^\star\), and the latest-timestamp rule selects the observation at \(t^\star\), which is outdated. The last clause of part (i) follows from the two characterizations.

Part (ii). Let \(F_1\) show \(y\) current and \(z\) outdated, and \(F_2\) show \(y\) outdated and \(z\) current, under the timestamps \(d_1\) and \(d_2\). The entries about \(y\) are its observations in \(F_1\) and \(F_2\), and so are the entries about \(z\). If \(d_1\le d_2\), the outdated observation of \(y\) in \(F_2\) carries the latest timestamp among the entries about \(y\), and the rule is wrong on \(y\). Otherwise \(d_2<d_1\), the outdated observation of \(z\) in \(F_1\) carries the latest timestamp among the entries about \(z\), and the rule is wrong on \(z\). Both frames show an outdated region of the target view, so filtering hands over no observation of \(y\) or \(z\) and is wrong on both.

Part (iii). With a timestamp per token, every observation of a region carries the timestamp that the region rule of \cref{sec:validtime} gives it, whatever else its frame shows, and \cref{thm:valid}(i) applies region by region.\end{proof}

\subsection{Proof of Theorem 4}

\renewcommand{\restatedname}{Theorem~\ref*{thm:view}}\begin{restatedplain}[View] Let \cref{asm:view} hold. (i) On a region whose current observations all have poses outside \(N(g)\), every memory of captured observations is wrong under every timestamp assignment. (ii) Under \ours{}'s timestamps the latest-timestamp rule is correct on every up-to-date and every reverted region with a current observation from a pose in \(N(g)\). (iii) Let a memory add to every history an observation of the current state with a pose in \(N(g)\). A timestamp assignment makes the latest-timestamp rule correct on every such region of every history exactly when it timestamps every current observation with a pose in \(N(g)\) after every outdated observation with a pose in \(N(g)\). With the other observations under \ours{}'s timestamps, the timestamp \(e_x\) of the added observation meets this condition, and a timestamp before an outdated observation from \(N(g)\) does not. (iv) \ours{} is correct on every region on which any memory of captured observations, under any timestamps and any selection rule, is correct.\end{restatedplain}

\begin{proof} Part (i). The entries in \(N(g)\) are observations of the history from poses in \(N(g)\), and on a region seen only from other poses every such observation is outdated, so the selection rule selects an outdated entry or has none. Part (ii). The proof of \cref{thm:valid}(i) bounds the timestamp of each observation by itself: every current observation is timestamped at least \(e_x\) and every outdated one below \(e_x\). The bound survives the restriction to \(N(g)\), and a region seen from the target pose has a current entry there. Part (iii). The first sentence is \cref{thm:valid}(iii) applied to the entries in \(N(g)\), which include the transported observation: an assignment that keeps every current entry from \(N(g)\) after every outdated one leaves only current entries at the latest timestamp, and one that places an outdated entry \(b\) at or after a current entry \(a\) is wrong on the history whose entries in \(N(g)\) are \(a\) and \(b\). Every outdated observation of \(x\) was captured before the state of \(x\) last changed, at a time below \(e_x\), and \ours{} leaves it there or moves it earlier, so a transported observation timestamped at \(e_x\), beside current observations that \ours{} timestamps at \(e_x\) or later, meets the condition. A transported observation timestamped before an outdated entry in \(N(g)\) is a current entry timestamped before an outdated one, and on the history of these two entries the rule selects the outdated one. On one fixed history a weaker condition suffices, namely that the latest timestamp from \(N(g)\) belongs to current entries alone: a current entry at time 0, an outdated one at 1 and a current one at 2 make the rule correct, and the quantifier over every history makes the condition of the theorem necessary. Part (iv). A memory of captured observations that is correct on a region selects a current entry from a pose in \(N(g)\), which is a captured current observation from \(N(g)\). The region then has a current observation, so it is up to date or reverted, it is seen from the target pose, and \ours{} is correct on it by part (ii).\end{proof}

\subsection{Proof of Corollary 3}

\begin{corollary}\label{cor:table} Under the latest-timestamp rule, and under \cref{asm:view} for the last column, each memory of \cref{tab:theory} is correct or wrong on one region as the table states.\end{corollary}

\begin{proof} Deletion. The deletion-based memory keeps the latest observation of a region, so on an up-to-date region the entry with the latest capture timestamp is current and the rule is correct, and on a reverted region it is wrong by \cref{thm:keep}. Of two crossing regions one is up to date and the other has reverted. Shared time or none. \Cref{lem:timestamp} gives, for up-to-date and for reverted regions, two histories on one of which every deterministic rule is wrong. Capture order. The rule is correct on up-to-date regions by \cref{lem:timestamp} and wrong on reverted regions, and \cref{lem:retime} extends both to every strictly increasing map of the capture times; decay by age lowers the chance that an old frame is retrieved and keeps the order of the timestamps the generator reads. Sink at the present. \Cref{lem:retime}, second part. State filtering. \Cref{thm:frame}(i), and (ii) for crossing regions. \ours{}, per frame. \Cref{thm:frame}(i) for up-to-date and reverted regions, and (ii) for crossing regions. \ours{}, per token. \Cref{thm:frame}(iii). Other poses. Every memory of the first seven rows holds only captured observations, so each is wrong on a region seen only from other poses by \cref{thm:view}(i). \ours{} with a frame of the current state at the target pose under the timestamp \(e_x\) timestamps every current entry after every outdated one, so it is correct on every kind of region by \cref{thm:valid}(iii) and by \cref{thm:view}(iii) under \cref{asm:view}.\end{proof}

\section{Implementation}\label{sec:implementation}

This appendix specifies \ours{} inside FAR (\cref{sec:redate}), its constants, the rule behind every compared design and the timestamp conditions of \cref{sec:finding}. The harness changes no weight and no sampler setting.

\subsection{The state tracker}\label{sec:tracker}

The state tracker reads the agent's action log, a list of actions with their frames and the objects they name, and keeps one record per \emph{manipulated region}: the spot where an item first stood, a receptacle, or a door (\cref{alg:tracker,tab:s-tracker}). A manipulated region is located at the mean of the points 1~m along the optical axis of the stored frames nearest its actions in time, and an action with no stored frame within 1~s places nothing. For receptacles and doors the record computes the function \(\mathrm{same}_L\) of \cref{asm:log} by \cref{lem:log}. A spot where an item stood falls outside that group form because the simulator sets a returned item down at a new position, and the tracker treats it as empty from the first pickup of its item.

\begin{algorithm}[tbp]
\caption{The state tracker, in Python-like pseudocode. \texttt{START} holds the values before a region's first action (\cref{tab:s-tracker}).\label{alg:tracker}}
\begin{lstlisting}
def track(log):  # (frame, verb, target), in order
    record = {}  # region -> [(frame, value)]
    held, site = None, {}  # item -> its receptacle
    def value(region):  # what the last action left
        if region in record:
            return record[region][-1][1]
        return START[region.kind]
    for frame, verb, target in log:
        if verb == "PickupObject":
            region = site.pop(target, Spot(target))
            new = (EMPTY if region.kind == "spot"
                   else value(region) - {target})
            held = target
        elif verb == "PutObject":
            region = Receptacle(target)
            new = value(region) | {held}
            site[held], held = region, None
        else:  # OpenObject, CloseObject
            region = Door(target)
            new = verb == "OpenObject"
        record.setdefault(region, [])
        record[region].append((frame, new))
    return record

def same(record, x, t, q):  # same_L(x, t, q)
    return (value_at(record, x, t)
            == value_at(record, x, q))
\end{lstlisting}
\end{algorithm}

\begin{algorithm}[tbp]
\caption{\ours{}'s timestamps and context at one step, in Python-like pseudocode, for the target pose \texttt{g} at frame \texttt{q}.\label{alg:redate}}
\begin{lstlisting}
def redate(retrieved, memory, record, log, g, q):
    play = [x for x in record if in_view(g, x.place)]
    extra = recall(memory, record, log, g, q)
    context = replace_lowest(retrieved, extra)
    for f in context:  # 12 frames
        f.tau = full(f.capture)  # one per token
        restored = [x for x in play if f.holds(x)
                    and acted(record, x, f.t, q)
                    and same(record, x, f.t, q)]
        if restored:  # valid time, 2 s after
            last = max(last_action(record, x, q)
                       for x in restored)
            f.tau[:] = min(last + 2.0, q - STEP)
    oldest = min(f.tau.min() for f in context)
    for f in context:
        for x in play:
            if (f.holds(x) and acted(record, x, f.t, q)
                    and not same(record, x, f.t, q)):
                f.tau[f.disc(x, 0.8)] = oldest - 2.0
    return context

def recall(memory, record, log, g, q):
    retimed = [f for f in memory if restored_in(f, q)]
    nearest = nearest_pose(retimed, g, k=2, apart=2.0)
    post = [frame_after(memory, a.frame + 2.0, 3.0)
            for a in log[-2:]]  # post-action retrieval
    return nearest + post
\end{lstlisting}
\end{algorithm}

\begin{table*}[tp]
\centering
\caption{The record the state tracker keeps for each kind of manipulated region.\label{tab:s-tracker}}
\small\setlength{\tabcolsep}{4pt}
\begin{tabularx}{\linewidth}{>{\raggedright\arraybackslash\hsize=0.906\hsize}X>{\raggedright\arraybackslash\hsize=0.952\hsize}Xl>{\raggedright\arraybackslash\hsize=1.014\hsize}X>{\raggedright\arraybackslash\hsize=1.128\hsize}X}
\toprule
Region & Value & Before its first action & Action log & Group of \cref{lem:log} \\
\midrule
Spot where an item first stood & holds the item, or empty & holds its item & empty from the first pickup of the item & none, see the text \\
Receptacle & the items the agent put on it and has not taken away & no item put by the agent & a put adds the held item and a pickup removes it & subsets of items under symmetric difference \\
Door & open or closed & the state opposite to its first action & opening and closing & the group of order two \\
\bottomrule
\end{tabularx}
\end{table*}

\paragraph{A log without object names.}

The tracker also runs on a log that keeps only the verb of each action and its place, the point 1~m ahead of the camera of the stored frame nearest the action, which an agent knows from its own gripper and pose. Openings and closings whose places lie within 0.8~m of each other, the radius of a region's disc, act on one door, and puts within 0.8~m act on one receptacle. A pickup takes the item that the agent last put down within 0.8~m and has not picked up since, and otherwise takes a new item whose spot is the pickup's place. \Cref{alg:tracker} runs on these names unchanged, and \cref{sec:analysis} reports \ours{} with this log.

\begin{table*}[tp]
\centering
\caption{The state tracker against the ground-truth frame on the views stored at the target pose, summed over the 14,297 audited rollout steps of the 249 clips at seed 1. Precision is the share of re-dated views that agree with the ground-truth frame on its changed tokens, recall the share of the views to re-date, agreeing views older than the newest disagreeing one, that it re-dates, and Disagreeing the share of views marked outdated that differ from the ground-truth frame (\texttt{far\_ledger\_audit.py}); 1,882 of the 7,926 agreeing re-dated views, and 1,032 of 6,114 without object names, are already the newest.\label{tab:s-tracker-audit}}
\small\setlength{\tabcolsep}{6pt}
\begin{tabular}{lrrrrrr}
\toprule
Action log & Re-dated views & Precision & Views to re-date & Recall & Views marked outdated & Disagreeing \\
\midrule
Object names & 12,713 & 0.62 & 10,702 & 0.56 & 43,441 & 0.88 \\
Verb and place only & 11,358 & 0.54 & 10,702 & 0.47 & 42,686 & 0.86 \\
\bottomrule
\end{tabular}
\end{table*}

\begin{table*}[tp]
\centering
\caption{Constants of \ours{}, the compared designs and the evaluation.\label{tab:s-constants}}
\small\setlength{\tabcolsep}{4pt}
\begin{tabularx}{\linewidth}{>{\raggedright\arraybackslash\hsize=0.630\hsize}X>{\raggedright\arraybackslash\hsize=1.076\hsize}X>{\raggedright\arraybackslash\hsize=1.428\hsize}X>{\raggedright\arraybackslash\hsize=0.867\hsize}X}
\toprule
Constant & Value & Role & Set \\
\midrule
Reach & 1~m & the location of a manipulated region along the optical axis & for post-action retrieval, before re-dating \\
Token radius & 0.8~m & the disc whose cells are a region's tokens & with the transport rules of \cref{sec:views} \\
Holding range & 8~m & the farthest region a stored frame holds & before any run \\
Settle time & 2~s & restored frames after the restoring action, post-action frames after the action, outdated tokens before the oldest frame & for post-action retrieval, before re-dating \\
Window & 3~s & the search window of post-action retrieval, the reveal window of FAR's analysis & for post-action retrieval \\
Recalled frames & 2 and 2 & re-dated frames, and post-action frames from the last two interactions & the second chosen between one and two on subset A, the first following it \\
Pose distance & 2~m & the camera distance that counts as one radian & before any run \\
Exact revisit & 5 mm, 0.2° & a stored frame at the target pose & before any run \\
Masked logit & −20 & a hidden token in the generator's attention & before any run \\
Decay by age & \(\lambda_{\min}=0.1\), \(H=0.3\) of the stored span & the factor of Mem-World in \cref{tab:s-designs} & before any run \\
Same-camera change & latent error above 0.25 of the pair's peak, a peak above 0.10, regions of at least 3 tokens covering at most 0.25 of the grid & the tokens that a later frame of the same camera shows changed, for the two Echo-Forcing rules in \cref{tab:s-designs} & before any run \\
Fixed thresholds & 0.2 and 0.1 & readings without a quality floor (\cref{sec:metric}) & before any table was read at them \\
\bottomrule
\end{tabularx}
\end{table*}

\subsection{Timestamps and context}\label{sec:retiming}

For a target view with camera \(g\) at frame \(q\), a manipulated region is \emph{in play} when its location lies inside the target view. A stored frame taken at frame \(t\) \emph{holds} a manipulated region when the location lies inside its own view within 8~m, and its \emph{tokens} of the region are those whose cells a disc of 0.8~m around the location covers in that view. A region in play that a frame holds is \emph{restored} in the frame when an action touched it in \([t,q)\) and it has the same value at \(t\) and at \(q\), and \emph{outdated} when an action touched it and its value differs. \Cref{alg:redate} gives the timestamps of the retrieved frames at one rollout step and the context they join. They are computed again at every step, so a frame re-dated for one target is timestamped by its capture for another. FAR embeds one timestamp per frame, and \ours{} adds to each token the difference between the embedding of its own timestamp and that of its frame's, before the generator's first block.

\emph{\ours{} per frame}, the variant with one timestamp per frame (\cref{tab:ablation}), re-dates only the frames whose touched regions in play are all restored, and moves no token.

The re-dated frames that join the context are the two nearest the target pose, by the sum of the differences in yaw and pitch in radians and the distance between camera centers over 2~m, newest first on ties and at least 2~s apart. \emph{Post-action retrieval} takes, for each of the last two interactions of the agent before the evaluation leg, the stored frame nearest 2~s after it within a window of 3~s. Both replace the lowest-ranked frames of FAR's retriever, which ranks a re-dated frame by its capture, and a frame that the set already holds takes no second slot. Frames re-dated together keep the order of their capture. \Cref{tab:s-constants} lists the constants, each set before any run or on subset A.

\begin{table*}[tp]
\centering
\caption{The memory designs run inside FAR's memory bank, the published rules they adapt, what each reads besides the frames, and the single change from the published rule. ‡ reads the action log and † the simulator.\label{tab:s-designs}}
\small\setlength{\tabcolsep}{4pt}
\begin{tabularx}{\linewidth}{>{\raggedright\arraybackslash\hsize=0.727\hsize}X>{\raggedright\arraybackslash\hsize=0.709\hsize}X>{\raggedright\arraybackslash\hsize=0.637\hsize}X>{\raggedright\arraybackslash\hsize=1.928\hsize}X}
\toprule
Design & Adapted from & Reads & Change from the published rule \\
\midrule
Latest 12 frames & FAR's temporal baseline~\citep{kim2026learning} & capture order & the memory bank is cut to the context, where the released baseline is a checkpoint of its own \\
FAR & FAR~\citep{kim2026learning} & capture times & none, the system as released \\
View-overlap retrieval & Context as Memory~\citep{yu2025context} & poses & FAR's geometric retriever with its recency term at zero, under the FAR generator \\
Geometric retrieval & WorldMem~\citep{xiao2025worldmem} & poses, capture times & the WorldMem baseline of FAR's release, and its retriever under the FAR generator \\
Decay by age & Mem-World~\citep{zheng2026mem} & capture times & the factor \(\lambda_{\min}+(1-\lambda_{\min})2^{-\mathrm{age}/H}\) ranks FAR's candidates again, not candidates of its own retriever \\
Re-timing in the true order & WorldPlay~\citep{sun2025worldplay} & capture order & the retrieved frames keep their order and take evenly spaced timestamps up to the newest, not a fixed distance \\
Sink at the present & Deep Forcing~\citep{yi2025deep}, Rolling Forcing~\citep{liu2025rolling} & capture order & the two oldest retrieved frames move, in order, between the newest retrieved frame and the current frame \\
Frame deletion by pose & Echo-Forcing~\citep{wu2026echo} & latents, poses & the conflict decay is made hard and moved to frames: a contradicted frame leaves the memory bank \\
Token deletion by later frame & Echo-Forcing~\citep{wu2026echo} & latents, poses & the decay is made hard: tokens that a later frame of the same camera shows changed are hidden \\
Frame deletion at each change† & PermaVid~\citep{yang2026permavid} & simulator change times & views of the camera are deleted at the change, and views from other cameras stay \\
Token deletion at each change† & PermaVid~\citep{yang2026permavid} & change times and masks & the tokens on the change mask are hidden, not whole views \\
Deletion of all earlier frames‡ & WorldCraft~\citep{gu2026worldcraft} & action times & every frame older than the last pickup or put counts as taken before the change, whatever its view \\
Changing content never stored† & LSM-World~\citep{wang2026latent} & change masks & the change mask marks the changing content in every stored frame of a camera \\
State filtering‡ & ONEIRA~\citep{yang2026oneira} & action log & \ours{}'s state tracker supplies the world state, and a hidden frame returns once its states do \\
Oracle token deletion† & PermaVid~\citep{yang2026permavid} & change masks & the stale tokens of every outdated frame are hidden at every read, with the current state known \\
Oracle frame deletion† & ONEIRA~\citep{yang2026oneira} & change masks & the mask decides which stored frames at the camera are stale, as a perfect state table would \\
Re-dating from ground truth† & \ours{} & ground-truth frame & the frames at the target pose that agree with the ground-truth frame on its changed tokens are re-dated \\
\bottomrule
\end{tabularx}
\end{table*}

\begin{table*}[tp]
\centering
\caption{The timestamp conditions, for the timestamps \(\tau\) of a retrieved set with extremes \(\tau_{\min}\) and \(\tau_{\max}\).\label{tab:s-conditions-def}}
\small\setlength{\tabcolsep}{6pt}
\begin{tabular}{lll}
\toprule
Condition & Timestamps the generator reads & Keeps \\
\midrule
Kept & \(\tau\) & order and gaps \\
One time & the median of the set, for every retrieved frame & neither \\
Permuted & a permutation of the set's timestamps & the values \\
Reversed & \(\tau_{\min}+\tau_{\max}-\tau\) & the gaps, in reverse \\
Equal gaps & evenly spaced in the true order between \(\tau_{\min}\) and \(\tau_{\max}\) & the order \\
Gaps times \(c\) & \(\tau_{\max}-c\,(\tau_{\max}-\tau)\), for \(c=0.1\) and \(c=4\) & the order \\
Slots permuted & \(\tau\), with the retrieved frames in one other fixed slot order & order and gaps \\
\bottomrule
\end{tabular}
\end{table*}

\subsection{Memory designs compared}

Every design runs inside FAR's memory bank with the released generator, retriever and sampler, so the rows of \cref{tab:s-designs} differ in the memory rule alone. The privileged rows (†) read the simulator and are references with exact information about one part of the problem, which bounds no other part.

\subsection{Timestamp conditions}

The timestamp conditions of \cref{sec:finding} rewrite the timestamps \(\tau\) of the retrieved frames inside the generator (\cref{tab:s-conditions-def}). They leave the memory bank, the retriever's weights, the timestamps the retriever reads and the timestamp of the current frame unchanged. A second switch gives a generator the memory module of the other checkpoint, which is how the FAR generator runs under geometric retrieval.

\section{Evaluation}\label{sec:evaluation}

\subsection{Protocol}\label{sec:protocol}

All runs execute FAR's released \texttt{eval\_rollout\_thor.py} under its published settings (\cref{tab:s-settings}), and \cref{tab:s-runs} lists the clips and seeds for each run. Within a seed every condition of a clip shares its sampling noise, and each run is paired with its reference on the clips both completed. The benchmark's evaluator measures state accuracy as the fraction of evaluated steps whose changed-region crop lies closer in LPIPS to the ground-truth frame than the pose-matched stale frame does, over steps where those two references differ by at least 0.05. Following FAR's analysis code, steps within 3~s of a container opened during the evaluation leg form the container stratum and the remaining steps form the surface stratum. Because few clips contain container openings, our analysis averages surface accuracy over clips and pools container steps; FAR's released analysis (\cref{sec:reported}) averages containers over clips after dropping the 0.8~s before each opening. FAR's project page reports surface and container accuracies of 0.802 and 0.757 on the full test set, and our rerun gives 0.804 and 0.727 on the first 400 clips. The exporter \texttt{far\_state\_export.py} writes per-clip counts, and the analysis \texttt{far\_state\_analysis.py} computes every number in \cref{sec:results} from them.

\begin{table*}[tp]
\centering
\caption{Rollout, memory and hardware settings of the evaluation runs.\label{tab:s-settings}}
\small\setlength{\tabcolsep}{6pt}
\begin{tabular}{ll}
\toprule
Setting & Value \\
\midrule
Checkpoints & step 700k on AI2-THOR, step 500k on AI2-THOR-dyn \\
Rollout & stride of four frames, context of 12 frames, 20 DDIM steps \\
Memory & store of 1,000 exploration frames, retrieval chunk of 10; the memory bank's history stays real \\
Labels & interaction labels fed as in FAR \\
Hardware & NVIDIA H100 GPUs, and an NVIDIA L20X GPU for the HY-WorldPlay probe \\
AI2-THOR-dyn & FAR's both-ends probe, five epochs, on trajectories whose second agent stands at exactly one corridor end \\
\bottomrule
\end{tabular}
\end{table*}

\begin{table*}[tp]
\centering
\caption{Clips and seeds for each run. Subsets A (shards 0 to 7, 102 of 104 clips evaluated, used to choose rules) and B (shards 8 to 15, 93 evaluated clips, run after rule selection) form the first 200 eligible test clips in the collector's seeded order.\label{tab:s-runs}}
\small\setlength{\tabcolsep}{4pt}
\begin{tabularx}{\linewidth}{>{\raggedright\arraybackslash\hsize=1.000\hsize}Xll}
\toprule
Runs & Clips & Seed \\
\midrule
Audit of the readings (\cref{sec:metric}) & 400 test clips, and 200 held-out audit clips &  \\
Timestamp conditions & subsets A and B & 0 \\
Timestamp conditions & the 249 clips with a repeat-state step & 1 \\
Timestamp conditions & the clips of shards 0 to 15 of the complete index & 0 \\
Memory designs, \ours{} and its parts, the stores & the 249 clips with a repeat-state step & 1 \\
The stores & subsets A and B & 0 and 1 \\
The seed-0 rows of \cref{tab:s-seeds} & the 249 clips with a repeat-state step & 0 \\
FAR, \ours{}, \ours{} per frame, recalled frames under capture timestamps, state filtering, two deletion rules & the 246 clips eligible at seed 2 & 2 \\
FAR and \ours{}, recall rules under the LPIPS test & the complete index, 1,309 clips & 0 \\
\bottomrule
\end{tabularx}
\end{table*}

\subsection{The state metric and its audit}\label{sec:metric}

\begin{figure}[tbp]
\centering
\includegraphics[width=\linewidth,height=0.82\textheight,keepaspectratio]{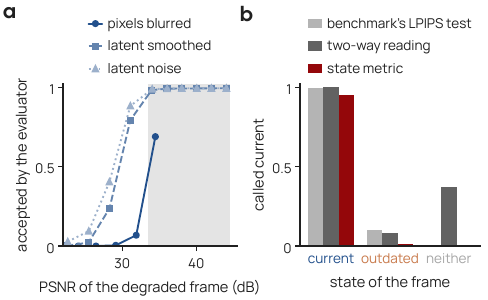}
\caption{(a) Simulator frames of the current state accepted by the benchmark's evaluator under three degradations, with FAR's rendering PSNR band in gray. (b) Simulator frames of known state called current by each reading, with the state metric in red.\label{fig:audit}}
\end{figure}

The benchmark's LPIPS evaluator places the stale frame on the decision boundary by calling a step correct whenever the rendered crop lies closer to the ground-truth frame than the stale frame does. We audit this rule by replacing the render with a second simulator frame of known state from the same camera, degrading 2,757 frames of the current state and 1,726 of the outdated state across 400 clips under latent noise, latent smoothing and pixel blur (\cref{fig:audit}a). After a tokenizer round trip the evaluator accepts 0.99 of current-state frames and 0.10 of outdated-state frames. It accepts 0.98 or more down to 34~dB under latent noise and smoothing, and 0.69 at 34.5~dB and 0.07 at 31.9~dB under pixel blur.

The state metric of \cref{sec:setup} separates the current and outdated states in the tokenizer's latent space. At each audit level \(\kappa\) is the 95th percentile of \(\delta_t/d\) over current-state frames, and a step takes the largest value among the three degradations nearest its PSNR, set by pixel blur at 0.295 above 33~dB and 0.475 at 31.9~dB. Steps below the 30.4~dB PSNR floor, where the calibration no longer separates the states, read as low quality, and outcome shares sum over current, outdated, ambiguous and low quality. Defining \(\kappa\) by 0.95 acceptance on current-state frames preserves 0.95 acceptance on 200 held-out audit clips (\cref{fig:audit}b).

Three checks verify that the metric's conclusions hold without the PSNR floor, which measures PSNR against the ground-truth frame and therefore penalizes sharp outdated renders. In latent space, a linear mixture with weight \(\lambda\) on the ground-truth frame has \(\delta_t=(1-\lambda)^2d\); an encoded pixel mixture of the two decoded frames, a ghost, lies off that line, and the metric calls 0.43 of 1,620 even ghosts current and 0.12 outdated. Every comparison is therefore repeated without a quality floor at the fixed thresholds 0.2 (which calls 0.19 of even ghosts current and 0.05 outdated) and 0.1 (which calls 0.02 of them a state); on the held-out audit clips, threshold 0.2 calls 0.96 of current-state frames current under latent noise and 0.91 under blur at 34.5~dB, and threshold 0.1 calls 0.95 and 0.64. Memory banks with known answers confirm the reading on generated frames (\cref{tab:s-reading}): with the bank cut before the change, the metric calls 0.01 current and 0.56 outdated, and with current views alone at the evaluated camera it calls 0.68 current, a lower bound on its sensitivity on renders, since every generator failure counts against it.

\begin{table*}[tp]
\centering
\caption{Share of inputs of known answer that each reading calls current and the evaluator calls correct, for simulator frames under latent noise, 1,620 even mixtures of the ground-truth and outdated frames (ghosts), and FAR renders from memory banks with a known answer. The two-way reading calls a frame current when its changed tokens lie nearer the ground-truth frame than the stale frame, with no threshold and no quality floor.\label{tab:s-reading}}
\small\setlength{\tabcolsep}{5pt}
\begin{tabular}{lrrrrr}
\toprule
Input whose answer is known & Evaluator's LPIPS accuracy & Two-way reading & State reading & Threshold 0.2 & Threshold 0.1 \\
\midrule
Simulator, current state & 0.99 & 1.00 & 0.95 & 0.96 & 0.95 \\
Simulator, outdated state & 0.05 to 0.16 & 0.08 & 0.01 & 0.01 & 0.01 \\
Simulator, another clip & 0.00 & 0.37 & 0.00 & 0.00 & 0.00 \\
Simulator, tokens erased & 0.02 to 0.17 & 0.34 to 0.37 & 0.00 & 0.00 & 0.00 \\
Even mixture, a ghost & 0.98 & 0.70 & 0.43 & 0.19 & 0.02 \\
FAR, current views alone & 0.80 & 0.83 & 0.68 & 0.65 & 0.63 \\
FAR, cut before change & 0.31 & 0.10 & 0.01 & 0.01 & 0.01 \\
FAR, view withheld & 0.62 & 0.77 & 0.06 & 0.01 & 0.00 \\
\bottomrule
\end{tabular}
\end{table*}

\begin{table*}[tp]
\centering
\caption{Decoded renders read in AlexNet LPIPS (\(\kappa=0.35\)) and RGB pixels (\(\kappa=0.45\)) at seed 1 on the 452-step sample; token-deletion rows match \cref{tab:samelog}. Nearer current is the share closer in LPIPS to the ground-truth frame than to the stale frame, Seen and Unseen split the 352 reverted steps by whether the target pose stored the current state, and † reads the simulator.\label{tab:s-features}}
\small\setlength{\tabcolsep}{5pt}
\begin{tabular}{lrrrrrrrr}
\toprule
 & \multicolumn{5}{c}{Reverted} & \multicolumn{1}{c}{Seen} & \multicolumn{1}{c}{Unseen} & \multicolumn{1}{c}{New-state} \\
\cmidrule(lr){2-6}\cmidrule(lr){7-7}\cmidrule(lr){8-8}\cmidrule(lr){9-9}
Candidate & \begin{tabular}[b]{@{}r@{}}LPIPS\\current↑\end{tabular} & \begin{tabular}[b]{@{}r@{}}LPIPS\\outd.↓\end{tabular} & \begin{tabular}[b]{@{}r@{}}pixel\\current↑\end{tabular} & \begin{tabular}[b]{@{}r@{}}pixel\\outd.↓\end{tabular} & \begin{tabular}[b]{@{}r@{}}nearer\\current↑\end{tabular} & \begin{tabular}[b]{@{}r@{}}LPIPS\\current↑\end{tabular} & \begin{tabular}[b]{@{}r@{}}LPIPS\\current↑\end{tabular} & \begin{tabular}[b]{@{}r@{}}LPIPS\\current↑\end{tabular} \\
\midrule
FAR~\citep{kim2026learning} & 0.32 & 0.30 & 0.42 & 0.38 & 0.52 & 0.45 & 0.12 & 0.44 \\
\rowcolor{oursrow}\textbf{\ours{} (ours)} & \textbf{0.55} & 0.05 & \textbf{0.71} & 0.10 & 0.83 & \textbf{0.75} & \textbf{0.22} & \textbf{0.46} \\
\ours{} per frame & \underline{0.53} & 0.09 & \underline{0.66} & 0.16 & 0.77 & \underline{0.73} & \underline{0.19} & \textbf{0.46} \\
State filtering~\citep{yang2026oneira} & 0.37 & \textbf{0.01} & 0.59 & \textbf{0.02} & 0.86 & 0.58 & 0.02 & 0.38 \\
Oracle token deletion†~\citep{yang2026permavid} & 0.35 & 0.11 & 0.48 & 0.14 & 0.67 & 0.53 & 0.05 & 0.42 \\
Frame deletion by pose~\citep{wu2026echo} & 0.20 & 0.21 & 0.38 & 0.31 & 0.57 & 0.26 & 0.11 & 0.41 \\
Token deletion, \ours{}'s log, capture time & 0.36 & \textbf{0.01} & 0.60 & \underline{0.03} & \textbf{0.88} & 0.57 & 0.02 & 0.37 \\
Token deletion, \ours{}'s log, valid time & 0.44 & 0.02 & 0.61 & \underline{0.03} & \underline{0.87} & 0.69 & 0.02 & 0.38 \\
\midrule
ground-truth frame† & 1.00 & 0.00 & 1.00 & 0.00 & 1.00 & 1.00 & 1.00 & 1.00 \\
stale frame† & 0.00 & 1.00 & 0.00 & 1.00 & 0.00 & 0.00 & 0.00 & 0.00 \\
even mixture† & 0.07 & 0.40 & 0.00 & 0.00 & 0.28 & 0.06 & 0.08 & 0.31 \\
\bottomrule
\end{tabular}
\end{table*}

A second reading leaves FAR's latent space (\cref{tab:s-features}). It decodes the renders of eight memories on a sample of 452 steps fixed before any render was read, all 352 reverted steps and 100 new-state steps drawn once, crops the changed region padded by 0.35 of its side, and measures each render against the ground-truth frame and the stale frame with LPIPS on AlexNet features and with the mean absolute RGB difference over the changed pixels. The three-way rule of the latent reading applies in each space, with \(\kappa\) the largest value on a grid of 0.05 at which at most 0.10 of the even mixtures of the reverted steps read current, 0.35 for LPIPS and 0.45 for pixels, fixed by the known inputs alone. \ours{} reads 0.55 and 0.71 current on reverted steps against FAR's 0.32 and 0.42, and it leads every other memory in both spaces, in the order of the latent reading. Token deletion with \ours{}'s own log reads 0.36 and 0.60 under capture time and 0.44 and 0.61 under valid time, so moving the outdated tokens into the past beats hiding them by 0.11 and 0.10 here, against 0.06 under the latent reading. The two-way reading, which asks only which reference a render lies nearer, places state filtering at 0.86, the token-deletion rows at 0.88 and 0.87 and \ours{} at 0.83 in LPIPS: these memories withhold every outdated view or token, and many of their renders lie nearer the current state without coming close to it, the renders that the quality floor counts as low. LPIPS agrees with the latent reading on 0.75 of the renders that the latent reading assigns to a state and pixels on 0.95. Where only other poses saw the current state, pixels read more renders as current than the latent reading does for every memory, and LPIPS does so for the memories that leave outdated tokens in view, \ours{} 0.34 and 0.22.

\section{Further results}\label{sec:results}

Each row of this appendix is paired with its reference at the same seed on the clips that both ran. Shares are of evaluated steps under the state metric unless a table names the evaluator, and Clips +/− counts the clips with a higher and a lower accuracy on the first step set of a table. \ours{} names the method of \cref{sec:redate}, with timestamps per token, and \ours{} per frame its variant with one timestamp per frame (\cref{sec:retiming}).

\Cref{tab:ablation} builds \ours{} from FAR one part at a time at seed 1, the build that \cref{sec:ablation} describes. Re-dating only the frames at the target pose that agree with the ground-truth frame gives 0.45 against the tracker's 0.57, since the tracker also re-dates frames from other poses that hold the restored region.

\begin{table*}[tp]
\centering
\caption{\ours{} and its parts at seed 1. Frames counts the re-dated frames added to the context and Post marks post-action retrieval; re-dating from ground truth† re-dates the frames at the target pose that agree with the ground-truth frame. Acc. and New give accuracy on the 352 reverted and 2,968 new-state steps, and Outd. and Low q. the outdated and low-quality shares on reverted steps; bold and underline mark the best and second best without †.\label{tab:ablation}}
\small\setlength{\tabcolsep}{6pt}
\begin{tabular}{lrcrrrr}
\toprule
Run & Frames & Post & Acc.↑ & Outd.↓ & Low q.↓ & New↑ \\
\midrule
\rowcolor{oursrow}\textbf{\ours{} (ours)} & 2 & ✓ & \textbf{0.62} & \textbf{0.09} & \textbf{0.09} & 0.44 \\
\ours{} per frame & 2 & ✓ & \underline{0.58} & \underline{0.15} & \underline{0.10} & \textbf{0.46} \\
FAR + valid time + two recalled frames & 2 &  & 0.57 & 0.17 & 0.11 & \textbf{0.46} \\
Re-dating from ground truth† & 2 &  & 0.45 & 0.24 & 0.17 & 0.44 \\
FAR + valid time & 0 &  & 0.55 & 0.19 & 0.11 & \textbf{0.46} \\
one time for every frame & 2 & ✓ & 0.21 & 0.44 & 0.26 & 0.33 \\
new timestamps on random frames & 0 &  & 0.30 & 0.32 & 0.23 & 0.41 \\
FAR + post-action retrieval & 0 & ✓ & 0.41 & 0.24 & 0.18 & 0.43 \\
FAR~\citep{kim2026learning} & 0 &  & 0.36 & 0.27 & 0.22 & 0.43 \\
\bottomrule
\end{tabular}
\end{table*}

\subsection{Where the generated frames lie}

\begin{figure*}[tp]
\centering
\includegraphics[width=\linewidth,height=0.82\textheight,keepaspectratio]{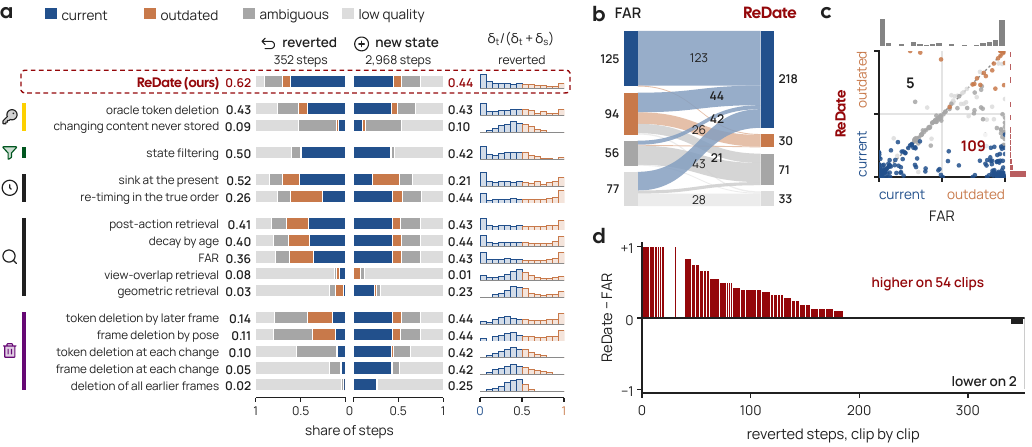}
\caption{(a) Outcome shares and latent-plane positions of memory designs on reverted and new-state steps at seed 1. (b, c) Per-step outcomes and distances under FAR and \ours{} on the 352 reverted steps. (d) Per-clip accuracy difference between \ours{} and FAR.\label{fig:results}}
\end{figure*}

\begin{figure*}[tp]
\centering
\includegraphics[width=\linewidth,height=0.82\textheight,keepaspectratio]{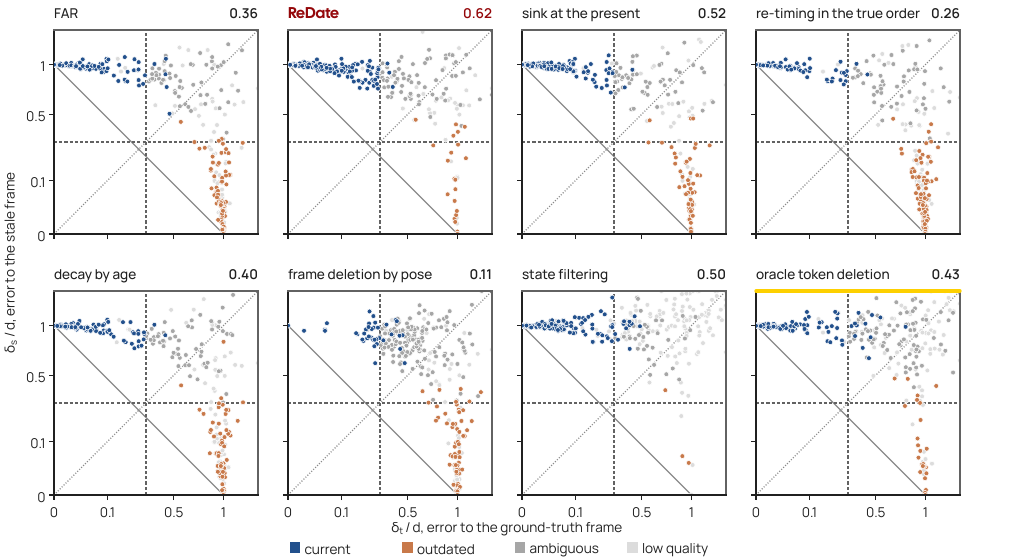}
\caption{Generated frames of the 352 reverted steps at seed 1 under eight memories, on square-root axes where mixtures of the two frames form the line from \((0,1)\) to \((1,0)\). Dashed lines mark \(\kappa=0.295\), numbers give accuracy, and yellow marks a memory that reads the simulator.\label{fig:plane}}
\end{figure*}

\Cref{fig:plane} places every reverted step of \cref{tab:main} by the errors of its generated frame to the ground-truth and the stale frame. FAR's outdated renders sit at the stale frame, \((1,0)\), as copies of it, and \ours{}'s renders gather at the ground-truth frame, \((0,1)\). Frame deletion by pose spreads its renders over the ambiguous middle, where the region shows neither state, and state filtering leaves 0.39 of its renders below the quality floor against 0.22 for FAR, the cost that \cref{thm:frame} names for a memory that withholds whole frames.

\subsection{Seeds, clip sets and readings}

\begin{figure*}[tp]
\centering
\includegraphics[width=\linewidth,height=0.82\textheight,keepaspectratio]{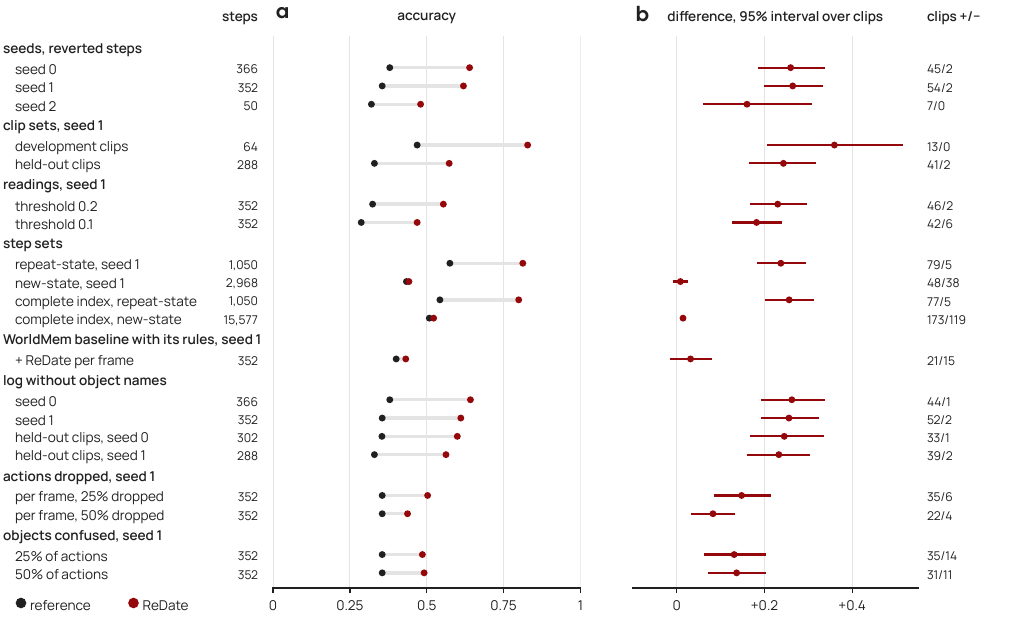}
\caption{(a) Accuracy of \ours{} and its reference (FAR unless noted) on matched steps. (b) Accuracy difference with clip-resampling 95\% intervals and clip win/loss counts.\label{fig:robust}}
\end{figure*}

\Cref{fig:robust} pairs \ours{} with its reference across seeds, clip sets, readings, step sets, a second memory system, a log without object names and a lossy action log, and the tables below give the four outcomes behind each pair. \ours{}'s gain over FAR on reverted steps holds at every seed and on the held-out clips, at both fixed thresholds and on the complete index, and new-state steps stay at FAR's level. Under the WorldMem baseline with target-pose frames and post-action retrieval, the generator that follows timestamps least firmly, \ours{} per frame adds +0.031. With the log without object names, \ours{} is higher than FAR on 52 clips and lower on 2 at seed 1, and the two logs give the same reading on 341 of the 352 reverted steps. Letting FAR's retriever read the new timestamps as well moves accuracy on reverted steps by at most 0.007 at seed 0 and on either clip set (\cref{tab:s-heldout,tab:s-seeds}), so \ours{}'s timestamps act through the generator.

\Cref{tab:s-samelog-seed0} repeats the comparison of \cref{tab:samelog} at seed 0: moving the outdated tokens again leads hiding them, at a higher PSNR. A hole also filters what the quality floor reads: among renders above the floor at seed 1, token deletion under valid time shows the current state on 0.88 and \ours{} on 0.68, and hiding as many tokens half the grid away clears \ours{}'s outdated renders in the same way (\cref{sec:ablation}).

\begin{table*}[tp]
\centering
\caption{Re-dating against token deletion with the same log on the 366 reverted steps of seed 0, with the columns and marks of \cref{tab:samelog}.\label{tab:s-samelog-seed0}}
\small\setlength{\tabcolsep}{4pt}
\begin{tabular}{llrrrrrrr}
\toprule
Recalled frames & Outdated tokens & Acc.↑ & At 0.2↑ & LPIPS acc.↑ & PSNR↑ & Outd.↓ & Low q.↓ & New↑ \\
\midrule
capture time & hidden & 0.56 & 0.54 & 0.72 & 34.5 & \textbf{0.01} & 0.37 & 0.40 \\
valid time & hidden & \underline{0.59} & \underline{0.58} & \underline{0.73} & \underline{35.4} & \textbf{0.01} & \underline{0.34} & \underline{0.41} \\
\rowcolor{oursrow}\textbf{valid time, \ours{} (ours)} & moved before the oldest frame & \textbf{0.64} & \textbf{0.61} & \textbf{0.87} & \textbf{39.6} & 0.06 & \textbf{0.12} & \textbf{0.44} \\
\bottomrule
\end{tabular}
\end{table*}

\begin{table*}[tp]
\centering
\caption{Reverted steps of the 52 development clips (inside the first 200 clips) and the 197 held-out clips, paired with FAR at seed 1 unless noted. Pooled rows over seeds 0 and 1 count clips once per seed against FAR at 0.342.\label{tab:s-heldout}}
\small\setlength{\tabcolsep}{4pt}
\begin{tabular}{lrrrrrrr}
\toprule
 &  &  &  &  &  & \multicolumn{2}{c}{Comparison} \\
\cmidrule(lr){7-8}
Row & Steps & Current & Outd. & Ambig. & Low q. & Difference & Clips +/− \\
\midrule
\rowcolor{oursrow}\textbf{\ours{} (ours)}, development clips & 64 & 0.828 & 0.062 & 0.078 & 0.031 & +0.359 & 13/0 \\
FAR + valid time & 64 & 0.750 & 0.141 & 0.047 & 0.062 & +0.281 & 10/0 \\
+ two recalled frames & 64 & 0.797 & 0.078 & 0.062 & 0.062 & +0.328 & 11/0 \\
+ post-action retrieval, \ours{} per frame & 64 & 0.812 & 0.078 & 0.062 & 0.047 & +0.344 & 12/0 \\
FAR + valid time, read by the retriever too & 64 & 0.750 & 0.125 & 0.062 & 0.062 & +0.281 & 10/0 \\
\midrule
\rowcolor{oursrow}\textbf{\ours{} (ours)}, held-out clips & 288 & 0.573 & 0.090 & 0.229 & 0.108 & +0.243 & 41/2 \\
FAR + valid time & 288 & 0.507 & 0.194 & 0.174 & 0.125 & +0.177 & 31/3 \\
+ two recalled frames & 288 & 0.524 & 0.184 & 0.174 & 0.118 & +0.194 & 33/2 \\
+ post-action retrieval, \ours{} per frame & 288 & 0.528 & 0.167 & 0.194 & 0.111 & +0.198 & 36/4 \\
FAR + valid time, read by the retriever too & 288 & 0.514 & 0.191 & 0.170 & 0.125 & +0.184 & 32/3 \\
\ours{}, log without object names, held-out clips & 288 & 0.562 & 0.097 & 0.219 & 0.122 & +0.233 & 39/2 \\
\midrule
\rowcolor{oursrow}\textbf{\ours{} (ours)}, held-out clips, seed 0 & 302 & 0.596 & 0.063 & 0.205 & 0.136 & +0.242 & 34/2 \\
\rowcolor{oursrow}\textbf{\ours{} (ours)}, held-out clips, seeds 0 and 1 & 590 & 0.585 & 0.076 & 0.217 & 0.122 & +0.242 & 75/4 \\
\ours{}, log without object names, held-out clips, seed 0 & 302 & 0.599 & 0.083 & 0.185 & 0.132 & +0.245 & 33/1 \\
\ours{}, log without object names, held-out clips, seeds 0 and 1 & 590 & 0.581 & 0.090 & 0.202 & 0.127 & +0.239 & 72/3 \\
\bottomrule
\end{tabular}
\end{table*}

\begin{table*}[tp]
\centering
\caption{Memory designs at seeds 0 and 2, and at seed 1 under the fixed threshold 0.2 (no quality floor; Low q. gives the ambiguous share), paired with FAR unless noted. † reads the simulator; bold and underline mark the best and second best per block without †.\label{tab:s-seeds}}
\small\setlength{\tabcolsep}{4pt}
\begin{tabularx}{\linewidth}{>{\raggedright\arraybackslash\hsize=1.000\hsize}Xrrrrrrrrrr}
\toprule
 & \multicolumn{3}{c}{Reverted} & \multicolumn{3}{c}{Repeat-state} & \multicolumn{3}{c}{New-state} &  \\
\cmidrule(lr){2-4}\cmidrule(lr){5-7}\cmidrule(lr){8-10}
Memory & Current↑ & Outd.↓ & Low~q.↓ & Current↑ & Outd.↓ & Low~q.↓ & Current↑ & Outd.↓ & Low~q.↓ & \begin{tabular}[b]{@{}r@{}}Clips\\+/−\end{tabular} \\
\midrule
\multicolumn{11}{l}{\emph{Seed 0, 249 clips}} \\
Latest 12 frames & 0.003 & \textbf{0.000} & 0.995 & 0.008 & \textbf{0.000} & 0.991 & 0.001 & \textbf{0.000} & 0.997 & 1/75 \\
FAR~\citep{kim2026learning} & 0.380 & 0.284 & 0.186 & 0.543 & 0.209 & 0.160 & 0.416 & 0.101 & 0.267 & ref. \\
View-overlap retrieval~\citep{yu2025context} & 0.060 & 0.019 & 0.896 & 0.079 & 0.010 & 0.872 & 0.007 & 0.067 & 0.869 & 6/70 \\
Frame deletion by pose~\citep{wu2026echo} & 0.109 & 0.254 & 0.219 & 0.477 & 0.112 & 0.181 & 0.433 & 0.079 & 0.261 & 9/58 \\
Token deletion by later frame~\citep{wu2026echo} & 0.126 & 0.276 & 0.210 & 0.465 & 0.125 & 0.177 & 0.421 & 0.086 & 0.264 & 2/51 \\
Frame deletion at each change†~\citep{yang2026permavid} & 0.052 & 0.008 & 0.806 & 0.405 & 0.032 & 0.437 & 0.414 & 0.008 & 0.502 & 5/68 \\
Changing content never stored†~\citep{wang2026latent} & 0.085 & 0.011 & 0.464 & 0.158 & 0.027 & 0.416 & 0.101 & 0.014 & 0.487 & 11/68 \\
State filtering~\citep{yang2026oneira} & 0.516 & 0.014 & 0.377 & 0.699 & 0.008 & 0.224 & 0.412 & \underline{0.004} & 0.536 & 37/17 \\
FAR + post-action retrieval & 0.423 & 0.230 & 0.180 & 0.636 & 0.135 & 0.136 & 0.425 & 0.096 & 0.259 & 16/7 \\
\ours{} per frame & 0.617 & 0.126 & \textbf{0.109} & 0.755 & 0.078 & 0.090 & \textbf{0.451} & 0.080 & 0.245 & 41/2 \\
\rowcolor{oursrow}\textbf{\ours{} (ours)} & \underline{0.639} & 0.060 & 0.120 & \textbf{0.799} & 0.031 & \textbf{0.081} & 0.438 & 0.089 & 0.253 & 45/2 \\
\ours{}, log without object names & \textbf{0.642} & 0.077 & \underline{0.117} & \underline{0.794} & 0.039 & \underline{0.086} & 0.435 & 0.095 & 0.253 & 44/1 \\
Token deletion, \ours{}'s log, capture time & 0.563 & \underline{0.005} & 0.374 & 0.733 & \underline{0.002} & 0.210 & 0.395 & 0.005 & 0.561 & 41/16 \\
Token deletion, \ours{}'s log, valid time, paired with \ours{} & 0.590 & 0.008 & 0.339 & 0.734 & 0.008 & 0.196 & 0.412 & 0.008 & 0.547 & 6/22 \\
Oracle frame deletion†~\citep{yang2026oneira} & 0.478 & 0.049 & 0.322 & 0.583 & 0.050 & 0.227 & 0.416 & 0.019 & 0.453 & 28/14 \\
Re-dating from ground truth† & 0.437 & 0.235 & 0.183 & 0.614 & 0.171 & 0.136 & 0.421 & 0.100 & 0.264 & 16/2 \\
FAR + valid time & 0.596 & 0.126 & 0.131 & 0.730 & 0.092 & 0.100 & 0.450 & 0.087 & \underline{0.243} & 35/1 \\
+ two recalled frames & 0.601 & 0.131 & 0.123 & 0.732 & 0.096 & 0.095 & \textbf{0.451} & 0.086 & \textbf{0.240} & 38/2 \\
FAR + valid time, read by the retriever too & 0.593 & 0.134 & 0.126 & 0.722 & 0.098 & 0.101 & 0.447 & 0.090 & 0.244 & 35/1 \\
Recalled frames, capture time & 0.383 & 0.279 & 0.202 & 0.554 & 0.201 & 0.162 & 0.423 & 0.097 & 0.263 & 3/3 \\
\midrule
\multicolumn{11}{l}{\emph{Seed 2, 246 clips}} \\
FAR~\citep{kim2026learning} & 0.320 & 0.260 & 0.200 & 0.557 & 0.193 & 0.164 & 0.523 & 0.074 & 0.201 & ref. \\
\rowcolor{oursrow}\textbf{\ours{} (ours)} & \underline{0.480} & \underline{0.140} & \underline{0.160} & \textbf{0.736} & \underline{0.064} & \textbf{0.121} & \underline{0.539} & 0.069 & \underline{0.184} & 7/0 \\
\ours{} per frame & \textbf{0.500} & \underline{0.140} & \textbf{0.140} & \underline{0.721} & 0.071 & \textbf{0.121} & \textbf{0.551} & 0.068 & \textbf{0.167} & 8/0 \\
Recalled frames, capture time & 0.300 & 0.280 & 0.200 & 0.543 & 0.200 & 0.164 & 0.526 & 0.074 & 0.194 & 0/1 \\
State filtering~\citep{yang2026oneira} & 0.320 & \textbf{0.000} & 0.460 & 0.593 & \textbf{0.000} & 0.286 & 0.496 & \textbf{0.004} & 0.451 & 3/3 \\
Frame deletion by pose~\citep{wu2026echo} & 0.120 & 0.260 & 0.180 & 0.514 & 0.121 & 0.157 & 0.529 & \underline{0.061} & 0.198 & 1/9 \\
Frame deletion at each change†~\citep{yang2026permavid} & 0.020 & 0.000 & 0.800 & 0.329 & 0.029 & 0.464 & 0.503 & 0.005 & 0.435 & 0/10 \\
\midrule
\multicolumn{11}{l}{\emph{Seed 1 at the fixed threshold 0.2, 249 clips}} \\
FAR~\citep{kim2026learning} & 0.324 & 0.358 & \underline{0.318} & 0.558 & 0.265 & \underline{0.177} & 0.414 & 0.153 & \textbf{0.433} & ref. \\
FAR + post-action retrieval & 0.378 & 0.290 & 0.332 & 0.637 & 0.175 & 0.188 & 0.419 & 0.124 & 0.457 & 17/5 \\
\ours{} per frame & 0.531 & 0.136 & 0.332 & 0.748 & 0.085 & \textbf{0.168} & \textbf{0.437} & \underline{0.113} & 0.451 & 42/4 \\
\rowcolor{oursrow}\textbf{\ours{} (ours)} & \textbf{0.554} & \underline{0.074} & 0.372 & \textbf{0.780} & \underline{0.035} & 0.185 & 0.422 & 0.125 & 0.453 & 46/2 \\
\ours{}, log without object names & \underline{0.545} & 0.091 & 0.364 & \underline{0.770} & 0.049 & 0.181 & 0.421 & 0.131 & 0.448 & 45/3 \\
Recalled frames, capture time & 0.301 & 0.386 & \textbf{0.312} & 0.535 & 0.287 & 0.178 & 0.416 & 0.150 & \underline{0.434} & 1/7 \\
State filtering~\citep{yang2026oneira} & 0.455 & \textbf{0.011} & 0.534 & 0.676 & \textbf{0.007} & 0.317 & 0.418 & \textbf{0.010} & 0.572 & 33/8 \\
Oracle token deletion†~\citep{yang2026permavid} & 0.381 & 0.102 & 0.517 & 0.572 & 0.141 & 0.287 & 0.407 & 0.076 & 0.517 & 21/13 \\
Frame deletion by pose~\citep{wu2026echo} & 0.028 & 0.315 & 0.656 & 0.474 & 0.117 & 0.409 & \underline{0.423} & 0.118 & 0.459 & 1/62 \\
Token deletion at each change†~\citep{yang2026permavid} & 0.014 & 0.000 & 0.986 & 0.407 & 0.009 & 0.585 & 0.408 & 0.011 & 0.581 & 2/67 \\
\bottomrule
\end{tabularx}
\end{table*}

The gain holds across the rooms of the benchmark (\cref{tab:s-rooms}). The 141 clips with reverted steps come from 20 AI2-THOR rooms, and \ours{}'s accuracy on reverted steps is higher than FAR's in 15 of them, equal in 4 and lower in 1, against 11, 5 and 4 for state filtering and 14, 4 and 2 for token deletion with \ours{}'s log under valid time. Kitchens rarely store the current state at the target pose, on 6 of their 31 reverted steps, and on those 6 \ours{} reads 0.67 against FAR's 0.33.

\begin{table*}[tp]
\centering
\caption{Reverted-step accuracy at seed 1 by room type on all steps and on seen steps. Token deletion reads \ours{}'s log under valid time (\cref{tab:samelog}).\label{tab:s-rooms}}
\small\setlength{\tabcolsep}{4pt}
\begin{tabular}{lrrrrrrrrrr}
\toprule
Room type & Rooms & Clips & Reverted steps & FAR & \ours{} & State filtering & Hidden, valid time & Seen steps & FAR & \ours{} \\
\midrule
living rooms & 5 & 47 & 100 & 0.38 & 0.78 & 0.70 & 0.65 & 78 & 0.46 & 0.96 \\
bedrooms & 5 & 26 & 102 & 0.33 & 0.61 & 0.45 & 0.59 & 61 & 0.56 & 0.98 \\
bathrooms & 6 & 59 & 119 & 0.40 & 0.61 & 0.45 & 0.55 & 74 & 0.62 & 0.89 \\
kitchens & 4 & 9 & 31 & 0.16 & 0.16 & 0.16 & 0.23 & 6 & 0.33 & 0.67 \\
\bottomrule
\end{tabular}
\end{table*}

Paired on the same steps, \ours{} is higher than every strong memory (\cref{tab:s-paired}). On all 352 reverted steps of seed 1 and on the 288 of the held-out clips, its current share exceeds that of FAR, of its per-frame variant, of token deletion with its own log under either timestamp, of state filtering, of the sink and of oracle token deletion, and every interval from resampling clips lies above zero. Against its strongest rival, token deletion with its own log under valid time, it is higher on 16 clips and lower on 4, and on the held-out clips by 0.042 with 11 clips higher and 4 lower. The sink at the present pays for its reverted steps on new-state steps, where it falls from FAR's 0.43 to 0.21, below FAR on 170 clips and above on 3.

Per-token timestamps improve over per-frame timestamps specifically on steps whose target-pose views hold outdated tokens (\texttt{far\_token\_split.py}). \ours{}'s saved context at seed 1 records, for every view of the target pose, its date of capture, the date it is read at and the date of its outdated tokens. On the 312 reverted steps (in 132 clips) where target-pose views hold outdated tokens and no view is re-dated, \ours{} gives 0.641 current and 0.061 outdated against 0.599 and 0.141 for \ours{} per frame. On the 18 crossing steps (\cref{thm:frame}(ii)), where a view is re-dated for one restored region and holds outdated tokens of another, the two variants read 0.278 and 0.222 current and 0.444 and 0.333 outdated. On the 22 steps without outdated tokens, the two variants assign identical timestamps and both read 0.591 current.

\begin{table*}[tp]
\centering
\caption{\ours{} paired with each strong memory on all 352 reverted steps of seed 1 (141 clips, \ours{} at 0.619) and on the 288 held-out steps (111 clips, \ours{} at 0.573). Difference is \ours{}'s accuracy minus the baseline's, interval is the 95\% interval over 2,000 clip resamples, and † reads the simulator.\label{tab:s-paired}}
\footnotesize\setlength{\tabcolsep}{4pt}
\begin{tabular}{lrrrrrrrr}
\toprule
 & \multicolumn{4}{c}{All} & \multicolumn{4}{c}{Held-out} \\
\cmidrule(lr){2-5}\cmidrule(lr){6-9}
Memory & memory & difference & interval & clips +/− & memory & difference & interval & clips +/− \\
\midrule
FAR~\citep{kim2026learning} & 0.355 & +0.264 & 0.199 to 0.334 & 54/2 & 0.330 & +0.243 & 0.164 to 0.318 & 41/2 \\
\ours{} per frame & 0.580 & +0.040 & 0.013 to 0.071 & 11/1 & 0.528 & +0.045 & 0.016 to 0.084 & 9/0 \\
Token deletion, \ours{}'s log, valid time & 0.562 & +0.057 & 0.022 to 0.096 & 16/4 & 0.531 & +0.042 & 0.006 to 0.083 & 11/4 \\
Token deletion, \ours{}'s log, capture time & 0.523 & +0.097 & 0.058 to 0.141 & 27/3 & 0.472 & +0.101 & 0.056 to 0.146 & 23/2 \\
State filtering~\citep{yang2026oneira} & 0.497 & +0.122 & 0.071 to 0.179 & 31/4 & 0.434 & +0.139 & 0.083 to 0.205 & 27/2 \\
State filtering, \ours{}'s context & 0.526 & +0.094 & 0.049 to 0.145 & 24/3 & 0.476 & +0.097 & 0.049 to 0.154 & 19/1 \\
Sink at the present~\citep{yi2025deep} & 0.520 & +0.099 & 0.049 to 0.156 & 33/8 & 0.465 & +0.108 & 0.052 to 0.169 & 28/5 \\
Sink at the present, \ours{}'s context & 0.574 & +0.046 & 0.005 to 0.087 & 23/8 & 0.514 & +0.059 & 0.015 to 0.107 & 21/5 \\
Oracle token deletion†~\citep{yang2026permavid} & 0.429 & +0.190 & 0.119 to 0.269 & 46/7 & 0.410 & +0.163 & 0.086 to 0.246 & 34/7 \\
\bottomrule
\end{tabular}
\end{table*}

Under the retrieval baselines of FAR's release, adding frames from the same camera and post-action retrieval lowers the low-quality share on repeat-state steps from 0.70 and 0.65 to 0.20 and 0.18 (\cref{tab:s-systems}). Among readable steps the accuracy is 0.77 and 0.76 for the WorldMem baseline and 0.76 and 0.74 for the FAR generator, and on subset B the low-quality share falls from 0.90 to 0.61.

\begin{table*}[tp]
\centering
\caption{Three retrieval and generator systems at seed 1. Each geometric baseline is paired with FAR, each recall-augmented row with its geometric baseline, and each \ours{} row with FAR or its recall-augmented baseline.\label{tab:s-systems}}
\footnotesize\setlength{\tabcolsep}{4pt}
\begin{tabular}{lrrrrrrrrrr}
\toprule
 & \multicolumn{3}{c}{Reverted} & \multicolumn{3}{c}{Repeat-state} & \multicolumn{3}{c}{New-state} &  \\
\cmidrule(lr){2-4}\cmidrule(lr){5-7}\cmidrule(lr){8-10}
Memory & Current & Outd. & Low~q. & Current & Outd. & Low~q. & Current & Outd. & Low~q. & Clips +/− \\
\midrule
FAR~\citep{kim2026learning} & 0.355 & 0.267 & 0.219 & 0.575 & 0.184 & 0.162 & 0.434 & 0.103 & 0.252 & ref. \\
FAR with \ours{} per frame & 0.580 & 0.151 & 0.099 & 0.769 & 0.084 & 0.071 & 0.458 & 0.083 & 0.240 & 48/4 \\
FAR with \ours{} & 0.619 & 0.085 & 0.094 & 0.812 & 0.037 & 0.063 & 0.442 & 0.089 & 0.259 & 54/2 \\
FAR generator, geometric retrieval & 0.026 & 0.091 & 0.818 & 0.265 & 0.030 & 0.651 & 0.233 & 0.029 & 0.703 & 3/71 \\
with target-pose frames and post-action retrieval & 0.381 & 0.222 & 0.233 & 0.608 & 0.130 & 0.178 & 0.383 & 0.086 & 0.379 & 69/2 \\
with \ours{} per frame & 0.565 & 0.122 & 0.162 & 0.727 & 0.099 & 0.108 & 0.407 & 0.081 & 0.352 & 45/6 \\
with \ours{} & 0.597 & 0.088 & 0.159 & 0.750 & 0.077 & 0.103 & 0.396 & 0.102 & 0.344 & 46/6 \\
WorldMem baseline~\citep{xiao2025worldmem} & 0.028 & 0.062 & 0.861 & 0.232 & 0.022 & 0.700 & 0.194 & 0.022 & 0.735 & 2/71 \\
with target-pose frames and post-action retrieval & 0.401 & 0.074 & 0.295 & 0.602 & 0.044 & 0.205 & 0.283 & 0.066 & 0.436 & 68/1 \\
with \ours{} per frame & 0.432 & 0.060 & 0.278 & 0.577 & 0.021 & 0.224 & 0.289 & 0.064 & 0.418 & 21/15 \\
with \ours{} & 0.432 & 0.045 & 0.273 & 0.580 & 0.019 & 0.218 & 0.289 & 0.075 & 0.412 & 21/14 \\
\bottomrule
\end{tabular}
\end{table*}

Writing generated frames back into the memory bank changes FAR and \ours{} by at most 0.008 on reverted steps and leaves \ours{} 0.26 above FAR (\cref{tab:s-writeback}). Each generated frame enters behind the current frame under either its \emph{render time} (when it was drawn) or its \emph{evidence time} (the timestamp of the newest frame it was drawn from, minus one rollout step and half a frame).

\begin{table*}[tp]
\centering
\caption{Rollouts that write generated frames into the memory bank at seed 1, each paired with the same memory on real history; bold and underline mark the best and second best within each memory.\label{tab:s-writeback}}
\small\setlength{\tabcolsep}{6pt}
\begin{tabular}{lrrrrrr}
\toprule
 & \multicolumn{3}{c}{Reverted} & \multicolumn{1}{c}{Repeat-state} & \multicolumn{1}{c}{New-state} &  \\
\cmidrule(lr){2-4}\cmidrule(lr){5-5}\cmidrule(lr){6-6}
Memory bank & Current↑ & Outd.↓ & Low q.↓ & Current↑ & Current↑ & Clips +/− \\
\midrule
FAR, real history & 0.355 & 0.267 & 0.219 & 0.575 & 0.434 & ref. \\
FAR, generated frames at render time & 0.352 & 0.287 & 0.216 & 0.529 & 0.430 & 9/11 \\
FAR, generated frames at evidence time & 0.349 & 0.264 & 0.230 & 0.516 & 0.429 & 9/13 \\
\rowcolor{oursrow}\textbf{\ours{} (ours)}, real history & \textbf{0.619} & \underline{0.085} & \textbf{0.094} & \textbf{0.812} & 0.442 & ref. \\
\ours{}, generated frames at render time & \underline{0.611} & 0.088 & \underline{0.099} & \underline{0.800} & \underline{0.444} & 5/8 \\
\ours{}, generated frames at evidence time & \underline{0.611} & \textbf{0.082} & 0.102 & 0.799 & \textbf{0.445} & 4/7 \\
\bottomrule
\end{tabular}
\end{table*}

\begin{table*}[tp]
\centering
\caption{Complete index at seed 0 (1,281 of 1,309 clips have evaluated steps), paired with FAR on each step set. The 1,379 steps outside the repeat-state and new-state subsets either mix a repeat-state and a new-state region or have no attributed region change.\label{tab:s-complete}}
\small\setlength{\tabcolsep}{5pt}
\begin{tabular}{lrrrrrrrr}
\toprule
 &  &  &  &  &  &  & \multicolumn{2}{c}{Comparison} \\
\cmidrule(lr){8-9}
Row & Steps & Current & Outd. & Ambig. & Low~q. & LPIPS acc. & Difference & Clips +/− \\
\midrule
FAR, all steps & 18,006 & 0.493 & 0.102 & 0.211 & 0.194 & 0.761 & ref. & ref. \\
\ours{} per frame, all steps & 18,006 & 0.523 & 0.084 & 0.217 & 0.176 & 0.787 & +0.030 & 218/88 \\
\rowcolor{oursrow}\textbf{\ours{} (ours)}, all steps & 18,006 & 0.523 & 0.075 & 0.219 & 0.184 & 0.781 & +0.030 & 233/122 \\
FAR, clips outside the first 200 & 15,607 & 0.488 & 0.105 & 0.222 & 0.185 & 0.754 & ref. & ref. \\
\ours{} per frame, clips outside the first 200 & 15,607 & 0.510 & 0.087 & 0.228 & 0.175 & 0.777 & +0.022 & 167/79 \\
\rowcolor{oursrow}\textbf{\ours{} (ours)}, clips outside the first 200 & 15,607 & 0.509 & 0.078 & 0.231 & 0.182 & 0.771 & +0.021 & 178/107 \\
FAR, repeat-state steps & 1,050 & 0.543 & 0.209 & 0.089 & 0.160 & 0.738 & ref. & ref. \\
\ours{} per frame, repeat-state steps & 1,050 & 0.755 & 0.078 & 0.077 & 0.090 & 0.871 & +0.212 & 67/6 \\
\rowcolor{oursrow}\textbf{\ours{} (ours)}, repeat-state steps & 1,050 & 0.799 & 0.031 & 0.089 & 0.081 & 0.918 & +0.256 & 77/5 \\
FAR, new-state steps & 15,577 & 0.508 & 0.086 & 0.215 & 0.191 & 0.775 & ref. & ref. \\
\ours{} per frame, new-state steps & 15,577 & 0.527 & 0.076 & 0.221 & 0.176 & 0.795 & +0.019 & 163/81 \\
\rowcolor{oursrow}\textbf{\ours{} (ours)}, new-state steps & 15,577 & 0.522 & 0.070 & 0.222 & 0.186 & 0.785 & +0.014 & 173/119 \\
\bottomrule
\end{tabular}
\end{table*}

\subsection{Results under the benchmark's evaluator}\label{sec:reported}

This subsection reads every run with the benchmark's evaluator and image measures alone. The released systems reproduce their reported values (\cref{tab:s-reported}): the WorldMem baseline and FAR match their surface accuracies within 0.001, and the temporal baseline, which conditions on the latest frames alone, gives 0.129 against the reported 0.057.

\ours{} improves both surface and container accuracy over FAR on the complete index (\cref{tab:s-recall}), and restricting to the 1,109 clips outside subsets A and B shifts each row by at most 0.006. A stored frame at the exact target pose exists on 0.90 of rollout steps, and recalling the two oldest target-pose frames gives 0.591 against 0.603 for all target-pose frames on the WorldMem baseline and 0.785 against 0.788 on FAR. No deletion-based rule improved FAR on both subsets under the LPIPS evaluator, and adding frame deletion by pose to the WorldMem baseline leaves every clip unchanged, since its geometric retriever never selected a deleted frame.

\begin{table*}[tp]
\centering
\caption{Published benchmark values alongside our reruns (FAR room values from its project page, corridor values from its Section 4.3).\label{tab:s-reported}}
\small\setlength{\tabcolsep}{4pt}
\begin{tabularx}{\linewidth}{ll>{\raggedright\arraybackslash\hsize=1.198\hsize}Xl>{\raggedright\arraybackslash\hsize=0.802\hsize}X}
\toprule
Benchmark & Measure & System & Reported & Our rerun \\
\midrule
FAR's changing rooms, 1,309 clips & surface accuracy & Temporal baseline~\citep{kim2026learning} & 0.057 & 0.129 \\
 &  & WorldMem baseline~\citep{xiao2025worldmem} & 0.421 & 0.420 \\
 &  & FAR~\citep{kim2026learning} & 0.802 & 0.802 \\
 &  & \ours{} (ours) &  & 0.824 \\
 & container accuracy & FAR~\citep{kim2026learning} & 0.757 & 0.760 \\
 &  & \ours{} (ours) &  & 0.817 \\
FAR's corridor with a second agent & trajectory accuracy & WorldMem baseline~\citep{xiao2025worldmem}, FAR~\citep{kim2026learning} with time and pose, FAR with the agent cue & 0.324, 0.671, 0.946 & 0.953 for FAR with timestamps kept, on the timestamp probe \\
\bottomrule
\end{tabularx}
\end{table*}

\begin{table*}[tp]
\centering
\caption{Recall rules under the benchmark's evaluator at seed 0. Container accuracy pools the 1,020 container steps of the complete index, and the released analysis drops the 0.8~s before a container opens and averages over clips (\cref{sec:protocol}). † reads the simulator.\label{tab:s-recall}}
\small\setlength{\tabcolsep}{6pt}
\begin{tabular}{llrrr}
\toprule
Memory & Clip set & Surface & Container & Clips +/− \\
\midrule
FAR~\citep{kim2026learning} & complete index & 0.802 & 0.705 & ref. \\
FAR + post-action retrieval & complete index & 0.817 & 0.705 & 259/179 \\
\rowcolor{oursrow}\textbf{\ours{} (ours)} & complete index & 0.824 & 0.754 & 312/214 \\
\midrule
FAR, released analysis & complete index &  & 0.760 &  \\
\ours{}, released analysis & complete index & 0.824 & 0.817 &  \\
\midrule
FAR + post-action retrieval & subset A & 0.829 &  &  \\
Oracle recalling the true post-change frame† & subset A & 0.821 &  &  \\
Frames taken before the interactions & subset A & 0.791 &  &  \\
Context of 14 frames, no rule & subset A & 0.812 &  &  \\
\bottomrule
\end{tabular}
\end{table*}

\subsection{Timestamps steer the generator}

With frames, weights and noise fixed, timestamp conditions on subsets A and B at seed 0 steer FAR (\cref{fig:timestamps,tab:s-conditions}). Of the 754 steps of subset A that show the current state with timestamps kept, reversed timestamps send 439 to the outdated state, 222 below the floor and 17 to the ambiguous state, and leave 76, so 95 clips lose accuracy and none gains; exchanging context slots moves no share by more than 0.003. On the 502 steps of subset A with one change, the current share among the steps that show either state falls from 0.88 with timestamps kept to 0.51 at one time and 0.02 reversed, and to 0.86, 0.43 and 0.01 at the fixed threshold 0.2, and reversal moves the current and outdated shares of all 502 steps from 0.689 and 0.096 to 0.012 and 0.633. Among the steps of subsets A and B that show either state, gaps times 4 keep the current state at 0.82 and 0.90 against 0.81 with timestamps kept and send 0.62 and 0.67 of the renders to low quality, and gaps times 0.1 give 0.45 and 0.52, and 0.81 on the steps with one change.

The benchmark's own LPIPS test follows the timestamps as well (\cref{tab:s-conditions}). Compressing the gaps tenfold in the true order gives about the accuracy of one time, and a fourfold stretch raises whole-image LPIPS from 0.018 to 0.078 on subset A, so a timestamp pattern outside training lowers the LPIPS accuracy about as much as removing the order does. Under plain geometric retrieval with no recall rule, hiding the timestamps raises the LPIPS accuracy from 0.444 to 0.504 for the FAR generator and from 0.455 to 0.497 for the WorldMem baseline. Deleting the 212 outdated frames per clip that oracle frame deletion removes takes whole-image LPIPS from 0.018 to 0.049, and deleting 184 random frames gives 0.025.

\begin{figure*}[tp]
\centering
\includegraphics[width=\linewidth,height=0.82\textheight,keepaspectratio]{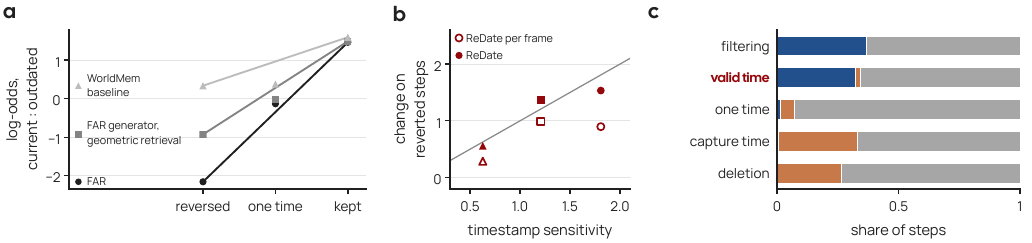}
\caption{(a) Log-odds of the current against the outdated state on subset A at seed 0 under reversed, one-time and kept timestamps. (b) Change of the log-odds on the 352 reverted steps at seed 1 from each context under capture time to \ours{} per frame and \ours{}, against the timestamp sensitivity, with the line of equal change. (c) HY-WorldPlay at the fixed threshold 0.2 on 218 reverted steps under five histories of two views.\label{fig:readers}}
\end{figure*}

Other generators follow the timestamps less far. With the frames of the target pose added to the geometric retrieval of the WorldMem baseline, on subset A the FAR generator shows the current state on 0.48, 0.36 and 0.19 of the steps under kept, one-time and reversed timestamps and the outdated state on 0.11, 0.36 and 0.49, and the WorldMem baseline shows the current state on 0.38, 0.31 and 0.25 and the outdated state on 0.08, 0.21 and 0.18, keeping the current state above the outdated one under reversal. On the reverted steps at seed 1, reversed capture timestamps raise FAR and lose new-state steps (\cref{tab:s-reverted-conditions}), as \cref{thm:valid}(ii) states for every rule that does not read the log. Write \(\ell=\log(p_C/p_O)\) for the log-odds of the current against the outdated share. The soft rule of \cref{prop:soft} gives \(\ell=\ell_0+\beta\Delta\) with timestamps kept and \(\ell_0-\beta\Delta\) with them reversed, so half the difference of the two, the \emph{timestamp sensitivity}, estimates \(\beta\Delta\) on the same steps. On subset A it is 1.81 for FAR, 1.21 for the FAR generator under geometric retrieval and 0.63 for the WorldMem baseline (\cref{fig:readers}a), with 95\% intervals from resampling clips of 1.53 to 2.15, 1.03 to 1.41 and 0.46 to 0.81. One time lies near the midpoint of the two, 0.21 above it for FAR and 0.30 and 0.59 below it for the other two generators, so the linear form of the soft rule describes these generators to within these offsets.

\begin{table*}[tp]
\centering
\caption{Outcomes of FAR under timestamp and slot conditions at seed 0 on subsets A (1,478 steps) and B (1,351 steps) and at the fixed threshold 0.2 on subset A, mean PSNR in dB on subset A, and LPIPS accuracy on both subsets and on the 825 trajectories per condition of the corridor benchmark.\label{tab:s-conditions}}
\footnotesize\setlength{\tabcolsep}{5pt}
\begin{tabular}{lrrrrrrrrrr}
\toprule
 & \multicolumn{3}{c}{Subset A} & \multicolumn{2}{c}{Subset B} & \multicolumn{2}{c}{A at 0.2} & \multicolumn{3}{c}{LPIPS acc.} \\
\cmidrule(lr){2-4}\cmidrule(lr){5-6}\cmidrule(lr){7-8}\cmidrule(lr){9-11}
Timestamps given to the FAR generator & Current & Outdated & PSNR & Current & Outdated & Current & Outdated & A & B & Corridor \\
\midrule
Kept & 0.510 & 0.118 & 42.0 & 0.457 & 0.107 & 0.479 & 0.152 & 0.787 & 0.788 & 0.953 \\
Equal gaps, true order & 0.524 & 0.124 & 41.0 & 0.473 & 0.115 & 0.485 & 0.145 &  &  &  \\
Gaps times 0.1, true order & 0.317 & 0.380 & 39.6 & 0.343 & 0.323 & 0.314 & 0.480 & 0.600 & 0.621 &  \\
Gaps times 4, true order & 0.225 & 0.048 & 29.4 & 0.203 & 0.021 & 0.155 & 0.043 & 0.508 & 0.452 &  \\
One time & 0.297 & 0.340 & 37.5 & 0.309 & 0.283 & 0.274 & 0.438 & 0.576 & 0.574 & 0.909 \\
Permuted & 0.272 & 0.354 & 37.0 & 0.264 & 0.317 & 0.261 & 0.453 & 0.541 & 0.524 &  \\
Reversed & 0.067 & 0.578 & 33.1 & 0.082 & 0.491 & 0.066 & 0.734 & 0.283 & 0.268 & 0.847 \\
Slots permuted, timestamps kept & 0.509 & 0.118 & 42.0 & 0.457 & 0.105 & 0.479 & 0.153 & 0.789 & 0.789 &  \\
Slots permuted, timestamps reversed & 0.066 & 0.579 & 33.1 & 0.081 & 0.491 & 0.066 & 0.737 & 0.282 & 0.268 &  \\
\bottomrule
\end{tabular}
\end{table*}

\begin{table*}[tp]
\centering
\caption{Timestamp conditions on the 249 clips at seed 1.\label{tab:s-reverted-conditions}}
\footnotesize\setlength{\tabcolsep}{5pt}
\begin{tabular}{lrrrrrrrrrr}
\toprule
 & \multicolumn{3}{c}{Reverted} & \multicolumn{3}{c}{Repeat-state} & \multicolumn{3}{c}{New-state} &  \\
\cmidrule(lr){2-4}\cmidrule(lr){5-7}\cmidrule(lr){8-10}
Memory & Current & Outd. & Low~q. & Current & Outd. & Low~q. & Current & Outd. & Low~q. & Clips +/− \\
\midrule
FAR~\citep{kim2026learning} & 0.355 & 0.267 & 0.219 & 0.575 & 0.184 & 0.162 & 0.434 & 0.103 & 0.252 & ref. \\
one time for every frame & 0.355 & 0.247 & 0.233 & 0.513 & 0.181 & 0.216 & 0.269 & 0.253 & 0.352 & 31/30 \\
timestamps reversed & 0.537 & 0.148 & 0.170 & 0.589 & 0.146 & 0.189 & 0.034 & 0.476 & 0.436 & 44/19 \\
\ours{} per frame & 0.580 & 0.151 & 0.099 & 0.769 & 0.084 & 0.071 & 0.458 & 0.083 & 0.240 & 48/4 \\
\rowcolor{oursrow}\textbf{\ours{} (ours)} & 0.619 & 0.085 & 0.094 & 0.812 & 0.037 & 0.063 & 0.442 & 0.089 & 0.259 & 54/2 \\
\ours{} per frame, one time for every frame & 0.213 & 0.438 & 0.261 & 0.393 & 0.318 & 0.224 & 0.325 & 0.237 & 0.305 & 3/59 \\
\bottomrule
\end{tabular}
\end{table*}

From each pair's context under capture timestamps, \ours{} raises the log-odds of the current against the outdated state by 0.56 for the WorldMem baseline, 1.37 for the FAR generator under geometric retrieval, 1.53 for FAR and 5.32 for HY-WorldPlay, in the order of the timestamp sensitivities of \cref{sec:finding} (\cref{fig:readers}b), which were measured at another seed on other steps or, for HY-WorldPlay, by the probe; two of the four pairs share the FAR generator (\cref{tab:generators}).

\begin{table*}[tp]
\centering
\caption{\ours{} across generators on the 352 reverted steps of seed 1, from each memory's context under capture timestamps to \ours{}'s timestamps. The first three rows read the state metric and take sensitivity from subset A at seed 0; HY-WorldPlay reads the target-pose views in \ours{}'s context at the fixed threshold 0.2, takes sensitivity from the probe of \cref{sec:finding}, and, with the reference image from memory, has 1 current and 99 outdated renders under capture time.\label{tab:generators}}
\small\setlength{\tabcolsep}{6pt}
\begin{tabular}{llrrrrrr}
\toprule
 &  &  & \multicolumn{2}{c}{Capture} & \multicolumn{2}{c}{\ours{}} &  \\
\cmidrule(lr){4-5}\cmidrule(lr){6-7}
Generator and memory & Time enters as & Sensitivity & Cur. & Outd. & Cur. & Outd. & Gain in log-odds \\
\midrule
WorldMem baseline~\citep{xiao2025worldmem} & timestamp embedding & 0.63 & 0.40 & 0.07 & 0.43 & 0.05 & 0.56 \\
FAR generator, geometric retrieval & timestamp embedding & 1.21 & 0.38 & 0.22 & 0.60 & 0.09 & 1.37 \\
FAR~\citep{kim2026learning} & timestamp embedding & 1.81 & 0.39 & 0.25 & 0.62 & 0.09 & 1.53 \\
HY-WorldPlay~\citep{sun2025worldplay} & rotary positions & 3.45 & 0.00 & 0.28 & 0.19 & 0.09 & 5.32 \\
\bottomrule
\end{tabular}
\end{table*}

Read at the fixed threshold 0.2, the reading of HY-WorldPlay, the three generators that read a timestamp embedding give sensitivities of 1.78, 1.20 and 0.51 and gains of 1.81, 1.32 and 0.58, so all four pairs of \cref{tab:generators} keep one order under one reading. The constants of the timestamps leave \ours{} within 0.01 on reverted steps (\cref{tab:s-timestamps}): restored frames 0~s or 6~s after the restoring action give 0.616 and 0.625, and outdated tokens 8~s before the oldest retrieved frame give 0.611 with 0.097 at low quality.

\begin{table*}[tp]
\centering
\caption{\ours{} against its context under capture timestamps for three generators, and the timestamp constants, at seed 1 on the 352 reverted, 1,050 repeat-state and 2,968 new-state steps of the 249 clips. \(\ell\) is the log-odds of the current against the outdated share on reverted steps; bold and underline mark the best and second best within a block.\label{tab:s-timestamps}}
\small\setlength{\tabcolsep}{6pt}
\begin{tabular}{lrrrrrr}
\toprule
 & \multicolumn{4}{c}{Reverted} & \multicolumn{1}{c}{Repeat-state} & \multicolumn{1}{c}{New-state} \\
\cmidrule(lr){2-5}\cmidrule(lr){6-6}\cmidrule(lr){7-7}
Generator and memory & Current↑ & Outd.↓ & Low q.↓ & \(\ell\) & Current↑ & Current↑ \\
\midrule
\multicolumn{7}{l}{\emph{FAR}} \\
capture timestamps~\citep{kim2026learning} & 0.355 & 0.267 & 0.219 & 0.29 & 0.575 & 0.434 \\
\ours{}'s context, capture timestamps & 0.392 & 0.250 & 0.188 & 0.45 & 0.646 & 0.440 \\
\rowcolor{oursrow}\textbf{\ours{} (ours)} & \underline{0.619} & \textbf{0.085} & \textbf{0.094} & 1.98 & \underline{0.812} & 0.442 \\
restored frames 0~s after the action & 0.616 & 0.088 & \textbf{0.094} & 1.95 & 0.809 & 0.442 \\
restored frames 6~s after the action & \textbf{0.625} & \textbf{0.085} & \textbf{0.094} & 1.99 & \textbf{0.813} & \underline{0.445} \\
outdated tokens 8~s before the oldest frame & 0.611 & \textbf{0.085} & 0.097 & 1.97 & 0.808 & \textbf{0.451} \\
\midrule
\multicolumn{7}{l}{\emph{FAR generator, geometric retrieval with target-pose frames and post-action retrieval}} \\
capture timestamps & 0.381 & 0.222 & 0.233 & 0.54 & 0.608 & 0.383 \\
\rowcolor{oursrow}\textbf{\ours{} (ours)} & \textbf{0.597} & \textbf{0.088} & \textbf{0.159} & 1.91 & \textbf{0.750} & \textbf{0.396} \\
\midrule
\multicolumn{7}{l}{\emph{WorldMem baseline with target-pose frames and post-action retrieval}} \\
capture timestamps~\citep{xiao2025worldmem} & 0.401 & 0.074 & 0.295 & 1.69 & \textbf{0.602} & 0.283 \\
\ours{} per frame & \textbf{0.432} & \underline{0.060} & \underline{0.278} & 1.97 & 0.577 & \textbf{0.289} \\
\rowcolor{oursrow}\textbf{\ours{} (ours)} & \textbf{0.432} & \textbf{0.045} & \textbf{0.273} & 2.25 & \underline{0.580} & \textbf{0.289} \\
\bottomrule
\end{tabular}
\end{table*}

\subsection{A second reader: HY-WorldPlay}

HY-WorldPlay~\citep{sun2025worldplay} generates video in chunks of four latents from a reference frame, a camera path, a prompt and a history of earlier latents; we run the released autoregressive checkpoint in its 480p image-to-video configuration with 20 sampling steps. The history enters attention through temporal rotary positions and carries no other signal of time. Our probe takes the 219 seen reverted steps of seed 1, of which 218 have both stored views. For each step it decodes three of FAR's latents to 256 px: the latest frame stored at the target pose, which shows the outdated state (B), the newest earlier frame from that pose whose latent matches the ground-truth frame (A), chosen by the ground-truth frame alone, and the ground-truth frame. The reference frame is the ground-truth frame with the changed tokens and their neighbors painted gray, so that the state of the region reaches the generator only through its history. The camera holds its pose, the prompt is empty, and the five histories of a step share one noise draw. Each history holds a view for one chunk of four temporal positions: B alone at positions 0 to 3 (deletion), A at 0 to 3 and B at 4 to 7 (capture time), A and B both at 0 to 3 (one time), B at 0 to 3 and A at 4 to 7 (valid time), and A alone (filtering). WorldPlay generates two chunks, the first following the reference frame and the second the history, and the last frame is encoded with FAR's tokenizer and read against the ground-truth and stale frames with the state metric of \cref{sec:setup}. Its renders, drawn at 480p from 256 px views, are read at the fixed threshold 0.2 (\cref{sec:metric}), set before any render existed.

\begin{figure*}[tp]
\centering
\includegraphics[width=\linewidth,height=0.82\textheight,keepaspectratio]{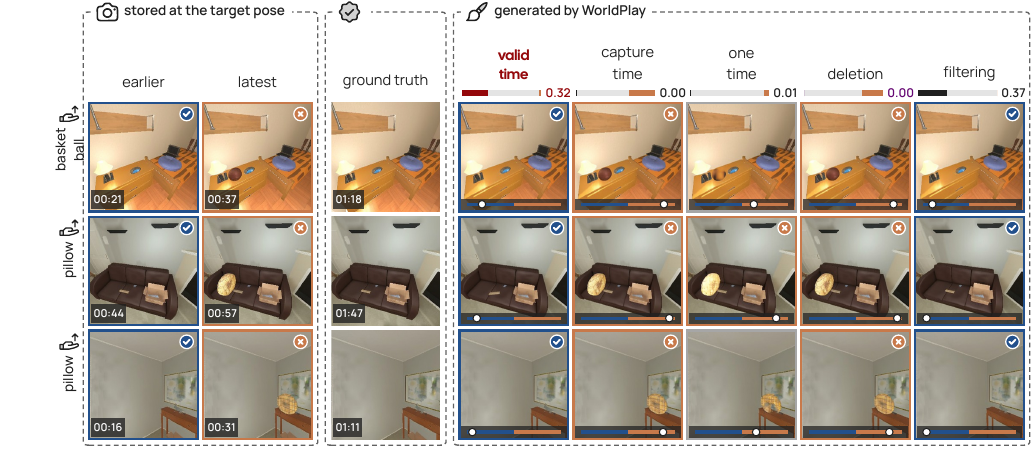}
\caption{HY-WorldPlay on the 218 seen reverted steps of seed 1 with both views. The earlier target-pose view is current and the latest is outdated; bars give the current share under five histories.\label{fig:worldplay}}
\end{figure*}

\begin{table*}[tp]
\centering
\caption{HY-WorldPlay on the 218 seen reverted steps of seed 1 (99 clips) under five histories of the earlier current view (A) and the latest view at the target pose (B). At 0.2 is the fixed threshold, and Nearer current is the share of renders closer in FAR's latent space to the ground-truth frame than to the stale frame.\label{tab:s-worldplay}}
\small\setlength{\tabcolsep}{4pt}
\begin{tabular}{llrrrrrrrr}
\toprule
 &  & \multicolumn{3}{c}{At 0.2} &  & \multicolumn{3}{c}{State metric} &  \\
\cmidrule(lr){3-5}\cmidrule(lr){7-9}
History & Positions & current & outdated & neither & Nearer current & current & outdated & low q. & PSNR \\
\midrule
Deletion & B & 0.000 & 0.266 & 0.734 & 0.225 & 0.000 & 0.193 & 0.771 & 25.9 \\
Capture time & A, then B & 0.005 & 0.326 & 0.670 & 0.193 & 0.014 & 0.248 & 0.702 & 27.0 \\
One time & A and B together & 0.014 & 0.060 & 0.927 & 0.404 & 0.051 & 0.142 & 0.578 & 28.8 \\
\rowcolor{oursrow}\textbf{Valid time, \ours{} (ours)} & B, then A & 0.321 & 0.023 & 0.656 & 0.899 & 0.408 & 0.041 & 0.477 & 29.9 \\
Filtering & A & 0.367 & 0.000 & 0.633 & 0.959 & 0.367 & 0.000 & 0.569 & 28.5 \\
\bottomrule
\end{tabular}
\end{table*}

From capture to valid time, 47 steps move from the outdated to the current state, 22 from neither state to the current state and 19 from the outdated state to neither, and no step leaves the current state (\cref{tab:s-worldplay}, \cref{fig:readers}c). The current share rises by 0.32, with a 95\% interval of 0.25 to 0.39 from resampling the 99 clips, and the share of renders nearer the current state rises from 0.19 to 0.90. Withholding B and keeping A, the change a filtering memory makes, raises the current share by 0.36, so exchanging two positions does nearly what withholding the outdated view does and keeps that view in the history. Deletion draws the outdated state less often than capture time, because a two-view history draws the region more sharply (\cref{tab:s-worldplay}).

The distance between the two temporal positions plays the part of the gap between timestamps (\texttt{worldplay\_reader.py\ -\/-offset}). The probe places the later view one chunk (4 latent positions) after the first, and one time places both at one position; two further runs place the later view 1 and 2 positions after the first on the same 218 steps, views, reference, camera, prompt and noise under capture order and valid time (\cref{tab:s-offsets}). The separation in the log-odds \(\ell\) of the counts plus 0.5 rises with the offset, as \cref{cor:sensitivity} states for a gap, and flattens as capture order's current renders fall to 1 or 2 of 218, where the log-odds stop resolving the share.

\begin{table*}[tp]
\centering
\caption{HY-WorldPlay on the 218 probe steps with the later of its two views a given number of latent positions after the first, at the fixed threshold 0.2. \(\ell\) is the log-odds of the current against the outdated renders, each count plus 0.5; offsets 0 and 4 come from \cref{tab:s-worldplay}.\label{tab:s-offsets}}
\small\setlength{\tabcolsep}{6pt}
\begin{tabular}{rrrrrr}
\toprule
 & \multicolumn{2}{c}{Capture order} & \multicolumn{2}{c}{Valid time} &  \\
\cmidrule(lr){2-3}\cmidrule(lr){4-5}
Offset & current & outdated & current & outdated & Difference of \(\ell\) \\
\midrule
0 (one time) & 0.014 & 0.060 & 0.014 & 0.060 & 0.0 \\
1 & 0.028 & 0.262 & 0.289 & 0.060 & 3.7 \\
2 & 0.009 & 0.440 & 0.390 & 0.050 & 5.7 \\
4 (one chunk) & 0.005 & 0.326 & 0.321 & 0.023 & 6.4 \\
\bottomrule
\end{tabular}
\end{table*}

The probe hands WorldPlay two views chosen with the ground-truth frame; \ours{}'s own context tests the method on this generator with no ground-truth frame or change mask in the run (\texttt{worldplay\_context.py}, \cref{tab:s-worldplay-context}). At each reverted step (352 at seed 1 and 366 at seed 0), the retrieved frames of the target pose from FAR's saved \ours{} context enter WorldPlay's history, capped at four views by memory use before any render was scored (2.16 on average at seed 1): those \ours{} re-dates, then those with tokens of their own date, then the newest by capture. The same views enter under capture time and under \ours{}'s timestamps; each view takes the 4-latent chunk of its date rank, and under \ours{}'s timestamps the outdated tokens take the chunk of their own date, the per-token timestamps of \cref{thm:frame}(iii) in rotary positions. The reference image is the newest target-pose view in \ours{}'s context with the tokens that the state tracker dates apart and their neighbors painted gray, and the camera, prompt and noise follow the probe. From capture time to \ours{}'s timestamps, the current share at the fixed threshold 0.2 rises from 0.003 to 0.188 at seed 1, with a clip interval of 0.13 to 0.24, 43 clips higher and 1 lower, and the outdated share falls from 0.281 to 0.091; at seed 0 it rises from 0.003 to 0.208, with an interval of 0.16 to 0.26, 47 clips higher and 1 lower. On the held-out clips it rises from 0.003 to 0.181 at seed 1 (288 steps, 35 clips higher and 1 lower) and from 0.003 to 0.209 at seed 0 (302 steps), and on the seen steps of seed 1 from 0.005 to 0.301, near the 0.321 that the probe reaches with views that the ground-truth frame chooses.

Valid time also tells a memory what to keep. On the 222 reverted steps of seed 1 whose context holds a target-pose view that \ours{} re-dates, the newest such view was captured a median 77~s before the step (55~s to 119~s for four steps in five), so a memory of the last 30~s, 60~s or 120~s of capture keeps none, 0.28 or 0.91 of them; it became valid a median 44~s before the step (31~s to 72~s for four steps in five), and a memory of the same windows of valid time keeps 0.07, 0.82 and 221 of the 222.

On the same steps, views, reference, camera, prompt and noise, state filtering and token deletion read \ours{}'s log in WorldPlay (\cref{tab:s-worldplay-context}). Filtering keeps the views with no tokens dated apart at their capture ranks, and leaves WorldPlay the history generated from the reference image when no view remains. Token deletion places every view at its valid-time rank and hides the tokens dated apart from every attention, the history pass included, and from the cache. Both rules gain from \ours{}'s log as \ours{} does: token deletion reaches 0.233 at seed 1 and 0.257 at seed 0 at 25.7~dB, state filtering 0.173 and 0.186 at 24.0~dB, and \ours{} 0.188 and 0.208 at 28.0 and 27.8~dB. In WorldPlay the action log carries the gain, and hiding tokens costs image quality as it does in FAR.

\begin{table*}[tp]
\centering
\caption{HY-WorldPlay on the target-pose views in \ours{}'s context at seeds 1 and 0, with the reference image from memory; each block holds one set of steps, and state filtering and token deletion read \ours{}'s log. At 0.2 is the fixed threshold, Nearer current the two-way reading, and the state metric keeps its quality floor; bold and underline mark the best and second best within a block.\label{tab:s-worldplay-context}}
\small\setlength{\tabcolsep}{6pt}
\begin{tabular}{lrrrrrrr}
\toprule
 & \multicolumn{3}{c}{At 0.2} &  & \multicolumn{2}{c}{State metric} &  \\
\cmidrule(lr){2-4}\cmidrule(lr){6-7}
History & current↑ & outdated↓ & neither & Nearer current↑ & current↑ & low q.↓ & PSNR↑ \\
\midrule
\multicolumn{8}{l}{\emph{Seed 1, all 352 reverted steps}} \\
capture time & 0.003 & 0.281 & 0.716 & 0.19 & 0.003 & \underline{0.713} & \underline{27.1} \\
state filtering, \ours{}'s log & 0.173 & \textbf{0.020} & 0.807 & \textbf{0.84} & 0.190 & 0.767 & 24.0 \\
token deletion, \ours{}'s log & \textbf{0.233} & \underline{0.028} & 0.739 & \underline{0.76} & \textbf{0.256} & \underline{0.713} & 25.7 \\
\rowcolor{oursrow}\textbf{\ours{} (ours)} & \underline{0.188} & 0.091 & 0.722 & 0.63 & \underline{0.213} & \textbf{0.625} & \textbf{28.0} \\
\midrule
\multicolumn{8}{l}{\emph{Seed 1, 219 steps whose target pose stored the current state}} \\
capture time & 0.005 & 0.297 & 0.699 & 0.19 & 0.005 & 0.735 & 27.1 \\
state filtering, \ours{}'s log & 0.279 & \textbf{0.018} & 0.703 & \underline{0.91} & 0.306 & 0.635 & 27.8 \\
token deletion, \ours{}'s log & \textbf{0.374} & \textbf{0.018} & 0.607 & \textbf{0.93} & \textbf{0.411} & \textbf{0.553} & \textbf{29.0} \\
\rowcolor{oursrow}\textbf{\ours{} (ours)} & \underline{0.301} & \textbf{0.018} & 0.680 & 0.89 & \underline{0.342} & \underline{0.580} & \underline{28.7} \\
\midrule
\multicolumn{8}{l}{\emph{Seed 0, all 366 reverted steps}} \\
capture time & 0.003 & 0.249 & 0.749 & 0.19 & 0.000 & 0.746 & \underline{26.7} \\
state filtering, \ours{}'s log & 0.186 & \textbf{0.014} & 0.801 & \textbf{0.82} & 0.208 & 0.762 & 24.0 \\
token deletion, \ours{}'s log & \textbf{0.257} & \textbf{0.014} & 0.730 & \underline{0.77} & \textbf{0.262} & \underline{0.708} & 25.7 \\
\rowcolor{oursrow}\textbf{\ours{} (ours)} & \underline{0.208} & 0.060 & 0.732 & 0.61 & \underline{0.224} & \textbf{0.648} & \textbf{27.8} \\
\midrule
\multicolumn{8}{l}{\emph{Seed 0, 238 steps whose target pose stored the current state}} \\
capture time & 0.004 & 0.244 & 0.752 & 0.18 & 0.000 & 0.769 & 26.8 \\
state filtering, \ours{}'s log & 0.286 & \textbf{0.017} & 0.698 & 0.86 & 0.319 & 0.639 & 27.7 \\
token deletion, \ours{}'s log & \textbf{0.395} & \textbf{0.017} & 0.588 & \textbf{0.90} & \textbf{0.403} & \textbf{0.555} & \textbf{28.8} \\
\rowcolor{oursrow}\textbf{\ours{} (ours)} & \underline{0.319} & 0.025 & 0.655 & \underline{0.87} & \underline{0.345} & \underline{0.588} & \underline{28.5} \\
\bottomrule
\end{tabular}
\end{table*}

\subsection{Undo in real homes}

Every other result of the paper comes from AI2-THOR; this subsection tests whether temporal position steers HY-WorldPlay after an undo in real homes. Charades~\citep{sigurdsson2016hollywood} films scripted activities in homes with a mostly stationary camera and annotates every action with its interval, so its annotations serve as the action log. An event is an opening of a door, a closet or cabinet, a refrigerator or a window and the next closing of the same kind at least 1~s after the opening ends, which returns the furniture to the state it held before the opening (\texttt{charades\_undo.py}). Three frames of the 480p video enter: A, 1~s before the opening, the older view of the closed state; B, midway between the opening and the closing, the newest view, open; and G, 1~s after the closing, which only evaluates the renders. \ours{}'s state tracker reads the two annotated actions and dates A after B. A and G are registered to B by a homography from SIFT matches with RANSAC, and the three frames are cropped to their central 256 px square. Of the 317 events of the release, 55 keep both registrations at 40 inliers or more with the crop covered and between 4 and 400 changed tokens, and 31 keep at least 4 changed tokens once the tokens that a person covers or borders in any of the three frames, found by torchvision's Faster R-CNN at a score of 0.5, leave the evaluation mask, so that the renders are read on the furniture and its surroundings. WorldPlay renders every event under the five histories of the probe with two reference images, B with the tokens where A and B differ painted gray so that no frame after the closing enters the run, and G with the tokens of the evaluation mask painted gray as in the probe (\texttt{worldplay\_real.py}). The camera holds its pose, the prompt is empty, and the histories of an event share one noise draw.

\begin{figure*}[tp]
\centering
\includegraphics[width=\linewidth,height=0.82\textheight,keepaspectratio]{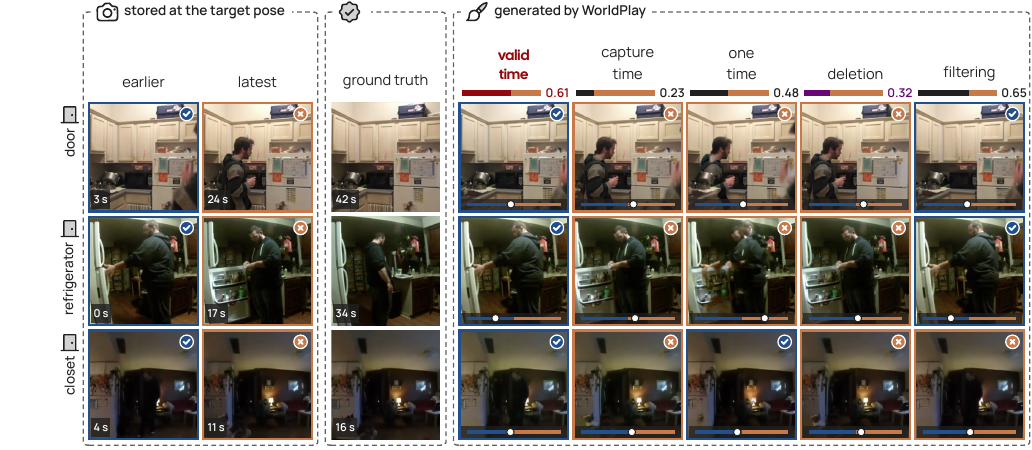}
\caption{HY-WorldPlay after an undo in Charades homes with the reference image from memory. Badges give the two-way reading, and bars give the share of the 31 events nearer the post-closing frame G.\label{fig:homes}}
\end{figure*}

WorldPlay draws these real frames at 15 to 16~dB of PSNR against G, so we read them with the two-way reading, which asks which of G and B a render lies nearer over the evaluation mask; it follows the order of the AI2-THOR probe (\cref{tab:s-homes,fig:homes}). With the reference from memory, the share nearer G rises from capture time to valid time by 0.39, with a 95\% interval of 0.19 to 0.58 from resampling the 31 videos, 13 events moving toward G and 1 away, and one time lies between the two orders.

\begin{table*}[tp]
\centering
\caption{HY-WorldPlay on 31 undo events of Charades under five histories of the older closed view (A) and the newest open view (B). Nearer G is the share of renders closer in FAR's latent space to the post-closing frame G than to B over the evaluation mask, current at 0.2 is the fixed-threshold current share (where no render reads outdated), and PSNR is measured against G over the whole frame; bold and underline mark the best and second best.\label{tab:s-homes}}
\small\setlength{\tabcolsep}{6pt}
\begin{tabular}{llrrrrrr}
\toprule
 &  & \multicolumn{3}{c}{Memory reference} & \multicolumn{3}{c}{Probe reference} \\
\cmidrule(lr){3-5}\cmidrule(lr){6-8}
History & Positions & nearer G↑ & current at 0.2↑ & PSNR↑ & nearer G↑ & current at 0.2↑ & PSNR↑ \\
\midrule
Deletion & B & 0.32 & 0.00 & 14.8 & 0.29 & 0.00 & 14.8 \\
Capture time & A, then B & 0.23 & 0.00 & 15.1 & 0.23 & 0.00 & 15.1 \\
One time & A and B together & 0.48 & \underline{0.03} & \textbf{16.3} & 0.58 & \underline{0.03} & \textbf{16.3} \\
\rowcolor{oursrow}\textbf{Valid time, \ours{} (ours)} & B, then A & \underline{0.61} & \textbf{0.06} & \underline{16.2} & \underline{0.68} & \textbf{0.06} & \textbf{16.3} \\
Filtering & A & \textbf{0.65} & \underline{0.03} & 15.6 & \textbf{0.71} & \underline{0.03} & 15.9 \\
\bottomrule
\end{tabular}
\end{table*}

\subsection{Deletion, oracles and post-action retrieval}

\Cref{tab:s-stores} compares ten stores under FAR's trained retriever: FAR as released, three read-time rules, two oracles and four deletion rules (\cref{tab:s-designs}). The deletion rules lose the current state on reverted steps because the frames alone do not tell which old tokens will be valid again. Of the 125 reverted steps that FAR draws in the current state, frame deletion by pose keeps 33 and moves 83 to the ambiguous state, and it gives 0.71 against 0.69 on the other 698 repeat-state steps. On new-state steps, deletion at the time of the change removes most outdated renders at a cost in quality. The oracles lower the outdated share more than they raise the current one. On subsets A and B at seed 0, oracle token deletion lowers the outdated share to 0.07 and 0.06 against 0.12 and 0.10 for its sham, which hides as many tokens of an unchanged region, and keeps the current share at 0.51 and 0.45, and oracle frame deletion lowers it to 0.03 and 0.03 at a cost of 3.7 and 3.5~dB, with the current share at 0.49 and 0.43 against 0.51 and 0.46 for FAR.

Post-action retrieval raises the current share on repeat-state steps at seed 1 from 0.58 to 0.66, with 28 clips higher and 10 lower, and by +0.079 and +0.077 at the fixed thresholds 0.2 and 0.1. Outside the first 200 clips it raises repeat-state accuracy from 0.554 to 0.610 at seed 1, with 19 clips higher and 10 lower, and from 0.527 to 0.606 at seed 0, with 24 higher and 8 lower, and on the first 200 clips from 0.64 to 0.80. On new-state steps it gives 0.434 beside 0.434.

\begin{table*}[tp]
\centering
\caption{Stores under FAR's trained retriever at seed 1 on the 1,050 repeat-state, 352 reverted and 2,968 new-state steps. Readable is the current share of readable frames and At 0.2 the current share at the fixed threshold 0.2, and a plus sign adds a read-time rule to FAR. ‡ reads the action log and † the simulator; bold and underline mark the best and second best without †.\label{tab:s-stores}}
\footnotesize\setlength{\tabcolsep}{4pt}
\begin{tabular}{lrrrrrrrrrr}
\toprule
 & \multicolumn{5}{c}{Repeat-state} & \multicolumn{2}{c}{Reverted} & \multicolumn{3}{c}{New-state} \\
\cmidrule(lr){2-6}\cmidrule(lr){7-8}\cmidrule(lr){9-11}
Store read by FAR & Current↑ & Outd.↓ & Low q.↓ & Readable↑ & At 0.2↑ & Current↑ & Low q.↓ & Current↑ & Outd.↓ & Low q.↓ \\
\midrule
FAR~\citep{kim2026learning} & 0.575 & 0.184 & 0.162 & 0.69 & 0.558 & 0.355 & 0.219 & 0.434 & 0.103 & \textbf{0.252} \\
+ post-action retrieval & \textbf{0.659} & 0.137 & \textbf{0.116} & \underline{0.75} & \underline{0.637} & \underline{0.412} & \underline{0.176} & 0.434 & 0.087 & 0.258 \\
+ decay by age & 0.602 & 0.163 & 0.153 & 0.71 & 0.590 & 0.401 & 0.193 & 0.436 & 0.093 & 0.254 \\
+ post-action retrieval and decay & \underline{0.656} & 0.135 & \underline{0.124} & \underline{0.75} & \textbf{0.639} & \textbf{0.426} & \textbf{0.173} & \underline{0.437} & \underline{0.083} & 0.255 \\
Frame deletion by pose~\citep{wu2026echo} & 0.507 & \underline{0.108} & 0.162 & 0.60 & 0.474 & 0.111 & 0.193 & \textbf{0.439} & 0.085 & \textbf{0.252} \\
Deletion of all earlier frames‡~\citep{gu2026worldcraft} & 0.200 & \textbf{0.000} & 0.773 & \textbf{0.88} & 0.193 & 0.023 & 0.946 & 0.253 & \textbf{0.000} & 0.732 \\
\midrule
Oracle token deletion†~\citep{yang2026permavid} & 0.610 & 0.104 & 0.162 & 0.73 & 0.572 & 0.429 & 0.236 & 0.426 & 0.064 & 0.311 \\
Oracle frame deletion†~\citep{yang2026oneira} & 0.539 & 0.045 & 0.262 & 0.73 & 0.514 & 0.406 & 0.381 & 0.409 & 0.027 & 0.449 \\
Token deletion at each change†~\citep{yang2026permavid} & 0.458 & 0.016 & 0.285 & 0.64 & 0.407 & 0.097 & 0.446 & 0.424 & 0.020 & 0.362 \\
Frame deletion at each change†~\citep{yang2026permavid} & 0.406 & 0.024 & 0.442 & 0.73 & 0.387 & 0.054 & 0.815 & 0.418 & 0.015 & 0.488 \\
\bottomrule
\end{tabular}
\end{table*}

\subsection{Frames stored at the target pose}\label{sec:views}

\begin{figure*}[tp]
\centering
\includegraphics[width=\linewidth,height=0.82\textheight,keepaspectratio]{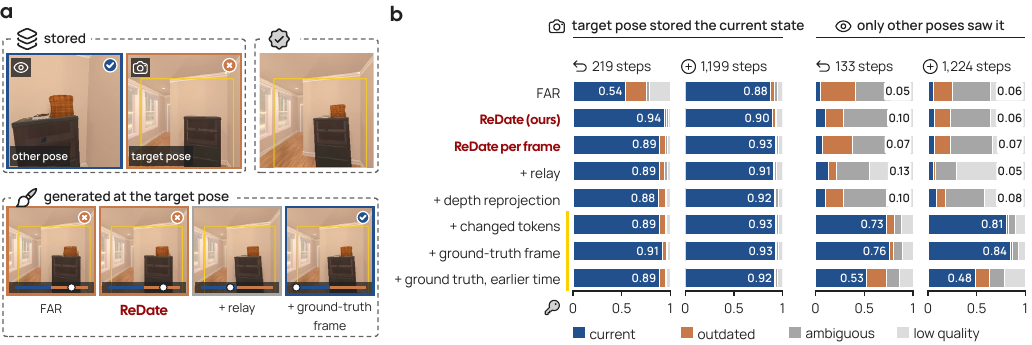}
\caption{(a) On an unseen step, FAR and \ours{} copy the stale frame at the target pose, and a ground-truth frame stored there restores the current dresser. (b) Outcomes on seen and unseen steps for the 352 reverted steps and the 2,423 new-state steps whose split is recorded.\label{fig:views}}
\end{figure*}

This subsection tests whether supplying a view at the target pose \(g\) resolves unseen steps (\cref{fig:views,tab:s-views}). Transport supplies a frame at \(g\) for frame \(q\) when an action changes a manipulated region inside the target view after the newest frame stored at \(g\) and every stored frame at \(g\) shows an outdated state. Its \emph{source} is the nearest stored frame from another pose that shows the region in its current state: a frame captured after the last action on the region, or one that \ours{} re-dates. Steps without a source have no current observation and stay as \ours{} leaves them. Two transports build a frame at \(g\) from the source without privileged information and store it under valid time. The \emph{relay} walks FAR's rollout from the source to \(g\) in at most 16 steps along a straight path that keeps the region in view, withholding frames in an outdated state and keeping the re-dated context frames; the \emph{relay with splicing} keeps the newest frame at \(g\) and takes the walk's final tokens where they change within 0.8~m of the region. \emph{Depth reprojection} predicts VGGT~\citep{wang2025vggt} depth on a metric scale from reprojection across nearby frames and splats source pixels within 0.8~m of the region into \(g\) to update the newest frame there. Each frame takes the timestamp 2~s after the last action on the region and the first retrieval slot of \ours{}. Three privileged references read the simulator. The \emph{ground-truth frame} stores the simulator's frame at the first served step, the \emph{changed tokens} store the newest frame at \(g\) with the simulator's tokens on that step's change mask, and the control \emph{timestamped before} places the transported or ground-truth frame one rollout step before the oldest frame at \(g\) in an outdated state.

In \cref{tab:s-views} the relay draws the region in neither state as the rest of the frame drifts, and depth reprojection leaves most of it in an outdated state, so both stay at or below 0.13 on unseen steps. The ground-truth frame raises those steps far more under valid time than when timestamped before, as \cref{thm:view}(iii) states. Rendering the current state into the target pose is the next problem, and \cref{sec:analysis} isolates it.

\begin{table*}[tp]
\centering
\caption{Frames stored at the target pose at seed 1. The rows below \ours{} per frame each add one frame to it, and Seen and Unseen split steps by whether the target pose stored a frame of the current state. † reads the simulator; bold and underline mark the best and second best without †.\label{tab:s-views}}
\small\setlength{\tabcolsep}{6pt}
\begin{tabular}{lrrrrrr}
\toprule
 & \multicolumn{2}{c}{Reverted} & \multicolumn{2}{c}{New-state} & \multicolumn{2}{c}{All} \\
\cmidrule(lr){2-3}\cmidrule(lr){4-5}\cmidrule(lr){6-7}
Frame stored at the target pose & Seen↑ & Unseen↑ & Seen↑ & Unseen↑ & Reverted↑ & New-state↑ \\
\midrule
FAR & 0.54 & 0.05 & 0.88 & 0.06 & 0.36 & 0.43 \\
\rowcolor{oursrow}\textbf{\ours{} (ours)} & \textbf{0.94} & 0.10 & 0.90 & 0.06 & \textbf{0.62} & 0.44 \\
\ours{} per frame & 0.89 & 0.07 & \textbf{0.93} & 0.07 & 0.58 & \textbf{0.46} \\
relay & 0.89 & \textbf{0.13} & 0.91 & 0.05 & \underline{0.60} & 0.45 \\
relay, spliced & 0.86 & \underline{0.11} & 0.91 & 0.06 & 0.58 & 0.45 \\
depth reprojection & 0.88 & 0.10 & 0.92 & \textbf{0.08} & 0.58 & \textbf{0.46} \\
depth reprojection, timestamped before & \underline{0.91} & 0.10 & \textbf{0.93} & \textbf{0.08} & \underline{0.60} & \textbf{0.46} \\
changed tokens† & 0.89 & 0.73 & 0.93 & 0.81 & 0.83 & 0.81 \\
ground-truth frame† & 0.91 & 0.76 & 0.93 & 0.84 & 0.86 & 0.83 \\
ground-truth frame timestamped before† & 0.89 & 0.53 & 0.92 & 0.48 & 0.75 & 0.65 \\
\bottomrule
\end{tabular}
\end{table*}

\subsection{Cost}

\Cref{tab:s-cost} gives the runtime and peak memory per rollout step. The state tracker is a table update per action, and per-token timestamps add one embedding difference per token. \ours{} fills FAR's 12 context slots from the same stored frames, so a bank held in host memory sends the GPU as many frames per step as FAR's, the host-to-device traffic on the critical path of offloaded decoding~\citep{cao2026specfit}.

\begin{table}[tbp]
\centering
\caption{Seconds per rollout step over 1,312 rollout steps of 13 clips, each pair measured in one run, with the peak memory of both.\label{tab:s-cost}}
\small\setlength{\tabcolsep}{6pt}
\begin{tabular}{lrrr}
\toprule
Comparison & FAR (s) & \ours{} (s) & Peak memory \\
\midrule
\ours{} per frame & 0.443 & 0.452 & 6.1~GiB \\
\rowcolor{oursrow}\textbf{\ours{} (ours)} & 0.436 & 0.472 & 6.1~GiB \\
\bottomrule
\end{tabular}
\end{table}

\subsection{More reverted steps}

\begin{figure*}[tp]
\centering
\includegraphics[width=\linewidth,height=0.82\textheight,keepaspectratio]{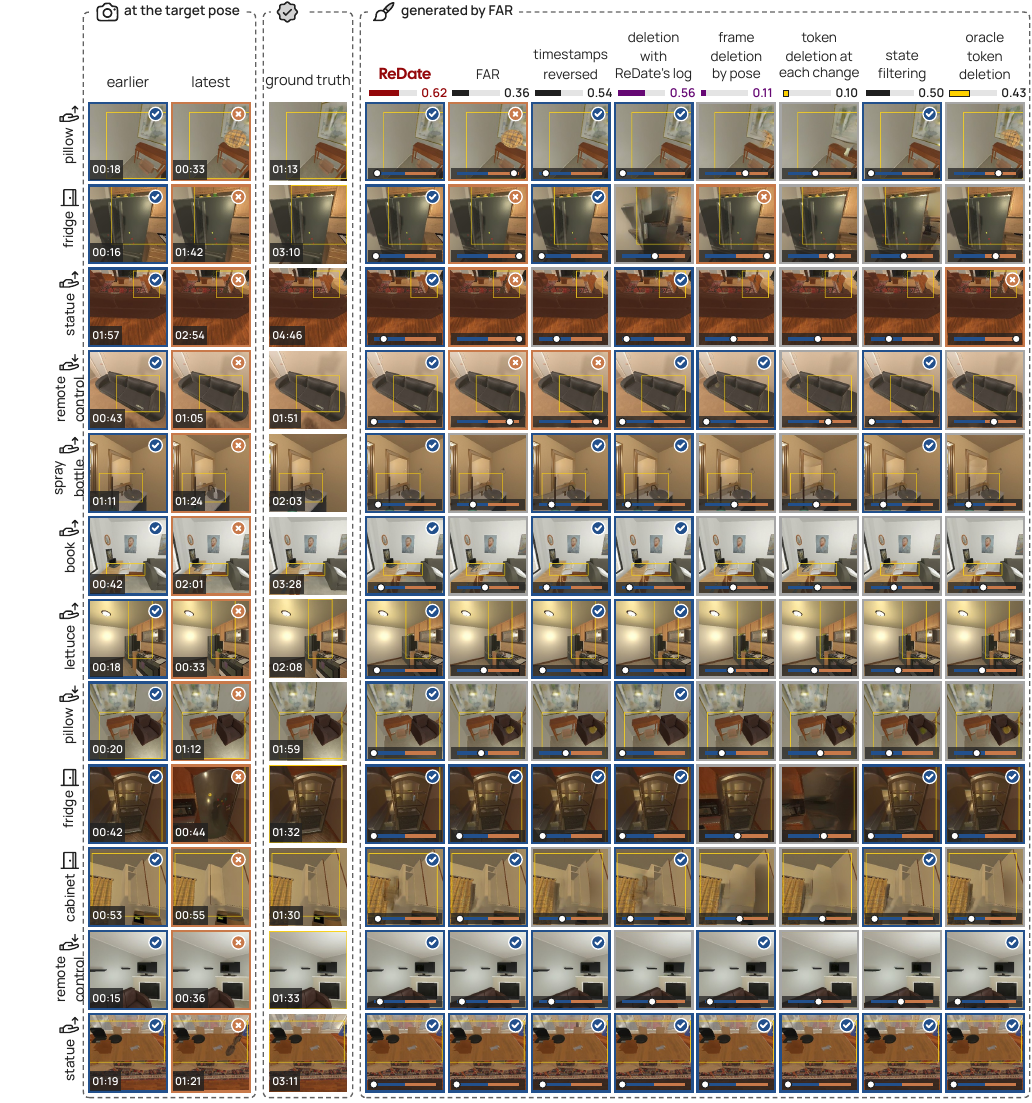}
\caption{Twelve more reverted steps at seed 1 on which \ours{} shows the current state, four for each reading of FAR's output. Columns follow \cref{fig:examples}, with FAR under reversed timestamps and token deletion at each change added.\label{fig:gallery}}
\end{figure*}

\Cref{fig:gallery} extends \cref{fig:examples} with twelve more reverted steps at seed 1, four for each reading of FAR's output (outdated, ambiguous and current), chosen across objects and actions among the steps on which \ours{} shows the current state. Across the twelve rows reversed timestamps draw the current state on eight, state filtering on six, frame deletion by pose and oracle token deletion on three, and token deletion at each change on one, and the deletion-based memories mostly draw neither state, leaving a ghost or a smeared region. \Cref{sec:failures} shows steps whose current state only other poses saw.

\subsection{Steps seen only from other poses}\label{sec:failures}

\begin{figure*}[tp]
\centering
\includegraphics[width=\linewidth,height=0.82\textheight,keepaspectratio]{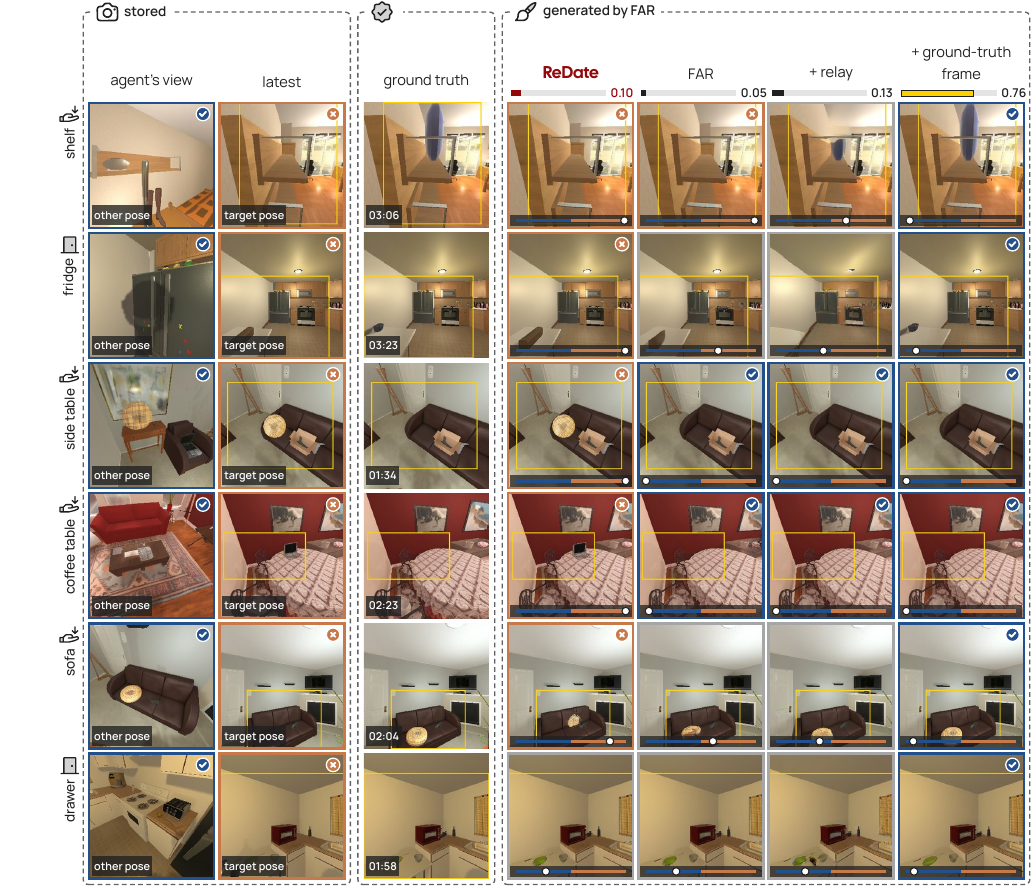}
\caption{Unseen reverted steps at seed 1, on which only other poses saw the current state and a ground-truth frame stored at the target pose restores it, showing the agent's view 2~s after its last action, the latest frame at the target pose, the ground-truth frame and FAR's output under four memories. Bars give accuracy on the 133 unseen reverted steps.\label{fig:failures}}
\end{figure*}

Of the 134 reverted steps on which \ours{} misses the current state, 120 lie in the 133 unseen steps (\cref{sec:analysis}), and \cref{fig:failures} shows six of them. In each the agent saw the current state from its own pose and the frames stored at the target pose show the outdated one, the case of \cref{thm:view}(i). A frame of the current state at the target pose, which \cref{thm:view}(iii) asks for and no timestamp supplies, restores every row.

\begin{table*}[tp]
\centering
\caption{The memories closest to \ours{}, by what each does with a contradicted view and what its generator reads once the agent restores the region, as each work describes its design.\label{tab:s-neighbours}}
\small\setlength{\tabcolsep}{4pt}
\begin{tabularx}{\linewidth}{l>{\raggedright\arraybackslash\hsize=0.999\hsize}X>{\raggedright\arraybackslash\hsize=0.680\hsize}X>{\raggedright\arraybackslash\hsize=1.320\hsize}X}
\toprule
Memory & A contradicted view & Signal & After the agent restores the region \\
\midrule
PermaVid~\citep{yang2026permavid} & invalidated or masked & change times and masks & the view of the restored state is gone \\
WorldCraft~\citep{gu2026worldcraft} & invalidated & the agent's edit & the view of the restored state is gone \\
Echo-Forcing~\citep{wu2026echo} & decayed or removed & a later frame of the same pose & the view of the restored state is gone or weakened \\
ONEIRA~\citep{yang2026oneira} & withheld while it disagrees & a world-state table & a frame returns when every region of the target view agrees \\
MemWorld~\citep{anon2026memworld} & kept as depth only & an edit & the geometry returns without the appearance \\
StateCommit~\citep{anon2026statecommit} & replaced by a committed symbolic state & the agent's interactions & the symbolic state returns, bound to the predicted view \\
Deep Forcing~\citep{yi2025deep} & moved next to the present when it is among the oldest & the position in the context & current and outdated views move alike \\
Zep~\citep{rasmussen2025zep} & kept with a closed validity interval & facts extracted from text & a text fact, read by a language model \\
\rowcolor{oursrow}\textbf{\ours{} (ours)} & kept, its tokens moved into the past & the agent's action log & the stored view, re-dated to the restoring action \\
\bottomrule
\end{tabularx}
\end{table*}

\section{How often worlds return}

\begin{figure}[tbp]
\centering
\includegraphics[width=\linewidth,height=0.82\textheight,keepaspectratio]{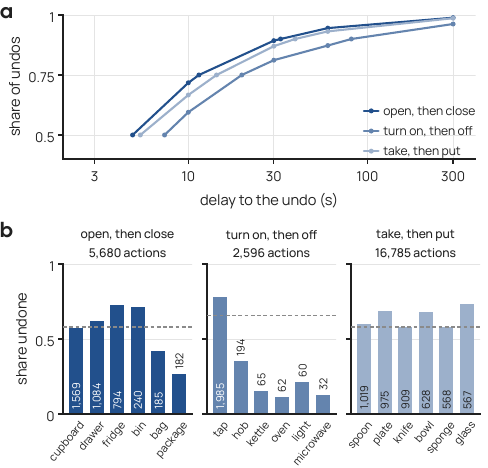}
\caption{Undos in EPIC-KITCHENS-100. (a) Share of undone actions reversed within a given delay for each verb pair. (b) Share of actions undone later in the same video for the six most frequent noun classes of each pair, with action counts; dashed lines mark the share over all noun classes.\label{fig:returns}}
\end{figure}

This appendix measures how often everyday actions undo earlier ones in EPIC-KITCHENS-100~\citep{damen2022rescaling}, supporting \cref{sec:introduction} (\cref{fig:returns,tab:s-epic}). We scan the 76,885 narrated actions across 633 videos and 86 hours of unscripted kitchen activity in the public training and validation annotations (\texttt{epic\_returns.py}). Within each video, actions are ordered by their start times. An opening is \emph{undone} when the next opening or closing of the same noun class in that video is a closing, and switching on and taking are undone in the same way by switching off and putting. Nouns are matched by class, so the opening and closing of two cupboards pair as one, and the narration omits some actions. Most undos follow within 30~s, and 0.58 of opened cupboards, 0.62 of opened drawers and 0.73 of opened fridges are closed again, and 0.78 of taps turned on are turned off. A put need not return an item to the spot it came from, so the last row of \cref{tab:s-epic} bounds item returns from above, and the first two rows count returns of the state of a place. Because the wearer's camera films these returns, it observes most of them; the benchmark of \cref{sec:experiments} measures the share of returns after which the camera of a later step holds no newer view, 352 of the 1,050 repeat-state steps at seed 1.

\begin{table}[tbp]
\centering
\caption{Actions of EPIC-KITCHENS-100 undone later in the same video, with the median delay to the reverse action and the share reversed within 30~s.\label{tab:s-epic}}
\small\setlength{\tabcolsep}{6pt}
\begin{tabular}{lrrrr}
\toprule
Pair & Actions & Undone & \begin{tabular}[b]{@{}r@{}}Median\\delay\\(s)\end{tabular} & \begin{tabular}[b]{@{}r@{}}Undone\\within\\30~s\end{tabular} \\
\midrule
Open, close & 5,680 & 0.58 & 4.9 & 0.89 \\
Turn on, turn off & 2,596 & 0.66 & 7.4 & 0.81 \\
Take, put & 16,785 & 0.58 & 5.4 & 0.87 \\
\bottomrule
\end{tabular}
\end{table}

\section{Extended related work}

This appendix extends \cref{sec:background}. \Cref{tab:s-neighbours} sets \ours{} beside the memories closest to it. After the agent restores a region, a deleted, masked or decayed view is gone, a filtered frame returns only when every region of the target view agrees with the state table, a symbolic state returns without the appearance that only a stored view holds, and a re-dated view returns with its appearance and a timestamp the generator trusts.

\paragraph{Long rollouts and spatial memory.}

Interactive generators condition each chunk on the past through differing frame noise or training on their own rollouts~\citep{chen2024diffusion,huang2025self,sun2025worldplay}, and spatial memories address stored frames by view overlap, surfels, point clouds or learned keys and keep them compact as key-value states or trained weights~\citep{wu2025video,yu2026mosaicmem,yi2026worldkv,wu2026world}; a controlled study compares such designs on revisits~\citep{king2026echo}. \ours{} leaves retrieval and compression as they are and changes only the timestamps of frames whose content the agent restored and of tokens whose content the agent changed, so it applies to every such memory that dates its frames.

\paragraph{Time-aware memory beyond video.}

Long-term mapping keeps experiences of semi-static scenes and estimates how long features persist~\citep{churchill2013experience,rosen2016lifelong,saavedraruiz2025perpetua}, and language agents weight memories by recency and keep outdated facts on a timeline~\citep{park2023generative,dhingra2022time,rasmussen2025zep,wang2026toki}. \ours{} brings valid time to the frames that condition a generator, where the agent's own actions supply it. Where actors keep no log, video reasoners that ground when events happen~\citep{zhu2026tempo} can write one.
\end{document}